\documentclass[times,twocolumn,final,longtitle]{elsarticle}

\usepackage{marvosym}

\newcommand{\arxiv}{1} %% uncomment this line to build the arXiv version instead of MedIA

\ifdefined\arxiv
    \usepackage{medima-arxiv}
    \newcommand{\writer}[1]{~}
    \newcommand{\orcid}[1]{\href{https://orcid.org/#1}{\includegraphics[scale=0.09]{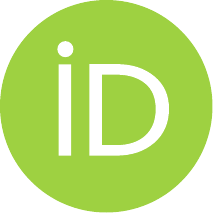}}}
    \newcommand{\mailto}[1]{\href{mailto:#1}\textsuperscript{\textsuperscript{\Letter,}} }
    \journal{}
    \date{}

\else
    \usepackage{medima}
    \newcommand{\writer}[1]{}
    \newcommand{\orcid}[1]{}
    \newcommand{\mailto}[1]{}
\fi

\usepackage{framed}
\usepackage{multirow}
\usepackage{multicol}
\usepackage{tabularx}
\usepackage{booktabs}
\usepackage{amsmath}
\usepackage{amssymb}
\usepackage{latexsym}
\usepackage[obeyspaces]{url}
\usepackage[dvipsnames]{xcolor}
\usepackage{colortbl}
\usepackage{rotating}
\usepackage{makecell}
\usepackage{pifont}
\usepackage{array}
\usepackage{etoolbox}
\usepackage{subcaption}
\usepackage[colorlinks=true,citecolor=blue,urlcolor=blue]{hyperref}
\AtBeginDocument{%
  \hypersetup{
    citecolor=teal,
    linkcolor=blue,   
    urlcolor=Blue}}

\newcolumntype{P}[1]{>{\centering\arraybackslash}p{#1}}

\newif\ifrevisionmode
\revisionmodefalse
\definecolor{revcolor}{HTML}{C00000}
\newcommand{\rev}[1]{\ifrevisionmode{\color{revcolor}#1}\else#1\fi}
\newcommand{\revtable}{\ifrevisionmode\color{revcolor}\fi}

\preto\tabular{\setcounter{magicrownumbers}{0}}
\newcounter{magicrownumbers}

\definecolor{newcolor}{rgb}{.8,.349,.1}
\journal{Medical Image Analysis}

\begin{document}

\verso{Shi Li \textit{et~al.}}

\begin{frontmatter}

\title{{\bf SurgTEMP: Temporal-Aware Surgical Video Question Answering with Text-guided Visual Memory for Laparoscopic Cholecystectomy}}%
% \tnotetext[tnote1]{This is an example for title footnote coding.}

\author[1]{ 
\orcid{0009-0005-9874-3957} Shi \snm{Li} \mailto{shi.li@ext.ihu-strasbourg.eu}\corref{cor1}} \cortext[cor1]{Corresponding author
}
\author[1]{Vinkle \snm{Srivastav} \mailto{srivastav@unistra.fr}}
\author[4,1,2]{Shih-Min \snm{Yin} \mailto{shih-min.yin@ihu-strasbourg.eu}}
\author[1]{Nicolas \snm{Chanel} \mailto{nicolas.chanel@etu.unistra.fr }}
\author[1]{Saurav \snm{Sharma} \mailto{ssharma@unistra.fr}}
\author[1]{Nabani \snm{Banik} \mailto{nabani.banik@ext.ihu-strasbourg.eu}}
\author[1]{Lorenzo \snm{Arboit} \mailto{lorenzo.arboit@ext.ihu-strasbourg.eu}}
\author[1]{Kun \snm{Yuan} \mailto{Kun.YUAN@ext.ihu-strasbourg.eu}}
\author[1,3]{Pietro \snm{Mascagni} \mailto{p.mascagni@unistra.fr}\fnref{fn2}}
\author[1]{Nicolas \snm{Padoy} \mailto{npadoy@unistra.fr}\fnref{fn2}}

\address[1]{University of Strasbourg, CNRS, INSERM, ICube, UMR7357, Strasbourg, France}
\address[2]{IHU Strasbourg, Strasbourg, France}
\address[3]{Fondazione Policlinico Universitario Agostino Gemelli IRCCS, Rome, Italy}
\address[4]{Department of General Surgery, Kaohsiung Chang Gung Memorial Hospital, Chang Gung University College of Medicine, Kaohsiung, Taiwan}

\fntext[fn2]{These authors contributed equally as co-last authors.}

\received{: }
\finalform{:}
\accepted{:}
\availableonline{:}
\communicated{:}

\begin{abstract}
% 1. Context & Motivation
Surgical procedures are inherently complex and risky, requiring extensive expertise and constant focus to well navigate evolving intraoperative scenes. Computer-assisted systems such as surgical visual question answering (VQA) offer promises for education and intraoperative support.
% 2. Problem & Gap
Current surgical VQA research largely focuses on static frame analysis, overlooking rich temporal semantics. Surgical video question answering is further challenged by low visual contrast, its highly knowledge-driven nature, diverse analytical needs spanning scattered temporal windows, and the hierarchy from basic perception to high-level intraoperative assessment. To address these challenges, we propose \textit{SurgTEMP}, a multimodal LLM framework featuring (i) a query-guided token selection module that builds hierarchical visual memory (spatial and temporal memory banks) and (ii) a Surgical Competency Progression (SCP) training scheme. Together, these components enable effective modeling of variable-length surgical videos while preserving procedure-relevant cues and temporal coherence, and better support diverse downstream assessment tasks.
To support model development, we introduce CholeVidQA-32K, a surgical video question answering dataset comprising 32K open-ended QA pairs and 3,855 video segments (approximately 128 h total) from laparoscopic cholecystectomy. The dataset is organized into a three-level hierarchy---Perception, Assessment, and Reasoning---spanning 11 tasks from instrument/action/anatomy perception to Critical View of Safety (CVS), intraoperative difficulty, skill proficiency, and adverse event assessment.
In comprehensive evaluations against state-of-the-art open-source multimodal and video LLMs (fine-tuned and zero-shot), SurgTEMP achieves substantial performance improvements, advancing the state of video-based surgical VQA. The project page is available at \url{https://camma-public.github.io/SurgTEMP/}.
\end{abstract}

\begin{keyword}
%% Keywords
\vspace{-0.33in}
\KWD Surgical Video Question Answering \sep Surgical Safety Assessment \sep Multimodal Large Language Model.
\end{keyword}

\end{frontmatter}

%\linenumbers

%% main text
\section{Introduction}
\begin{figure*}[htbp]
    \centering
    \includegraphics[width=1.0\textwidth]{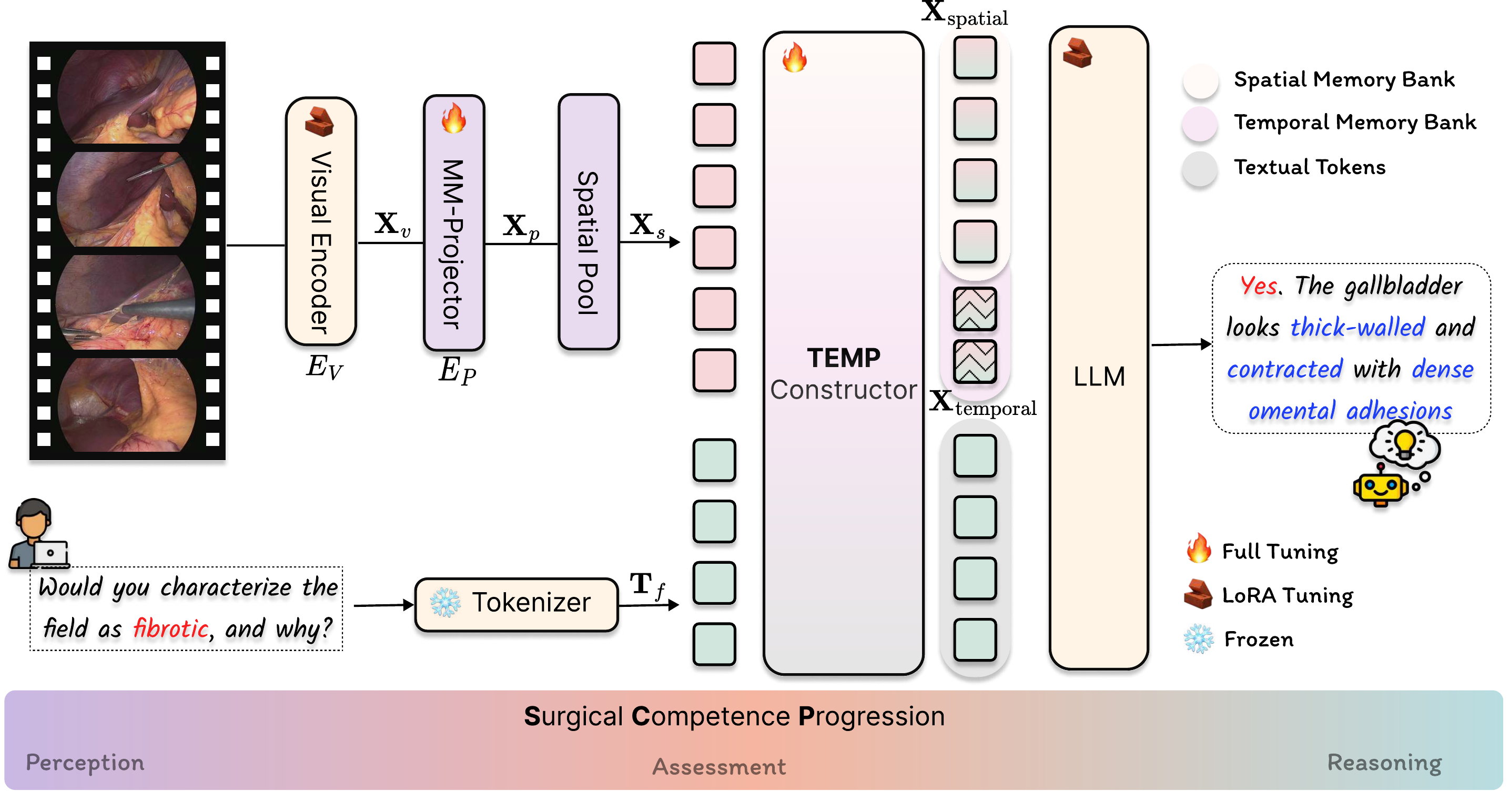}
    \caption{Architecture Overview of Our Proposed Model SurgTEMP. The sampled video frames first go through a feature extraction pipeline containing visual encoder, multi-modal projector and spatial pooling to obtain visual tokens $\mathbf{X}_s$. Textual input is processed through the textual tokenizer to yield textual tokens $\mathbf{T}_f$. Our proposed TEMP constructor takes tokens from both modalities to construct a hierarchical visual memory bank including spatial and temporal granularity. Then an LLM is employed as backbone to generate answers conditioned on visual input and textual query.}
    \label{fig:model_arch}
\end{figure*}

% 2 pages for intro
% 1. Broad Context / Importance (Why surgery is)
Surgery is a highly complex and skill-demanding discipline, characterized by substantial risk and a prolonged training period before practitioners achieve proficiency. At the beginning of training, a solid understanding of anatomy is generally assumed, as it provides the foundation for interpreting the operative field, for example the identification of key anatomical landmarks and the assessment of their status. Based on that, years of residency are typically required to master fundamental operative actions, such as suturing and tissue retraction, and to develop the ability to perform context-aware intraoperative assessments. For instance, in laparoscopic cholecystectomy surgeons must judge whether the cystic duct has been adequately dissected and exposed and whether it is safe to divide it. This competence is typically acquired only after numerous procedures. Such skill is essential for ensuring a smooth operation, minimizing complications, and achieving favorable postoperative outcomes.

Recent advances in surgical data science have endeavored to capture the aforementioned nuances in surgery. A substantial body of work has explored the perception of the surgical scene, including tools~\citep{allan20192017, Nwoye_2019, rocha2019selfsupervisedsurgicaltoolsegmentation, hasan2021detection, wagner2023comparative, tao2023last, zia2025intuitivesurgicalsurgtoollocchallenge}, anatomy~\citep{vlontzos2019multiplelandmarkdetectionusing, ALI2021102002, meyer2025s4msegment4extreme, murali2025cyclesamfewshotsurgicalscene}, and surgical actions~\citep{Nwoye_2020, Nwoye_2022, Sharma_2023, hu2024ophnet, 11084985, AYOBI2025103726}.
Beyond these foundational efforts in surgical scene understanding, work has also emerged focusing on more complex assessments of additional facets of surgery. One notable line of work centers on the Critical View of Safety (CVS)~\citep{mascagni2020formalizing}, a standardized checklist for specific procedures such as laparoscopic cholecystectomy. By ensuring that key anatomical landmarks are clearly dissected and exposed, CVS adherence reduces the risk of inadvertent injury to critical structures. Vision-based models have been proposed to predict CVS achievement by leveraging intraoperative visual cues~\citep{murali2023latent, murali2023encoding, alapatt2024jumpstarting, jaspers2025scaling}, offering automated support for this essential safety protocol.
Another promising direction involves LCOD (Laparoscopic Cholecystectomy Operative Difficulty) assessment~\citep{VANNUCCI20221158, Sharma_2025}. Scoring systems have been introduced to quantify surgical difficulty based on observable intraoperative findings (e.g., visceral fat, where thick adipose tissue may obscure anatomical landmarks and hinder instrument manipulation), providing decision support for critical choices such as whether to involve more experienced practitioners or to convert from a laparoscopic to an open approach.
Additionally, intraoperative adverse event (IAE) detection models~\citep{bose2025featuremixingapproachdetecting} have been developed to identify unexpected or undesirable incidents during surgery, including bleeding, mechanical injuries, and other deviations from the ideal intraoperative course. 
Moreover, surgical skill assessment~\citep{wagner2021comparativevalidationmachinelearning, pedrett2023technical} is a well developed modeling task that facilitates the quantification of proficiency levels in tissue manipulation, suturing, and other technical skills.

The intraoperative assessments highlighted above are inherently dynamic: they unfold progressively and can only be fully interpreted within a temporal context. Apart from the aforementioned vision-only solutions, surgical VQA systems have emerged as promising intraoperative analytical tools to support education by formulating detection tasks into interactive question-answering and reasoning processes between the user and the model, thereby offering greater flexibility and potential for broader applicability. Existing works—such as Surgical-VQA~\citep{seenivasan2022surgical}, Surgical-VQLA~\citep{bai2023surgical}, SSG-VQA~\citep{yuan2024advancing}, and PitVQA~\citep{he2024pitvqa}—have laid the foundation for this line of research by exploring the capability of VQA systems to reason about the detection and grounding of key elements (e.g., tools, anatomy, actions) in static frames from surgical videos. However, despite this promising potential, the dynamic information embedded in temporal context remains underexplored, even though it is essential for various intraoperative analytical needs, thereby limiting the practical applicability of such systems.

Addressing this gap and developing surgical video question answering systems for diverse practical intraoperative analytic needs presents several significant challenges. 

\textit{First—\textbf{domain-specific visual interpretation}:} Unlike natural images or videos, where visual semantics (i.e., objects and their defining attributes such as edges and color) are often clear and dominant, endoscopic surgical videos pose a unique challenge. Here, anatomical structures often lack strong visual contrast, and the identification of anatomical boundaries, tissue types, or pathological states often relies on domain-specific priors. Surgeons, for instance, infer granular anatomical elements by triangulating contextual information—such as procedure type, surgical phase, and the relative positions of known landmarks—alongside subtle visual cues and their understanding of typical spatial relationships and expected tissue behavior.

\textit{Second—\textbf{variable spatiotemporal granularity needs}:} Throughout the operation, numerous assessments are performed across many timespans with significantly different granularity needs. For example, the detection of LCOD findings and IAE usually requires the identification of shorter-duration events with finer granularity because they are often related to the status of anatomical landmarks, which rely on subtle visual cues such as shape and texture. In contrast, skill assessment needs a longer observation window to draw conclusions about proficiency based on the smoothness of operation and near-misses. This requires the model to have stronger spatiotemporal understanding capabilities that adapt to these various needs.

\textit{Third—\textbf{hierarchical task dependencies}:} All types of intraoperative assessments rely on the fundamental capabilities of interpreting the operative field, which form a clear hierarchical structure in which sequential atomic surgical scene understandings support higher-level assessments. For example, CVS relies on the clear identification of fully dissected and exposed anatomical landmarks. 

To address these challenges, we propose a novel surgical video QA framework named \textit{SurgTEMP}. It is a multimodal LLM-based framework that contains a visual feature extraction module and textual tokenization to produce tokens from both modalities and feed them into a backbone LLM to generate textual responses as shown in Figure~\ref{fig:model_arch}.
At its core is a \textbf{TEMP} (\textbf{TE}xt-guided \textbf{M}emory \textbf{P}yramid) constructor—a cross-modal learnable visual token arrangement module specifically designed to balance two key aspects in the knowledge-intensive surgical domain. First, it captures fine-grained details in short yet significant spatiotemporal contexts via a spatial memory bank (e.g., proper dissection to expose critical anatomy). Second, it preserves lengthy spans of progressive analytical signals via a temporal memory bank (e.g., skill proficiency assessment over a surgical phase, reflecting cumulative decision points in operative field manipulation). Additionally, a domain-specific training scheme—\textbf{S}urgical \textbf{C}ompetency \textbf{P}rogression (SCP)—progressively builds the model's capabilities from fundamental perception to higher-level assessment and reasoning tasks.
%  Yet, it differentiates itself with a TEMP constructor that uses input textual queries as guiding signals to construct a hierarchical visual memory reflecting both fine-grained visual cues and long-term temporal context. Specifically, tokenized textual embeddings attend over patch-wise visual tokens to construct a multi-level attention map (patch-level and frame-level). Visual tokens are selected based on frame-level importance scores by performing differentiable and learnable Gumbel-Softmax top-k selection and weighted by patch-level attention weights to form a spatial memory bank. Simultaneously, patch-wise visual tokens are pooled and reweighted by frame-level attention scores to construct a temporal memory bank.

The design choices and training strategy directly address the aforementioned domain challenges. 
\textit{First}, by utilizing the textual query as the guiding signal, the model learns to form informative spatial memory banks that benefit from the embedded knowledge in the pretrained backbone MLLM and accommodate the domain shift in the downstream fine-tuning dataset, which mitigates the challenge of limited visual contrast and takes advantage of the knowledge-driven nature of surgical videos.
\textit{Second}, by constructing a hierarchical memory bank that covers fine-grained and short-term memory as well as long-term yet coarser memory, our model gains enhanced capabilities for handling variable spatiotemporal attentional needs from different assessment tasks.
\textit{Third}, the SCP training scheme progressively trains the model from fundamental perception tasks to higher-level assessment and reasoning tasks, directly addressing the challenge of hierarchical task dependencies by explicitly modeling the structure in which basic surgical scene understanding serves as a prerequisite for more complex intraoperative assessments.

\begin{figure*}[t!]
    \centering
    \includegraphics[width=0.95\textwidth]{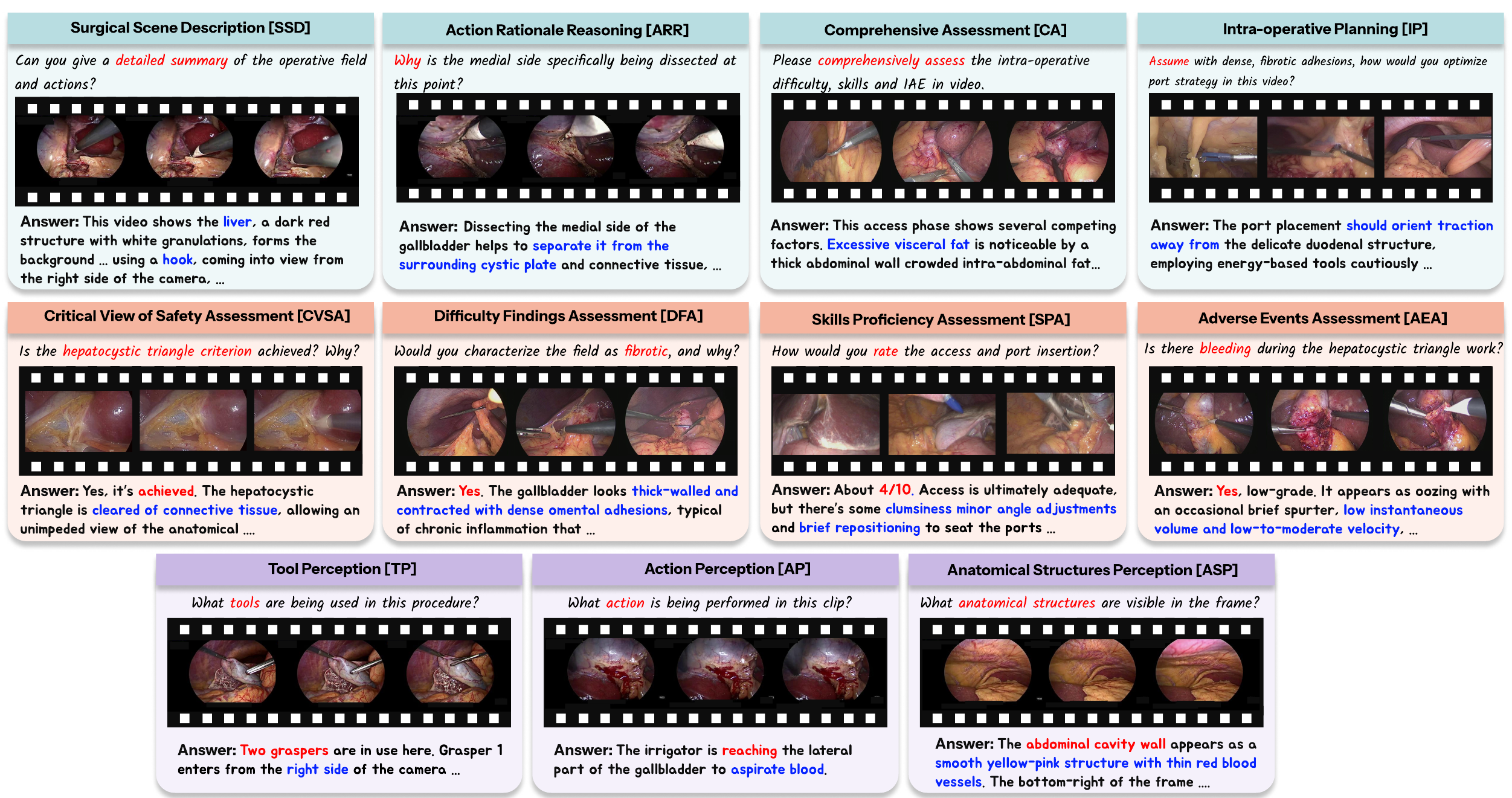}
    \caption{Hierarchical overview of the CholeVidQA-32K dataset. The 11 tasks are categorized into three capability levels: \textcolor[HTML]{6B4C9A}{\textbf{Perception}} (basic surgical scene understanding), \textcolor[HTML]{C85A3A}{\textbf{Assessment}} (surgical assessment for multiple ends), and \textcolor[HTML]{3D8A92}{\textbf{Reasoning}} (complex surgical scene analysis). In the question and answer examples, text indicating task-specific characteristics is shown in \textcolor[HTML]{FF0404}{red}, while descriptive elements are highlighted in \textcolor[HTML]{0724FF}{blue}.}
    \label{fig:hierarchy}
\end{figure*}

% 4. Your Contributions (What you propose)
Moreover, to facilitate model training and evaluation, we introduce CholeVidQA-32K, an instruction-tuning dataset that distinguishes itself from existing resources through dynamic temporal context and broader assessment coverage (Table~\ref{tab:dataset_comparison}). To address the challenge in the surgical domain regarding the hierarchical nature of intraoperative assessments, we formulate and organize 11 tasks into a three-level hierarchy (Figure~\ref{fig:hierarchy}).

At the \textit{Perception} level, we design three fundamental tasks for surgical scene understanding. \textit{Tool Perception} facilitates contextually dynamic responses that reference tools by position and movement. \textit{Action Perception} captures surgical actions with clinical rationale. \textit{Anatomical Structure Perception} provides fine-grained descriptions of anatomical landmarks, including attributes such as status, position, shape, and color. Data for this level is derived from the CholecT50~\citep{Nwoye_2022} dataset, where we segment videos by action continuity and incorporate structured caption annotations from clinicians, repurposed into semantically rich QA pairs. These short-span video segments preserve action-oriented visual dynamics. Together, these three tasks establish fundamental visual recognition capabilities for surgical scenes.

At the \textit{Assessment} level, we design tasks to support diverse intraoperative analysis needs, including \textit{Critical View of Safety}, \textit{Difficulty Findings}, \textit{Skills Proficiency} and \textit{Adverse Events} assessments. These tasks encompass different analytical scenarios and predictive formulations, including binary and multi-category classification. Data for this level is sourced from Endoscapes~\citep{Mascagni2025} and CholeScore~\citep{Sharma_2025}. We segment Endoscapes videos using fixed 5-second intervals and CholeScore videos by surgical phase, resulting in two distinct temporal scales—shorter clips for fine-grained assessment and longer segments spanning minutes to tens of minutes for comprehensive evaluation. We transform existing categorical annotations into open-ended question-answer pairs with enriched descriptive explanations.

Finally, at the \textit{Reasoning} level, we support more complex clinical analysis by combining operative field interpretation and assessments across video segments. With the assistance of large language models (LLMs), we generate condensed descriptive QA pairs on basic scene understanding (i.e., \textit{Surgical Scene Description}) and on various intra-operative assessment analysis (i.e., \textit{Comprehensive Assessment}). To further stimulate the models' reasoning capability, we create reasoning questions by conditioning on action intentions (i.e., \textit{Action Rationale Reasoning}) and intraoperative variables (i.e., \textit{Intra-operative Planning}), enabling scenario-based analysis.

We evaluate our proposed model using three tiers of metrics: classification metrics reflecting categorical capability, overlap metrics reflecting textual alignment with ground-truth answers, and multifaceted LLM-based scores reflecting response quality across multiple dimensions including correctness, relevance, and linguistic quality.

Experimental results broken down across multiple levels show that our model outperforms strong general-domain video QA baselines under both fine-tuned and zero-shot settings on our dataset, with notable margins on longer and more complex tasks.

\rev{Furthermore, we develop an offline expert-verified test subset consisting of a carefully selected subset of test videos and questions with high-quality answers manually written by clinicians. This set provides an additional evaluation of alignment with clinician-curated answers: SurgTEMP outperforms the baselines under the reported metrics.}

Our contributions can be summarized as follows:
\begin{itemize}
    \item We propose SurgTEMP, a model capable of handling variable-length surgical videos and tasks requiring different granularities of visual cues, advancing the state of surgical video question answering.
    \item We introduce a novel surgical video question answering dataset, CholeVidQA-32K, featuring dynamic temporal context and broad task coverage, facilitating video-based visual question answering research across various downstream applications.
    \rev{\item We evaluate our model on a carefully curated offline expert-verified test subset with high-quality answers manually written by clinicians. Results show superior alignment with these clinician-curated answers compared to all baselines, while remaining distinct from prospective clinical validation.}
\end{itemize}

\section{Related work}
% 2 pages for related work

\subsection{Surgical Computer Vision for Laparoscopic Videos}
% tools/action/anatomy/pose estimation/classification
Recent advances in deep learning have enabled sophisticated modeling of intraoperative dynamics, particularly in laparoscopic surgery, establishing the foundation for contemporary surgical computer vision research. The field has evolved to address surgical scene understanding across multiple levels of semantic granularity. Significant attention has been directed toward action triplet recognition~\citep{Nwoye_2020, Nwoye_2022, Nwoye_2023, Sharma_2023, sharma2023surgicalactiontripletdetection, hu2024ophnet, 11084985, AYOBI2025103726}, which involves predicting structured tool-action-anatomy relationships (e.g., grasper-grasp-gallbladder) from video sequences and is commonly formulated as a structured multi-label set prediction problem at the frame or short-clip level. A representative benchmark is CholecT50~\citep{Nwoye_2022}, which provides frame-level tool-action-anatomy triplet annotations for laparoscopic cholecystectomy videos. Complementing these temporal modeling efforts, the detection and tracking of anatomical structures~\citep{vlontzos2019multiplelandmarkdetectionusing, ALI2021102002, luengo20222020cataractssemanticsegmentation, Meyer_2024, meyer2025s4msegment4extreme} and surgical instruments~\citep{allan20192017, Nwoye_2019, rocha2019selfsupervisedsurgicaltoolsegmentation, hasan2021detection, wagner2023comparative, tao2023last, zia2025intuitivesurgicalsurgtoollocchallenge} have been extensively studied, typically formulated as object detection and semantic segmentation tasks.

% \subsection{Data-driven Intra-operative Assessment}
Recognizing the high-stakes nature of laparoscopic surgery, researchers have increasingly focused on data-driven approaches to capture and enhance surgical safety. A prominent line of research centers on the Critical View of Safety (CVS), a procedure-specific checklist designed to guide surgeons in achieving clear visual exposure of critical anatomical landmarks, thereby minimizing the risk of inadvertent injuries such as bile duct damage during laparoscopic cholecystectomy~\citep{Mascagni_2021}. The annotation process for CVS assessment typically requires expertise from multiple levels of clinical experience and adherence to detailed protocols~\citep{mascagni2021surgicaldatasciencesafe} to ensure consistency and minimize inter-annotator variability. The Endoscapes dataset~\citep{Mascagni2025} was developed to support this research direction, providing 201 laparoscopic videos with binary CVS labels at fixed 5-second intervals indicating whether each of the three established criteria is achieved, along with sparse bounding box and segmentation mask annotations. Models including LG-CVS~\citep{murali2023latent} and Jumpstart~\citep{alapatt2024jumpstarting} have been developed using this dataset to predict the achievement of each criterion, facilitating automated CVS assessment.

Apart from CVS evaluation, surgical difficulty assessment has emerged as another critical safety consideration. Established scoring systems, such as the Nassar and Sugrue scores and the Parkland Grading Scale (PGS), quantify procedural complexity based on intraoperative findings. Recent computational approaches, including SurgPrOD~\citep{Sharma_2025}, have demonstrated the feasibility of predicting difficulty levels from early-stage surgical videos and introduced CholeScore, which provides videos annotated with the binary presence of difficulty findings across three scales. Additionally, the detection of intraoperative adverse events (IAE)—such as bleeding and mechanical injuries—represents a crucial component of surgical safety monitoring. These events, while potentially emergent, have significant implications for surgical outcomes. Recent work, including BetaMixer~\citep{bose2025featuremixingapproachdetecting}, has explored automated detection of such events in gastric bypass procedures. Moreover, automated surgical skill assessment has emerged as a promising research direction for understanding and improving the human factor in the operating room~\citep{gao2014jhu, liu2021towards}. Most works use the OSATS~\citep{martin1997objective} protocol to evaluate proficiency levels for surgical operations.

Our dataset curation pipeline leverages existing videos from prior vision-centric datasets, including CholecT50, Endoscapes, and CholeScore. For generic surgical scene perception, we adopt clinician-annotated captions to provide visually contextual textual descriptions. For higher-level intraoperative assessments, in light of the high annotation cost, we repurpose existing structured labels into open-ended question–answer pairs and incorporate enriched descriptions from annotation definitions. Although our QA pairs are derived from these existing labels, the proposed tasks differ fundamentally from the original visual-centric formulations: they require generating open-ended, and explanatory, natural language answers conditioned on questions. Consequently, we focus on comparisons within the VQA formulation (including strong multimodal LLM baselines) and regard prior visual-centric works as complementary foundations for label curation rather than directly competing baselines.

\subsection{Multimodal Large Language Models}
The emergence of Large Language Models (LLMs) has significantly transformed artificial intelligence capabilities~\citep{openai2024gpt4technicalreport,deepseekai2025deepseekv3technicalreport}, which have demonstrated remarkable world knowledge and reasoning capabilities acquired through training on vast textual corpora, fundamentally enhancing human-AI interaction and problem-solving potential. The critical aspects of these models are the multimodal understanding. A common design is the encoder alignment approach~\citep{liu2023visualinstructiontuning, ye2024mplug, zhu2023minigpt4enhancingvisionlanguageunderstanding, li2023blip2bootstrappinglanguageimagepretraining, li2025ottermultimodalmodelincontext, laurençon2024mattersbuildingvisionlanguagemodels, lin2024vilapretrainingvisuallanguage}, which leverage strong pretrained vision encoders with lightweight adapters to align image features to an LLM. Another family uses a unified pretraining paradigm, where both image and text modalities are jointly trained from the start to ensure deeper integration~\citep{bai2023qwenvlversatilevisionlanguagemodel, li2022mplugeffectiveefficientvisionlanguage, alayrac2022flamingovisuallanguagemodel, chen2023palijointlyscaledmultilinguallanguageimage, chen2023palixscalingmultilingualvision, huang2023languageneedaligningperception, peng2023kosmos2groundingmultimodallarge}.

Extending beyond static image understanding, one line of work on video-enabled multimodal LLMs~\citep{maaz2024videochatgptdetailedvideounderstanding, li2024llavanextinterleavetacklingmultiimagevideo} builds upon image-focused frameworks and treats a sequence of frames as a set of independent images. A second line explicitly introduces a dedicated encoder for video input~\citep{li2024videochatchatcentricvideounderstanding, lin2024videollavalearningunitedvisual}. To address the challenges posed by long-form video and the limited context window of LLMs, recent models adopt diverse architectural strategies. mPLUG-Owl3~\citep{ye2024mplug} introduces a hyper-attention block as a fundamental architectural change for improved cross-modal fusion. LLaVA-Video~\citep{zhang2024video} incorporates a slow--fast visual feature pooling scheme with multiple strides to provide two levels of visual granularity. InternVideo2.5 uses token-merging and dropout to reduce the number of visual tokens and extend the effectively perceivable video length. LongVA transfers reasoning capabilities from image pretraining and scales the model at test time for long-context analysis by representing video as multi-grid images.

Our method is a domain-tailored extension of recent video LLM architectures. It departs from prior work through a text-guided hierarchical memory bank that jointly maintains fine- and coarse-grained surgical visual representations, balancing localized operative field interpretations with broader procedural context. This design foregrounds knowledge-driven cues (e.g., anatomical landmarks) that are less visually salient yet conceptually defined, in contrast to general-domain settings. Accordingly, we benchmark against open-source video LLMs to demonstrate the effectiveness of our architecture in capturing surgical nuance.

\begin{table*}[t!]
\centering
\caption{Comparison of surgical VQA datasets. Our dataset distinguishes itself through video-based input, significantly longer average word count, and comprehensive coverage across perception, assessment, and reasoning tasks.}
\label{tab:dataset_comparison}
\begin{tabular}{lccccc}
\hline
\textbf{Dataset} & \textbf{Visual Input} & \makecell{\textbf{Average}\\\textbf{Word Count}} & \makecell{\textbf{Perception}\\\textbf{tasks}} & \makecell{\textbf{Assessment}\\\textbf{tasks}} & \makecell{\textbf{Reasoning}\\\textbf{tasks}} \\
\hline
EndoVis-18-VQA~\citep{seenivasan2022surgical} & image & 5.8 & {\color{green!70!black}\ding{51}} & {\color{red}\ding{55}} & {\color{red}\ding{55}} \\
Cholec80-VQA~\citep{seenivasan2022surgical} & image & 2.0 & {\color{green!70!black}\ding{51}} & {\color{red}\ding{55}} & {\color{red}\ding{55}} \\
SSG-VQA~\citep{yuan2024advancing} & image & 12.8 & {\color{green!70!black}\ding{51}} & {\color{red}\ding{55}} & {\color{red}\ding{55}} \\
PitVQA~\citep{he2024pitvqa}         & image & 10.3 & {\color{green!70!black}\ding{51}} & {\color{red}\ding{55}} & {\color{red}\ding{55}} \\
Surg-396K~\citep{wang2025endochatgroundedmultimodallarge} & image & - & {\color{green!70!black}\ding{51}} & {\color{red}\ding{55}} & {\color{green!70!black}\ding{51}} \\
\hline
\textbf{CholeVidQA-32K} & \textbf{video} & \textbf{70.5} & {\color{green!70!black}\ding{51}} & {\color{green!70!black}\ding{51}} & {\color{green!70!black}\ding{51}} \\
\hline
\end{tabular}
\end{table*}

\subsection{Visual Question Answering: From General Domain to Surgery}
Visual Question Answering (VQA) has played a pivotal role in advancing multimodal understanding, with large-scale datasets providing the foundation for training and evaluation across diverse scenarios. The VQA v1 and v2 datasets~\citep{agrawal2016vqavisualquestionanswering} were among the earliest benchmarks, pairing open-ended questions with natural images to establish the fundamental paradigm. Subsequent developments introduced increasingly sophisticated challenges: GQA~\citep{hudson2019gqanewdatasetrealworld} emphasized compositional reasoning and scene graph grounding, while VizWiz~\citep{gurari2018vizwizgrandchallengeanswering} introduced real-world complexity through images captured by visually impaired users. The field further evolved toward knowledge-intensive reasoning with datasets like OK-VQA~\citep{marino2019okvqavisualquestionanswering} and A-OKVQA~\citep{schwenk2022aokvqabenchmarkvisualquestion}, which required external knowledge integration and commonsense understanding. More recently, with the emergence of multimodal large language models, comprehensive evaluation suites such as MMBench~\citep{liu2024mmbenchmultimodalmodelallaround}, SEED-Bench~\citep{li2023seedbenchbenchmarkingmultimodalllms}, and MM-Vet~\citep{yu2024mmvetevaluatinglargemultimodal} have been developed to systematically assess capabilities spanning perception, reasoning, and instruction-following across diverse scenarios.

Building upon these general-domain foundations, surgical visual question answering has emerged as a specialized paradigm that adapts VQA principles to address the unique challenges of surgical environments, offering promising potential for developing flexible interfaces and educational tools that bridge computer vision and natural language understanding in clinical contexts. Surgical-VQA~\citep{seenivasan2022surgical} pioneered this research direction by leveraging existing annotations from the EndoVis17 and EndoVis18 datasets~\citep{allan20192017} to create question-answer pairs focused on instrument recognition and basic scene understanding. Building upon this foundation, Surgical-VQLA~\citep{bai2023surgical} extended the framework by incorporating visual grounding capabilities, enabling models to localize relevant regions while answering questions. SSG-VQA~\citep{yuan2024advancing} further enhanced spatial reasoning by integrating scene graphs as additional input modalities to improve the model's spatial awareness and relational understanding. The scope of surgical VQA has also expanded beyond cholecystectomy procedures, with PitVQA~\citep{he2024pitvqa} extending the paradigm to pituitary surgery, demonstrating the generalizability of the approach across different surgical procedures. More recently, EndoChat~\citep{wang2025endochatgroundedmultimodallarge} has attempted to unify various spatial recognition tasks within a comprehensive VQA framework.

Despite these advances, existing surgical VQA systems remain predominantly focused on single-frame analysis and generic scene-understanding tasks. Our work extends this paradigm by introducing video-based input and broader coverage of assessment tasks, including safety- and procedure-oriented reasoning. Because prior models are predominantly trained on static imagery and are not optimized for temporal integration, their reported performance does not necessarily reflect capabilities under video-based evaluation. Accordingly, we position earlier systems as conceptual precedents rather than directly comparable modeling baselines. \rev{Within the broader perception--memory--reasoning--action continuum of embodied procedure-assistance systems~\citep{yao2026advancing}, SurgTEMP is a video-based perception and temporal-reasoning research component; it does not claim robotic embodiment, action planning, or closed-loop control.}

\section{Methodology}
% 5 pages for methodology (figures included)
\begin{figure*}[htbp]
    \centering
    \includegraphics[width=1.0\textwidth]{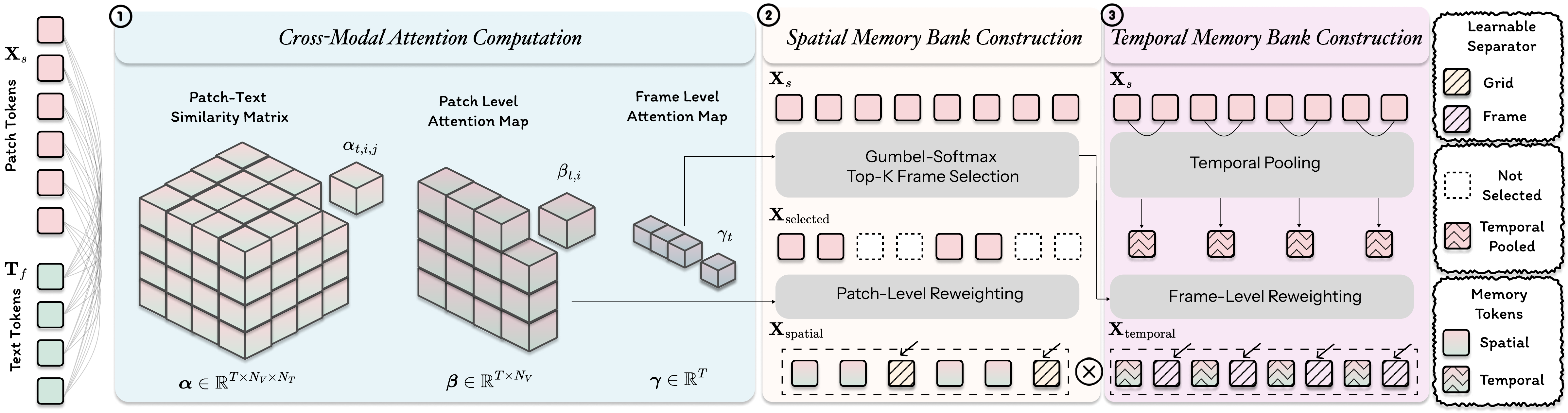}
    \caption{Illustration of our proposed TEMP module. It contains three processing steps. First, multi-level text-visual attention maps are computed. Second, the spatial memory bank is constructed by selecting frames based on frame-level attention and reweighting visual tokens with the patch-level attention map. Third, the temporal memory bank is formed by temporal pooling and reweighting with the frame-level attention map.}
    \label{fig:TEMP}
\end{figure*}

\subsection{SurgTEMP}

\subsubsection{Modeling Pipeline Overview}
As shown in Figure~\ref{fig:model_arch}, the architecture comprises four key components working synergistically: (1) a \textit{SigLIP Visual Encoder}~\citep{zhai2023sigmoid} employing a Vision Transformer for patch-based feature extraction; (2) a \textit{Multi-Modal Projector} with spatial pooling that bridges visual and textual representation spaces; (3) a novel \textit{Text-Guided Memory Pyramid (TEMP)} module that constructs hierarchical visual memories through cross-modal attention and differentiable frame selection; and (4) a \textit{Qwen2-7B Large Language Model} backend for multimodal fusion and response generation. \rev{The Qwen2-7B backbone is not selected as an isolated language model: SurgTEMP is initialised from the LLaVA-Video pretrained checkpoint~\citep{zhang2024video}, of which the language model, visual encoder, projector alignment, tokenizer, and video instruction-tuned weights form a single pretrained multimodal video backbone. Substituting a newer language model would require rebuilding that multimodal alignment and video instruction-tuning stage before any surgical-domain adaptation, and would therefore confound the contribution of the proposed TEMP module, which is what these experiments are designed to isolate.}

Our key contribution lies in the TEMP constructor module, which dynamically selects the most relevant frames using Gumbel-Softmax mechanisms and constructs a dual-granularity memory pyramid comprising both fine-grained spatial details for operative field analysis and coarse-grained temporal context for procedural awareness. This hierarchical representation is enhanced by learnable separator tokens that adaptively encode spatial-temporal boundaries between patch tokens. We employ a hybrid parameter optimization strategy combining parameter-efficient LoRA adapters for heavily parameterized components (visual encoder and LLM backbone) with full fine-tuning for lightweight architectural elements (multi-modal projector and separator tokens), enabling effective domain adaptation while maintaining computational efficiency. 

\subsubsection{Visual Feature Extraction}
\textbf{Uniform Frame Sampling Strategy:} To effectively process surgical videos of varying durations while maintaining computational efficiency, we employ a uniform frame sampling strategy that preserves temporal consistency across different video lengths. For each input video, we sample a fixed number of frames $T$ at regular intervals throughout the entire temporal sequence. The uniform sampling interval $\Delta \tau = \frac{L}{T-1}$ is computed based on the video length $L$ and desired frame count $T$, allowing the model to capture both fine-grained procedural details in short segments and long-range temporal dependencies in extended surgical phases. This strategy ensures that critical safety-relevant visual cues are systematically preserved across the temporal dimension while maintaining manageable computational complexity.

\textbf{SigLIP Visual Encoder:} The visual encoding process begins with input video frames $\mathbf{X} \in \mathbb{R}^{T \times C \times W \times H}$, where $T$, $C$, $W$, and $H$ respectively denote the number of uniformly sampled frames, color channels, frame width, and frame height. The visual encoder processes each frame independently through patch-based tokenization, dividing individual frames into non-overlapping spatial patches:
\begin{equation}
\label{eq:visual_encoder}
E_V: \mathbb{R}^{T \times C \times W \times H} \rightarrow \mathbb{R}^{T \times M \times D_V}
\end{equation}
in which the input frames are transformed into dense visual representations:
\begin{equation}
\label{eq:visual_features}
\mathbf{X}_v = E_V(\mathbf{X})
\end{equation}
where $M$ represents the total number of spatial patches per frame and $D_V$ denotes the dimensionality of the visual embedding space.

\textbf{Multi-Modal Projector:} The multi-modal projector serves as a critical alignment component that bridges the domain gap between vision-specific and language-specific representation spaces, which is formally defined as:
\begin{equation}
\label{eq:projector}
E_P: \mathbb{R}^{T \times M \times D_V} \rightarrow \mathbb{R}^{T \times M \times D}
\end{equation}
where $D$ represents the target embedding dimension shared between visual and textual modalities. The projector is implemented as a multi-layer perceptron (MLP) with non-linear activations between layers:
\begin{equation}
\label{eq:projection}
\begin{aligned}
h^{(0)} &= \mathbf{X}_v \\
h^{(i)} &= \text{GELU}(W_i h^{(i-1)} + b_i), \quad i = 1, \ldots, L-1 \\
\mathbf{X}_p &= W_L h^{(L-1)} + b_L
\end{aligned}
\end{equation}
where $h^{(i)}$ denotes the hidden representation at the $i$-th layer, $L$ represents the total number of layers, $W_i$ and $b_i$ are the learnable weight matrix and bias term for the $i$-th layer, and GELU denotes the Gaussian Error Linear Unit activation function. The final projected visual representation $\mathbf{X}_p \in \mathbb{R}^{T \times M \times D}$ is obtained after $L$ layers of transformation.

\textbf{Spatial Pooling:} To balance computational efficiency with spatial detail preservation, we subsequently apply spatial pooling operation that reduces the spatial resolution while maintaining essential visual information. The spatial pooling transformation is formally defined as:
\begin{equation}
\label{eq:spatial_pooling}
\mathrm{Spatial\_Pool}: \mathbb{R}^{T \times M \times D} \rightarrow \mathbb{R}^{T \times N_V \times D}
\end{equation}
with stride parameter $p$, yielding spatially-reduced visual representations:
\begin{equation}
\label{eq:spatial_reduction}
\mathbf{X}_s = \mathrm{Spatial\_Pool}(\mathbf{X}_p)
\end{equation}
where $N_V = \lfloor \sqrt{M}/p \rfloor^{2}$ represents the reduced number of visual tokens per frame, so that the pooled tokens retain a square grid layout (for the SigLIP ViT-SO400M/14-384 encoder used here, $M = 27 \times 27 = 729$ and $p = 2$ give $N_V = 13 \times 13 = 169$), effectively managing the trade-off between spatial granularity and computational complexity. These spatially-reduced representations $\mathbf{X}_s$ serve as input to the subsequent TEMP constructor.

\subsubsection{Text Processing Pipeline}
For textual input processing, we follow the implementation in Qwen2's tokenizer. The text processing differs between training and inference to accommodate distinct input configurations. During training, the textual input consists of both question $\mathbf{Q}_{\text{ues}}$ and ground-truth answer $\mathbf{A}_{\text{ns}}$ sequences concatenated as $\mathbf{S_{\text{train}}} = [\mathbf{Q}_{\text{ues}}, \langle\textrm{sep}\rangle, \mathbf{A}_{\text{ns}}]$ to enable supervised learning through teacher forcing. During inference, only the question sequence $\mathbf{Q}_{\text{ues}}$ is provided as input for autoregressive response generation. The processing begins with tokenization using the pre-trained tokenizer $\mathcal{T}$, which segments the input sequence into discrete tokens: $\{\tau_1, \tau_2, \ldots, \tau_{N_T}\} = \mathcal{T}(\mathbf{S})$, where $N_T$ represents the total number of textual tokens and $\mathbf{S}$ denotes either $\mathbf{S_{\text{train}}}$ or $\mathbf{Q}_{\text{ues}}$ depending on the stage. 

The tokens are then transformed into dense embeddings via the embedding layer: $\mathbf{T} = E_{\text{T}}(\{\tau_1, \ldots, \tau_{N_T}\}) \in \mathbb{R}^{N_T \times D}$, where $E_{\text{T}}: \mathbb{R}^{N_T} \rightarrow \mathbb{R}^{N_T \times D}$ represents the token embedding transformation. To incorporate positional information, Rotary Position Embedding (RoPE) is employed to the embeddings using position-dependent rotation angles $\theta_i = 10000^{-2i/D}$ for each dimension pair. The final positional-encoded textual representations are obtained as $\mathbf{T}_f = \textrm{RoPE}(\mathbf{T}, \{1, 2, \ldots, N_T\})$, where $D$ denotes the unified embedding dimension shared between textual and visual modalities.

\subsubsection{Text-Guided Memory Pyramid}
With the textual tokens $\mathbf{T}_f$ and visual tokens $\mathbf{X}_s$, TEMP constructor will go through three steps, namely cross modal attention computation, spatial memory bank construction, and temporal memory construction, to build the final visual memory pyramid $\mathbf{X}_f$. Critically, during this process, answer tokens are masked to prevent the model from using future answer information to guide visual attention, ensuring that only question tokens contribute to the construction of the visual memory pyramid.

\textbf{Cross-Modal Attention Computation:} The TEMP module employs a hierarchical attention mechanism that computes cross-modal similarity between textual representations $\mathbf{T}_f \in \mathbb{R}^{N_T \times D}$ and spatially-reduced visual representations $\mathbf{X}_s \in \mathbb{R}^{T \times N_V \times D}$, progressively aggregating from fine-grained patch-text interactions to frame-level importance scores.

\textit{Patch-Text Similarity Matrix:} For each visual patch token $\mathbf{x}_{t,i} \in \mathbb{R}^{D}$ at frame $t$ and spatial position $i$, we calculate the cosine similarity with respect to all textual tokens, yielding a comprehensive similarity tensor:
\begin{equation}
\alpha_{t,i,j} = \frac{\mathbf{x}_{t,i} \cdot \mathbf{t}_j}{\|\mathbf{x}_{t,i}\|_2 \cdot \|\mathbf{t}_j\|_2}
\end{equation}
where $\alpha_{t,i,j} \in [-1,1]$ represents the cosine similarity between the $i$-th visual patch in frame $t$ and the $j$-th textual token embedding $\mathbf{t}_j$. This produces a fine-grained attention tensor $\boldsymbol{\alpha} \in \mathbb{R}^{T \times N_V \times N_T}$ capturing all pairwise patch-text interactions.

\textit{Patch-Level Attention Map:} The patch-level attention weight is obtained by aggregating similarities across all textual tokens:
\begin{equation}
\label{eq:patch_attention}
\beta_{t,i} = \frac{1}{N_T} \sum_{j=1}^{N_T} \alpha_{t,i,j}
\end{equation}
where $\beta_{t,i}$ represents the average cross-modal similarity between the $i$-th visual patch in frame $t$ and all textual tokens, resulting in a patch-level attention map $\boldsymbol{\beta} \in \mathbb{R}^{T \times N_V}$ that provides a measure of textual relevance for each spatial patch within the surgical scene.

\textit{Frame-Level Attention Map:} To capture temporal dynamics while maintaining computational efficiency, we aggregate patch-level attention weights to derive frame-level importance scores. For each frame $t$, the frame-level attention weight is computed as:
\begin{equation}
\label{eq:frame_attention}
\gamma_t = \frac{1}{N_V} \sum_{i=1}^{N_V} \beta_{t,i}
\end{equation}
representing the average textual relevance of all spatial patches within that temporal frame, yielding a frame-level attention map $\boldsymbol{\gamma} \in \mathbb{R}^{T}$ that encodes the temporal importance of each frame based on text-visual alignment.

\textbf{Spatial Memory Bank Construction:} Based on the patch and frame level attention map, visual tokens are further arranged to form the visual memory bank following steps below.

\textit{Differentiable Frame Selection:} To enable learnable and efficient frame selection, we apply the Gumbel-Softmax mechanism to identify the top-$k$ most relevant frames based on their frame-level attention scores:
\begin{equation}
\label{eq:gumbel_softmax}
{\mathbf{s}}_t = \frac{\exp((\gamma_t + g_t) / \sigma)}{\sum_{t'=1}^{T} \exp((\gamma_{t'} + g_{t'}) / \sigma)}
\end{equation}
where ${\mathbf{s}}_t \in [0,1]$ represents the selection probability for frame $t$, $g_t = -\log(-\log(\mathcal{U}(0,1)))$ denotes Gumbel noise sampled from the standard Gumbel distribution with $\mathcal{U}(0,1)$ representing the uniform distribution over $[0,1]$, and $\sigma$ is the temperature parameter controlling the sharpness of the selection distribution. This produces a probability distribution $\mathbf{s} \in \mathbb{R}^{T}$ over all frames where $\sum_{t=1}^{T} {\mathbf{s}}_t = 1$, with each element indicating the likelihood of selecting the corresponding frame. This approach maintains differentiability during training while approximating discrete selection during inference, enabling end-to-end optimization of the frame selection process. The selected frame indices are obtained by:
\begin{equation}
\label{eq:frame_selection}
\{t_1, t_2, \ldots, t_k\} = \mathrm{TopK}({\mathbf{s}}_1, {\mathbf{s}}_2, \ldots, {\mathbf{s}}_T, k)
\end{equation}
where $\mathrm{TopK}(\cdot, k)$ returns the indices of the $k$ frames with the highest selection probabilities, ensuring that only the most textually-relevant temporal segments are retained for subsequent processing.

\textit{Patch-Level Reweighting:} After frame selection, we renormalize the patch-level attention weights among the selected patches using softmax:
\begin{equation}
\tilde{\beta}_{t,i} = \frac{\exp(\beta_{t,i})}{\sum_{(t',j) \in {\mathbf{X}}_{\mathrm{selected}}} \exp(\beta_{t',j})}
\end{equation}
where ${\mathbf{X}}_{\mathrm{selected}} = \{(t,i) : t \in \{t_1, t_2, \ldots, t_k\}, i \in \{1, 2, \ldots, N_V\}\}$ denotes the set of all patch indices from the selected top-$k$ frames, and $\tilde{\beta}_{t,i}$ represents the renormalized patch-level attention weight for the $i$-th patch in frame $t$. We then compute reweighted patch representations by applying the renormalized attention weights:
\begin{equation}
\hat{\mathbf{x}}_{t,i} = \tilde{\beta}_{t,i} \cdot {\mathbf{x}}_{t,i}
\end{equation}
for all $(t,i) \in {\mathbf{X}}_{\mathrm{selected}}$, where $\hat{\mathbf{x}}_{t,i}$ denotes the attention-reweighted visual patch token that emphasizes textually-relevant spatial regions.

\textit{Spatial Memory Bank:} At the fine-grained level, we construct the spatial memory bank by concatenating the reweighted selected visual patch tokens from the top-$k$ frames with learnable grid separator tokens ${\mathbf{sep}}_{\mathrm{grid}} \in \mathbb{R}^{D}$. To preserve the spatial layout structure of the original frames, we organize patches into grid rows where $N_G = \sqrt{N_V}$ represents the number of patches per row, and insert grid separators after each complete row:
\begin{equation}
\label{eq:x_spatial}
\begin{aligned}
{\mathbf{X}}_{\mathrm{spatial}} = [&\hat{\mathbf{x}}_{t_1,1}, \ldots, \hat{\mathbf{x}}_{t_1,N_G}, {\mathbf{sep}}_{\mathrm{grid}}, \\
&\hat{\mathbf{x}}_{t_1,N_G+1}, \ldots, \hat{\mathbf{x}}_{t_1,2N_G}, {\mathbf{sep}}_{\mathrm{grid}}, \\
&\ldots, \\
&\hat{\mathbf{x}}_{t_1,(N_G-1)N_G+1}, \ldots, \hat{\mathbf{x}}_{t_1,N_V}, {\mathbf{sep}}_{\mathrm{grid}}, \\
&\hat{\mathbf{x}}_{t_2,1}, \ldots, \hat{\mathbf{x}}_{t_k,N_V}]
\end{aligned}
\end{equation}
This spatial memory bank preserves fine-grained visual details essential for operative field perception, while the learnable grid separators adaptively encode spatial row boundaries within each frame, evolving during training to optimally represent the 2D spatial layout of surgical scenes.

\textbf{Temporal Memory Bank Construction:} To construct the temporal memory bank, we first perform temporal pooling to aggregate spatial information within each frame, then apply frame-level attention reweighting based on textual relevance.

\textit{Temporal Pooling:} At the frame level, we compute temporally-aggregated representations through average pooling within each frame:
\begin{equation}
\label{eq:frame_aggregation}
{\mathbf{f}}_t = \frac{1}{N_V}\sum_{i=1}^{N_V} {\mathbf{x}}_{t,i}
\end{equation}
where ${\mathbf{f}}_t \in \mathbb{R}^{D}$ represents the aggregated feature representation for frame $t$, computed for all $T$ frames in the video sequence.

\textit{Frame-Level Reweighting:} To emphasize temporally relevant frames based on text-visual alignment, we first renormalize the frame-level attention weights among all frames using softmax:
\begin{equation}
\label{eq:frame_renorm}
\tilde{\gamma}_t = \frac{\exp(\gamma_t)}{\sum_{t'=1}^{T} \exp(\gamma_{t'})}
\end{equation}
where $\tilde{\gamma}_t$ represents the renormalized frame-level attention weight for frame $t$ among all $T$ frames. We then compute reweighted frame representations:
\begin{equation}
\hat{\mathbf{f}}_t = \tilde{\gamma}_t \cdot {\mathbf{f}}_t
\end{equation}
for all frames $t \in \{1, 2, \ldots, T\}$.

\textit{Temporal Memory Bank:} Finally, we construct the temporal memory bank by concatenating the reweighted frame representations with learnable temporal separator tokens ${\mathbf{sep}}_{\mathrm{frame}} \in \mathbb{R}^{D}$. These learnable temporal separators adapt to capture procedure-specific temporal dynamics and enable the model to encode meaningful temporal transitions between surgical phases:
\begin{equation}
\label{eq:x_temporal}
{\mathbf{X}}_{\mathrm{temporal}} = [\hat{\mathbf{f}}_{1}, {\mathbf{sep}}_{\mathrm{frame}}, \hat{\mathbf{f}}_{2}, {\mathbf{sep}}_{\mathrm{frame}}, \ldots, \hat{\mathbf{f}}_{T}]
\end{equation}
where each $\hat{\mathbf{f}}_{t}$ corresponds to the reweighted aggregated representation of frame $t$. The temporal memory bank captures procedural flow and contextual dependencies.

\textbf{Hierarchical Visual Memory Pyramid:} The final hierarchical visual memory pyramid is constructed by concatenating both memory banks:
\begin{equation}
\label{eq:memory_pyramid}
{\mathbf{X}}_f = [{\mathbf{X}}_{\mathrm{spatial}}, {\mathbf{X}}_{\mathrm{temporal}}] \in \mathbb{R}^{(k \cdot N_V + k \cdot N_G + 2T - 1) \times D}
\end{equation}
where the dimensionality comprises $k \cdot N_V$ spatial memory tokens from the selected $k$ frames, $k \cdot N_G$ learnable grid separator tokens for spatial structure encoding, $T$ temporal memory tokens representing all frames, and $T-1$ learnable frame separator tokens for temporal boundary delineation.

This dual-memory architecture enables simultaneous access to both microscopic anatomical details and macroscopic procedural context within a unified representation space. The learnable separator tokens further enhance this capability by adaptively encoding structural boundaries that align with surgical workflow patterns, making the memory pyramid particularly effective for intraoperative assessment tasks that demand both spatial precision and temporal reasoning capabilities.

\subsubsection{Multimodal Fusion and Response Generation}
The processed hierarchical visual memory pyramid ${\mathbf{X}}_f$ is concatenated with textual representations $\mathbf{T}_f$ to form the unified multimodal input sequence $\mathbf{Z} = [{\mathbf{X}}_f, \mathbf{T}_f] \in \mathbb{R}^{N_{\text{total}} \times D}$, where $N_{\text{total}}$ represents the total sequence length combining visual and textual tokens.

\textbf{Large Language Model Backend:} We employ Qwen2-7B as our foundation large language model, denoted as $F_{\text{LLM}}: \mathbb{R}^{N_{\text{total}} \times D} \rightarrow \mathbb{R}^{N_{\text{total}} \times V}$, where $V$ represents the vocabulary size. The LLM processes the multimodal sequence through its transformer architecture with multi-head self-attention and causal attention masking. During inference, the model generates responses autoregressively, while during training, it optimizes cross-entropy loss between predicted and ground-truth tokens.

\begin{figure*}[htbp]
    \centering
    \includegraphics[width=1.0\textwidth]{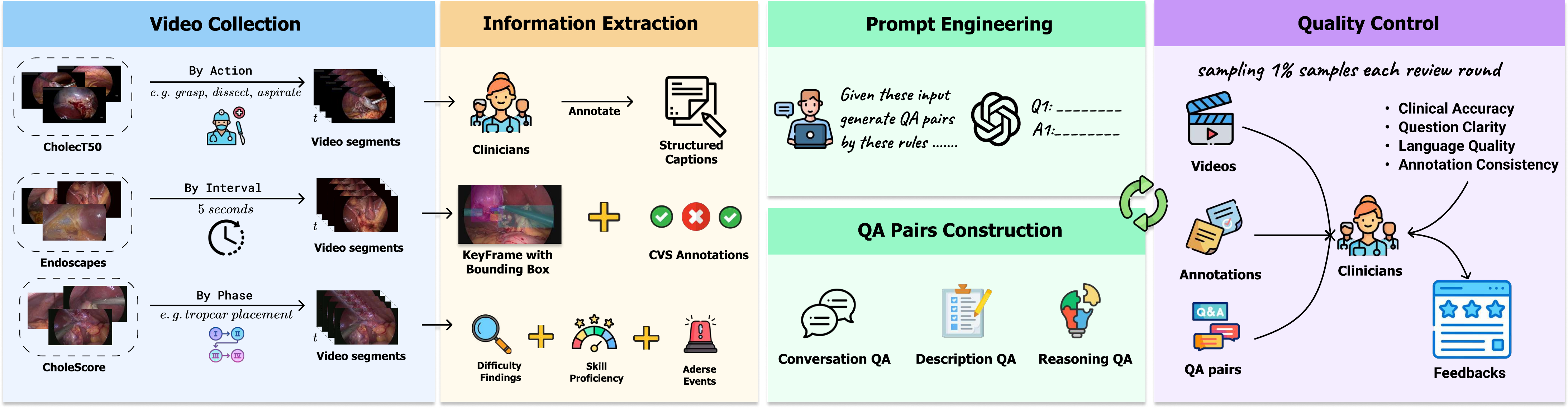}
    \caption{Curation pipeline of the proposed dataset CholeVidQA-32K.}
    \label{fig:curation_pipeline}
\end{figure*}

\subsubsection{Hybrid Parameter Training Strategy}
We employ a hybrid parameter optimization strategy that combines parameter-efficient LoRA adaptation for pre-trained components with full fine-tuning for novel architectural elements. Our trainable parameters consist of:
\begin{equation}
\label{eq:trainable_params}
\Theta = \Delta \boldsymbol{\theta}_{\text{LLM}} + \Delta \boldsymbol{\theta}_{E_V} + \boldsymbol{\theta}_{E_P} + {\mathbf{sep}}_{\mathrm{grid}} + {\mathbf{sep}}_{\mathrm{frame}}
\end{equation}
where LoRA adapters are applied to the large language model ($\Delta \boldsymbol{\theta}_{\text{LLM}}$) and visual encoder ($\Delta \boldsymbol{\theta}_{E_V}$), while the multi-modal projector parameters ($\boldsymbol{\theta}_{E_P}$) and learnable separator tokens ($\{{\mathbf{sep}}_{\mathrm{grid}}, {\mathbf{sep}}_{\mathrm{frame}}\}$) are fully parameterized. Specifically, the frozen LLM backbone $\boldsymbol{\theta}_{\text{LLM}}^{(0)}$ is augmented with low-rank adapters $\Delta \boldsymbol{\theta}_{\text{LLM}} = \mathbf{A}\mathbf{B}$ where $\mathbf{A} \in \mathbb{R}^{d \times r}$ and $\mathbf{B} \in \mathbb{R}^{r \times d}$ with $r \ll d$ and $d$ being the model's hidden dimension, and similarly for the visual encoder $\boldsymbol{\theta}_{E_V}^{(0)} + \Delta \boldsymbol{\theta}_{E_V}$. The multi-modal projector requires full optimization to establish effective cross-modal alignment, while the learnable separators ${\mathbf{sep}}_{\mathrm{grid}}$ and ${\mathbf{sep}}_{\mathrm{frame}}$ are trained from random initialization to capture surgical workflow dynamics. Following an instruction-tuning paradigm on dataset $\mathcal{D} = \left\{ \left(\mathbf{I}_i, \mathbf{y}_i\right) \right\}_{i=1}^N$, where $\mathbf{I}_i$ comprises video $\mathbf{X}$ and question $\mathbf{Q}_{\text{ues}}$, and $\mathbf{y}_i$ represents the ground-truth answer sequence $\mathbf{y}_i = \{y_{i,1}, y_{i,2}, \ldots, y_{i,L_i}\}$ of length $L_i$, we optimize the cross-entropy loss function:
\begin{equation}
\label{final_loss}
\mathcal{L}(\Theta) = -\sum_{i=1}^N \sum_{j=1}^{L_i} \log p_\Theta\left(y_{i,j} \mid \mathbf{I}_i, y_{i,<j}\right)
\end{equation}
where $y_{i,<j} = \{y_{i,1}, \ldots, y_{i,j-1}\}$ represents the preceding token sequence for autoregressive generation, and $p_\Theta\left(y_{i,j} \mid \mathbf{I}_i, y_{i,<j}\right)$ denotes the probability of generating the correct token $y_{i,j}$ given the multimodal input $\mathbf{I}_i$ and previous tokens, computed through the softmax-normalized output logits of $F_{\text{LLM}}$.

\subsubsection{Surgical Competency Progression (SCP)}
To address the inherent hierarchical dependency structure in surgical intraoperative assessments, where complex analytical tasks rely on fundamental surgical scene understanding capabilities, we design a Surgical Competency Progression (SCP) training scheme that mirrors the natural progression of surgical expertise acquisition. Our approach organizes training into three sequential stages aligned with the hierarchical task taxonomy: Perception, Assessment, and Reasoning.

\textbf{Stage-wise Task Organization:} We partition the dataset $\mathcal{D}$ into three task-level subsets corresponding to the hierarchical levels:
\begin{itemize}
    \item \textbf{Stage 1 (Perception)}: $\mathcal{D}_{\text{P}} = \{\mathcal{D}_{\text{TP}}, \mathcal{D}_{\text{AP}}, \mathcal{D}_{\text{ASP}}\}$, encompassing fundamental surgical scene understanding tasks including tool perception, action perception, and anatomical structure perception.
    \item \textbf{Stage 2 (Assessment)}: $\mathcal{D}_{\text{A}} = \{\mathcal{D}_{\text{CVSA}}, \mathcal{D}_{\text{DFA}}, \mathcal{D}_{\text{AEA}}, \mathcal{D}_{\text{SPA}}\}$, incorporating intraoperative assessment tasks including CVS evaluation, difficulty findings assessment, adverse events assessment, and surgical skill proficiency evaluation.
    \item \textbf{Stage 3 (Reasoning)}: $\mathcal{D}_{\text{R}} = \{\mathcal{D}_{\text{SSD}}, \mathcal{D}_{\text{ARR}}, \mathcal{D}_{\text{CA}}, \mathcal{D}_{\text{IP}}\}$, comprising higher-order reasoning tasks including surgical scene description, action rationale reasoning, comprehensive assessment, and intra-operative planning.
\end{itemize}

\textbf{Cumulative Training with Progressive Sampling:} Our SCP follows a cumulative training paradigm where tasks from previous stages are retained in subsequent stages, ensuring that foundational capabilities are maintained while more complex reasoning skills are developed. To balance computational efficiency with knowledge retention, we employ a progressive sampling strategy that reduces the sampling rate of earlier-stage data as training advances. Specifically, at stage $k$, the effective training set is constructed as:
\begin{equation}
\label{eq:curriculum_dataset}
\mathcal{D}^{(k)}_{\text{eff}} = \bigcup_{j=1}^{k} \lambda_{k,j} \cdot \mathcal{D}_j
\end{equation}
where $\mathcal{D}_j$ represents the task subset at stage $j$, and $\lambda_{k,j} \in (0,1]$ denotes the sampling rate for stage $j$ data when training at stage $k$. We set $\lambda_{k,k} = 1.0$ to fully utilize current-stage tasks, while applying reduced sampling rates $\lambda_{k,j} < 1.0$ for $j < k$ to maintain earlier capabilities without overwhelming the training distribution. This formulation ensures that the model progressively builds upon foundational skills while adapting to increasingly complex analytical requirements.

Each stage is trained for one complete epoch over its effective dataset $\mathcal{D}^{(k)}_{\text{eff}}$, allowing the model to adequately internalize the task-specific patterns before progressing to the next complexity level. The stage-specific optimization objective is:
\begin{equation}
\label{eq:curriculum_loss}
\mathcal{L}^{(k)}(\Theta) = -\sum_{(\mathbf{I}_i, \mathbf{y}_i) \in \mathcal{D}^{(k)}_{\text{eff}}} \sum_{j=1}^{L_i} \log p_\Theta\left(y_{i,j} \mid \mathbf{I}_i, y_{i,<j}\right)
\end{equation}
The model parameters $\Theta$ are continuously updated across all three stages, with the final trained model encapsulating knowledge from the complete hierarchical training progression. To ensure fair comparison with baseline models, we constrain the sample size at each stage to match that of the baselines, thereby maintaining equivalence in the total number of training samples observed across all methods.

\subsection{CholeVidQA-32K Dataset Curation}
\subsubsection{Overall Curation Pipeline}
As illustrated in Figure~\ref{fig:curation_pipeline}, our dataset construction follows a systematic four-stage pipeline designed to ensure both scalability and clinical accuracy. The first stage encompasses video collection and segmentation, wherein surgical video clips are extracted from multiple sources and partitioned according to their distinct characteristics and annotation types. The second stage involves comprehensive information extraction, wherein diverse annotations—including temporal keyframes, binary classification labels, continuous scoring metrics, spatial bounding boxes, and expert-generated textual captions—are systematically aggregated, processed, and standardized into a unified format. In the third stage, we employ carefully engineered prompts to guide GPT-5.1 in generating open-ended question-answer pairs from the extracted multimodal annotations. To address potential hallucination artifacts in large language model generation, we implement a stratified sampling protocol across each VQA category, resulting in 1\% of the dataset (320 QA pairs), and engage one clinical expert in a structured review process to assess the generated content across four critical dimensions: (1) \textit{Clinical Accuracy}, verifying factual correctness of medical terminology and procedural descriptions; (2) \textit{Question Clarity}, confirming questions are unambiguous and clinically meaningful; (3) \textit{Language Quality}, evaluating linguistic coherence and appropriate use of medical terminology; and (4) \textit{Annotation Consistency}, checking alignment with existing clinical annotations and assessment frameworks. Expert feedback identifying deficiencies across these dimensions is subsequently formalized into refinement prompts that guide iterative improvement of the generated QA-pairs. \rev{The failure modes surfaced by this review, the root cause attributed to each of them, the corrective action applied in the next iteration, and before/after examples of the affected QA pairs are reported in Appendix~\ref{appendix:feedback_loop}.} Finally, the fourth stage encompasses rigorous quality assurance, wherein all test set samples undergo manual verification to eliminate hallucinations and ensure clinical validity.

\subsubsection{Detailed Curation Process}
\label{subsubsec:detailed_curation_process}
To construct a dataset with rich hierarchical semantics, we adapt data from multiple established sources and partition them into three distinct subsets, designated with the VQA suffix: CholecT50-Caption-VQA, Endoscapes-VQA, and CholeScore-VQA. \rev{The 11 task categories were selected to follow the clinical progression from localized scene perception, to established intraoperative assessment constructs, to higher-level reasoning and planning; consequently, the temporal extent of each video segment is determined by the annotation unit needed to support that task rather than by a fixed arbitrary clip length.} As illustrated in Figure~\ref{fig:curation_pipeline}, the detailed curation process for each subset is described below:

\rev{The newly introduced annotations in this work are the clinician-written captions for CholecT50-Caption-VQA and the clinician-authored answers in the expert-verified test subset.}

\textbf{CholecT50-Caption-VQA}: 
We begin by leveraging videos from the publicly available CholecT50 dataset~\citep{Nwoye_2022}, which provides laparoscopic surgical videos annotated with action triplets at one-second intervals. Each temporal frame contains annotations indicating the presence or absence of specific action triplets, with multiple concurrent actions possible within individual frames. We segment videos based on action consistency, grouping consecutive frames where the same set of actions remains unchanged. \rev{This yields short clips because tool, action, anatomy, and action-rationale questions are grounded in visually coherent moments where the operative action remains stable.} Subsequently, we engage clinical experts to annotate each video segment following a structured annotation framework encompassing anatomical descriptions, tool dynamics, and surgical action rationales, yielding over 1,500 caption-clip pairs. Based on these rich textual descriptions, we employ GPT-5.1 to generate three distinct categories of question-answer pairs: (1) \textit{Conversational QA}, which isolates specific informational aspects from the captions including tools, anatomy, and actions, yielding the \textit{Tool Perception} (TP), \textit{Action Perception} (AP), and \textit{Anatomical Structure Perception} (ASP) tasks at the perception level; (2) \textit{Descriptive QA}, which synthesizes multiple aspects to provide comprehensive surgical scene elaboration, corresponding to the \textit{Surgical Scene Description} (SSD) task; and (3) \textit{Reasoning QA}, which poses questions conditioned on the rationale of current operative actions, corresponding to the \textit{Action Rationale Reasoning} (ARR) task (e.g., ``Why does the surgeon remain stationary during this phase rather than immediately performing dissection or manipulation?''). The detailed prompt engineering strategy for this subset is provided in Appendix~\ref{appendix:cholec_prompt}.

\textbf{Endoscapes-VQA}:
To incorporate Critical View of Safety (CVS)—one of the most critical surgical safety assessment protocols—we adapt the Endoscapes dataset~\citep{Mascagni2025}, which comprises 201 laparoscopic cholecystectomy videos with systematically annotated frames featuring CVS labels and spatial bounding boxes delineating critical anatomical structures. Each frame annotation consists of binary labels corresponding to the three established CVS criteria: the Two Structure criterion, the Hepatocystic triangle criterion, and the Cystic plate criterion. Each criterion focuses on the clear visual identification and adequate dissection of specific anatomical landmarks essential for safe surgical practice. Our objective extends beyond binary CVS prediction to enable interpretable reasoning that justifies the assessment decisions. To accomplish this, we implement temporal segmentation using fixed 5-second intervals that align with the original CVS annotation frequency, selecting the central frame of each segment as the representative keyframe. \rev{The 5-second window is therefore inherited from the CVS supervision granularity and provides enough local context for anatomical-relationship assessment without diluting the frame-level label.} We enhance these keyframes by overlaying the corresponding bounding box annotations to provide explicit visual guidance for anatomical structure localization. To ensure comprehensive understanding of CVS assessment rationale, we incorporate the complete CVS annotation guidelines into our generation prompts, enabling GPT-5.1 to deduce reasoning based on visual evidence from the annotated keyframes—including anatomical structure presence and spatial relationships—in conjunction with the detailed procedural guidelines that encompass various decision scenarios for CVS achievement assessment. This approach yields three comprehensive question-answer pairs corresponding to each CVS criterion, characterized by open-ended formulations. The detailed prompt engineering approach for this subset is provided in Appendix~\ref{appendix:endoscapes_prompt}.

\renewcommand{\arraystretch}{1.3}
\begin{table*}[t!]
\centering
\footnotesize
\begin{tabular}{l>{\centering\arraybackslash}m{0.8cm} >{\centering\arraybackslash}m{1.3cm} >{\centering\arraybackslash}m{0.9cm} >{\centering\arraybackslash}m{0.9cm} >{\centering\arraybackslash}m{0.9cm} >{\centering\arraybackslash}m{0.9cm} >{\centering\arraybackslash}m{1.4cm} >{\centering\arraybackslash}m{1.4cm} >{\centering\arraybackslash}m{2.3cm}}
\hline
\makecell{Dataset Subset} & \makecell{\#Videos} & \makecell{Avg Length\\(sec)} & \makecell{\#Conv.\\QA} & \makecell{\#Desc.\\QA} & \makecell{\#Reas.\\QA} & \makecell{\#Total\\QA} & \makecell{Avg WC\\Questions} & \makecell{Avg WC\\Answers} & \makecell{Tasks} \\
\hline
CholecT50-Caption-VQA & 1544 & 3 ± 4 & 9550 & 1544 & 7274 & 18368 & 14 ± 4 & 48 ± 45 & TP, AP, ASP, SSD, ARR \\
Endoscapes-VQA & 1812 & 5 ± 0 & 5436 & - & - & 5436 & 22 ± 7 & 54 ± 11 & CVSA \\
CholeScore-VQA & 499 & 711 ± 818 & 6405 & 1051 & 1051 & 8507 & 18 ± 9 & 64 ± 70 & DFA, SPA, AEA, CA, IP \\
\hline
\textbf{Total} & \textbf{3.9K} & \textbf{95 ± 378} & \textbf{21.4K} & \textbf{2.6K} & \textbf{8.3K} & \textbf{32.3K} & \textbf{17 ± 7} & \textbf{53 ± 50} & - \\
\hline
\end{tabular}
\caption{Comprehensive statistics of CholeVidQA-32K subsets covering number of videos, average video length, number of conversational (Conv.), descriptive (Desc.), and reasoning (Reas.) QA pairs, total QA pairs, and average word count (WC) for questions and answers, with corresponding task abbreviations (see Subsubsection~\ref{subsubsec:detailed_curation_process}).}\label{tab:dataset_stats}
\end{table*}

\textbf{CholeScore-VQA}:
To encompass comprehensive intraoperative assessment dimensions beyond CVS, including intraoperative difficulty evaluation, adverse event detection, and skill assessment, we introduce the CholeScore-VQA subset comprising 100 laparoscopic cholecystectomy videos from the CholeScore dataset~\citep{Sharma_2025}. Based on the nature of the annotations being phase-focused, we implement phase-based temporal segmentation, dividing procedures according to the six established surgical phases (e.g., trocar placement, hepatocystic triangle dissection), with individual procedures typically spanning 1-2 hours and resulting in extended video segments ranging from minutes to tens of minutes in duration. \rev{These longer segments are necessary because difficulty, adverse-event, and skill-proficiency judgments often depend on accumulated evidence across a procedural phase, such as tissue handling, exposure quality, bleeding control, and completion of phase-specific operative goals.} This subset incorporates three distinct categories of annotations:\\
(1) \textit{Operative Difficulty Assessment}, integrating three established difficulty scoring systems—the Nassar, Sugrue, and Parkland Grading Scale (PGS)—consolidating their evaluation criteria into discrete clinical findings represented as binary labels indicating the presence or absence of specific conditions (e.g., visceral fat, dense adhesions);\\
(2) \textit{Intraoperative Adverse Events (IAE)}, employing the SEVERE classification system to systematically catalog adverse events including thermal injury, bleeding, and mechanical complications, with each event categorized according to severity levels ranging from 3 to 5 gradations;\\
(3) \textit{Skill Assessment}, incorporating the Objective Structured Assessment of Technical Skills (OSATS) framework, which provides structured evaluation of procedural competency in critical anatomical manipulations (e.g., gallbladder fossa dissection) and operative goal completion (e.g., access port placement), with scores reflecting varying levels of surgical proficiency.\\
This comprehensive annotation set yields over 80 distinct assessment labels distributed across 499 video clips, with each surgical phase featuring phase-specific annotation sets that capture the unique safety considerations and technical challenges inherent to that procedural stage.\\
Following the consolidation of these multi-dimensional annotations, we formulate the task as phase-level video question answering. For each video segment, we provide GPT-5.1 with the complete set of phase-specific annotation labels accompanied by detailed descriptions of each assessment criterion. Consistent with the CholecT50-Caption-VQA methodology, we generate three distinct categories of question-answer pairs: (1) \textit{Conversational QA}, which isolates individual annotation labels to formulate targeted questions and derives clinical reasoning based on the provided label descriptions, yielding the \textit{Difficulty Findings Assessment} (DFA), \textit{Adverse Events Assessment} (AEA), and \textit{Skills Proficiency Assessment} (SPA) tasks; (2) \textit{Descriptive QA}, which synthesizes all available labels to generate comprehensive safety assessments that holistically evaluate the procedural segment, corresponding to the \textit{Comprehensive Assessment} (CA) task; and (3) \textit{Reasoning QA}, which poses hypothetical scenario-based questions by modifying specific clinical findings (e.g., ``Given dense fibrotic adhesions, how should the surgeon optimize port placement?''), strictly conditioned on the clinical definitions of each finding—including how it affects LCOD severity and what further intraoperative interventions it may require—corresponding to the \textit{Intraoperative Planning} (IP) task. This subset is didactic in nature, designed to build procedural reasoning rather than perform actual assessment, and is validated by checking whether model responses adhere to the original clinical descriptions. The detailed prompt engineering methodology for this subset is provided in Appendix~\ref{appendix:lcod_prompt}.

\begin{figure*}[htbp]
    \centering
    \includegraphics[width=1.0\textwidth]{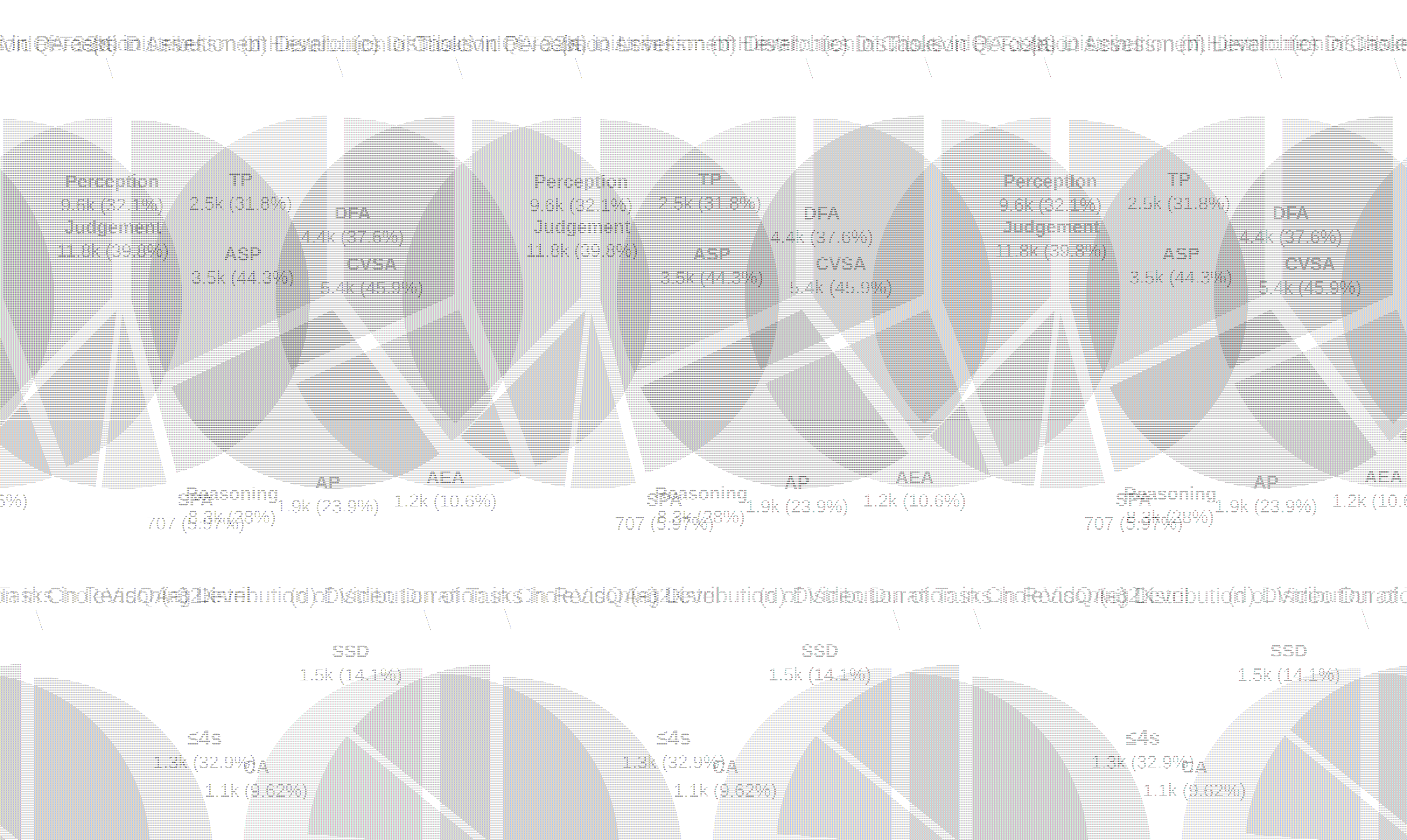}
    \caption{Breakdown of CholeVidQA-32K composition and characteristics across hierarchy, tasks, and temporal duration. Task abbreviations: TP (Tool Perception), AP (Action Perception), ASP (Anatomical Structure Perception), CVSA (Critical View of Safety Assessment), DFA (Difficulty Findings Assessment), AEA (Adverse Events Assessment), SPA (Skills Proficiency Assessment), SSD (Surgical Scene Description), ARR (Action Rationale Reasoning), CA (Comprehensive Assessment), and IP (Intraoperative Planning).}
    \label{fig:dataset_piecharts}
\end{figure*}

The detailed statistics of the resulting dataset are provided in Table~\ref{tab:dataset_stats}, and a thorough breakdown of the three hierarchies, along with further breakdowns of each hierarchy and temporal duration, can be found in Figure~\ref{fig:dataset_piecharts}.

\subsubsection{\rev{Expert-Verified Test Subset Curation}}
To explore the alignment between our proposed model and real-world clinicians' reasoning processes during operations, we extract a subset of samples from the assessment level and engage clinical experts to manually create answers based on given videos and questions. \rev{We refer to this set as the \emph{expert-verified test subset} rather than as a golden standard, since at $300$ QA pairs over $62$ videos it is large enough to compare models on clinician-written answers but not large enough to validate the full CholeVidQA-32K pipeline; its remaining scale limitation is discussed in Section~\ref{sec:limitations}.} The resulting statistics are shown in Table~\ref{tab:golden_set_stats}.

\begin{table}[htbp]
\centering
\revtable
\footnotesize
\begin{tabular}{lccccc}
\hline
\textbf{Statistic} & \textbf{CVS} & \textbf{LCOD} & \textbf{IAE} & \textbf{Skill} & \textbf{Total} \\
\hline
\textbf{\#QA Pairs} & 75 & 151 & 37 & 37 & 300 \\
\textbf{\#Videos} & 25 & 6 & 14 & 17 & 62 \\
\hline
\end{tabular}
\caption{\rev{Statistics of the expert-verified test subset with clinician-authored answers across different assessment subtasks.}}
\label{tab:golden_set_stats}
\end{table}

\section{Experiment }
% 3 to 8 pages for experiment

\subsection{Evaluation Pipeline}
\begin{figure}[htbp]
    \centering
    \includegraphics[width=\columnwidth]{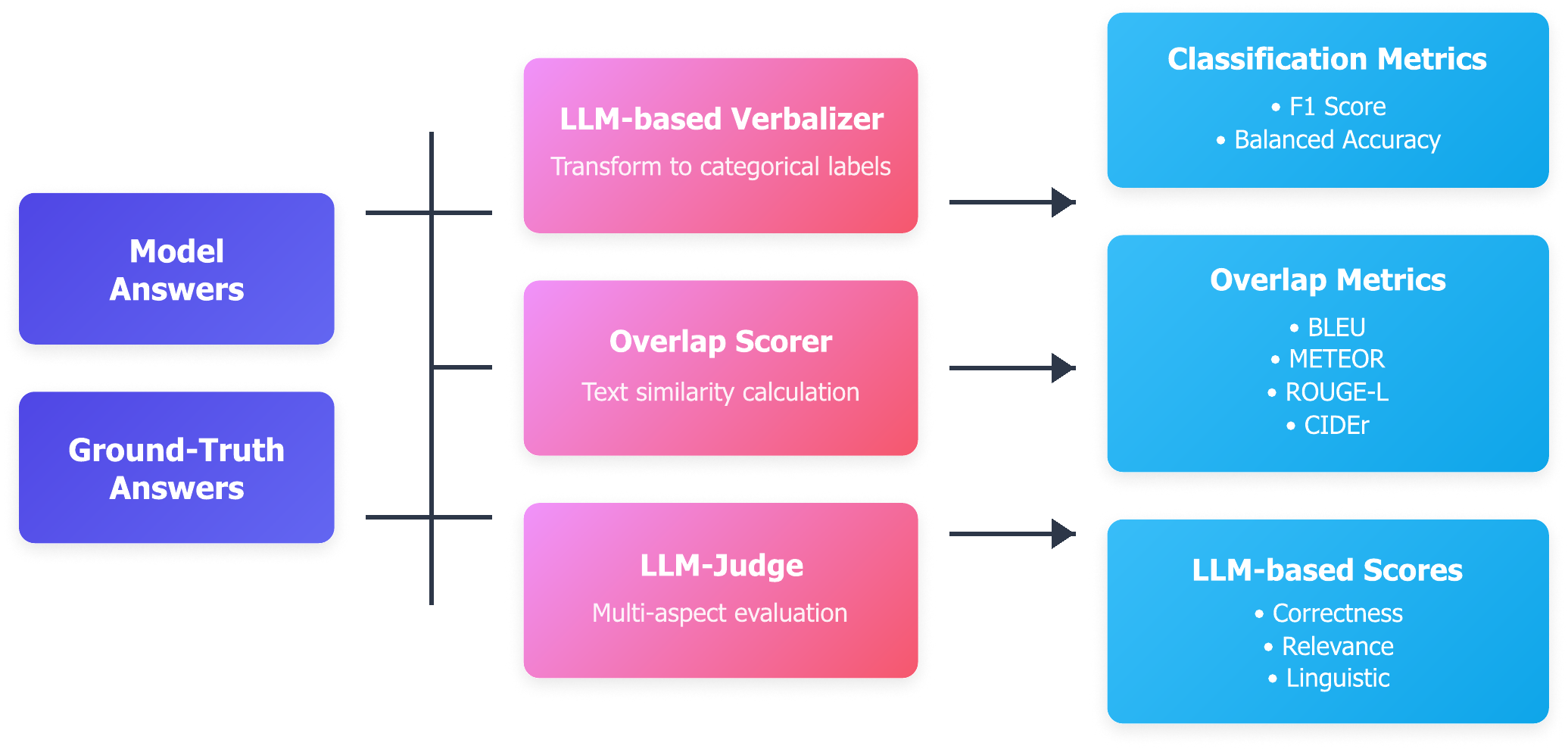}
    \caption{Comprehensive evaluation pipeline for CholeVidQA-32K assessment framework incorporating three complementary evaluation methodologies for robust performance measurement.}
    \label{fig:eval_pipeline}
\end{figure}

Given the complex nature of surgical video question answering tasks that encompass both categorical intraoperative assessments and open-ended clinical reasoning, we establish a comprehensive three-tier evaluation framework as illustrated in Figure~\ref{fig:eval_pipeline}. This multi-faceted approach addresses the inherent challenges in evaluating natural language responses for safety-critical surgical applications, where both factual accuracy and linguistic quality are paramount.

The evaluation pipeline processes model-generated answers and ground-truth answers through three distinct but complementary assessment streams: \textbf{(1) LLM-based Verbalizer} transforms both predicted and reference answers into standardized categorical labels, enabling systematic classification performance measurement through F1 scores (for DFA, SPA and AEA tasks) and balanced accuracy metrics (for CVSA task) that account for potential class imbalances inherent in clinical datasets. To reflect the model's capability in providing evaluable results, we also report answer rate along with the scores. \textbf{(2) Overlap Scorer} conducts direct text similarity analysis between generated responses and reference responses using established natural language generation metrics including BLEU for n-gram precision, METEOR for semantic alignment with synonyms and paraphrases, ROUGE-L for longest common subsequence matching, and CIDEr for consensus-based evaluation that emphasizes clinically relevant terminology. \textbf{(3) LLM-Judge} performs sophisticated multi-aspect evaluation wherein large language models assess generated responses across three critical dimensions: correctness (clinical accuracy and factual validity), relevance (appropriateness to the surgical context and question intent), and linguistic quality (coherence, clarity, and professional surgical terminology usage).

\subsection{Dataset Split}
We employ video-level stratified splitting strategies to ensure temporal independence and prevent data leakage across training, validation, and test sets. For \textbf{CholecT50-Caption-VQA}, we implement a 7:1:2 split ratio that balances training data sufficiency with robust evaluation set sizes. For \textbf{Endoscapes-VQA} and \textbf{CholeScore-VQA}, we adopt the original partitioning schemes from~\citep{Mascagni2025} and~\citep{Sharma_2025} respectively to maintain consistency with prior assessment benchmarks. Table~\ref{tab:dataset_split} summarizes the detailed statistics across all subsets.

\begin{table}[h!]
\centering
\scriptsize
\begin{tabular}{lccc|ccc}
\hline
\textbf{Subset} & \multicolumn{3}{c|}{\textbf{\#Videos}} & \multicolumn{3}{c}{\textbf{\#QA Pairs}} \\
\cline{2-7}
 & Train & Val & Test & Train & Val & Test \\
\hline
CholecT50-Caption-VQA & 1.1K & 155 & 309 & 12.9K & 1.8K & 3.7K \\
Endoscapes-VQA & 1.2K & 367 & 289 & 3.5K & 1.1K & 867 \\
CholeScore-VQA & 259 & 50 & 190 & 4.4K & 860 & 3.2K \\
\hline
\end{tabular}
\caption{Dataset split statistics across CholeVidQA-32K subsets.}
\label{tab:dataset_split}
\end{table}

\subsection{Baseline Models}
We evaluate the following open-source zero-shot models on our dataset: mPLUG-Owl3~\citep{ye2024mplug}, InternVideo2.5~\citep{wang2025internvideo2}, LongVA~\citep{zhang2024long}, LLaVA-Video~\citep{zhang2024video}, VideoGPT+~\citep{maaz2024videogptintegratingimagevideo}\rev{, and Qwen3-VL~\citep{bai2025qwen3vltechnicalreport}}. \rev{We additionally evaluate one proprietary zero-shot model, Gemini-3.1-Pro~\citep{googledeepmind2026gemini31pro}, accessed through its public video-capable API.} To isolate the effectiveness of our proposed model and ensure fair comparison while avoiding confounding effects from domain adaptation, we fine-tune LLaVA-Video\rev{,} VideoGPT+ \rev{and Qwen3-VL} on our CholeVidQA-32K dataset, creating the LLaVA-Video-ft\rev{,} VideoGPT+-ft \rev{and Qwen3-VL-ft} baselines.

\rev{These baselines are selected under three criteria rather than by leaderboard standing. First, they must be video-capable and accept the variable-length clips of CholeVidQA-32K under comparable input settings, so that differences in score reflect modelling rather than differences in how the video is presented to the model. Second, they must be reproducible: open weights that can be run, and where appropriate fine-tuned, under a single controlled protocol, which is what allows the fine-tuned comparisons in this work to isolate the contribution of the TEMP module rather than the effect of an unknown proprietary training recipe. Third, they must be sufficiently recent to serve as meaningful references, which is why we additionally include Qwen3-VL in both zero-shot and fine-tuned form and Gemini-3.1-Pro as a proprietary zero-shot reference point. General-purpose frontier multimodal models continue to be released at a rapid pace, and the intent of this comparison is not to track that leaderboard, but to measure what surgical-domain adaptation and query-conditioned temporal selection contribute under a fixed and reproducible protocol.} \rev{All fine-tuned conditions use the identical CholeVidQA-32K train/test split, question--answer pairs, task-balanced sampling strategy, and surgical instruction template, including the same system prompt, task framing, role tokens, and answer-formatting conventions. Thus, no baseline receives additional surgical supervision or model-specific prompt reformulation.}

For all zero-shot models, we standardize the input to 64 uniformly sampled frames per video to ensure consistent temporal coverage and fair comparison. For fine-tuned models, memory constraints necessitate different configurations: LLaVA-Video-ft\rev{,} VideoGPT+-ft \rev{and Qwen3-VL-ft process} 10 frames per video during training, and 64 frames are fed into the model during inference. Our proposed SurgTEMP leverages the TEMP module's text-guided attention mechanism to achieve efficient visual token reduction, enabling processing of 64 frames during both training and inference stage. \rev{All fine-tuned conditions therefore observe the same 64-frame candidate pool at inference time. The TEMP module is configured to retain the top $10$ of those $64$ frames, so SurgTEMP forwards strictly fewer visual tokens to its LLM than the baselines, which consume all $64$ frames directly. The comparison is thus not favourable to SurgTEMP in terms of inference-time visual budget, and any gain must be attributed to denser candidate sampling followed by query-conditioned selection rather than to a larger effective input.}

\subsection{Implementation Details}
We conduct all experiments on a dual-node infrastructure, each node equipped with NVIDIA A100 80GB GPUs. For SurgTEMP, we adopt the SigLIP ViT-SO400M/14-384 as the vision encoder, Qwen2-7B-Instruct as the LLM backbone, and the pretrained projector from~\cite{zhang2024video}. We follow the image preprocessing pipeline from~\citep{zhai2023sigmoid}, resizing input frames to 384$\times$384 pixels and applying normalization using ImageNet statistics. For spatial pooling after image encoding and cross-modal projection, we employ a bilinear pooling strategy with a stride of 2.

To ensure a fair comparison of learnable parameter budgets, we apply a consistent LoRA configuration of $r=64$ and $\alpha=16$ across all fine-tuned models. We employ a discriminative learning rate strategy: the vision encoder is updated with a learning rate of $2 \times 10^{-6}$, while all other trainable parameters use $1 \times 10^{-5}$. Training uses a cosine annealing scheduler with linear warmup over 3\% of total training steps, with mixed-precision training using bfloat16 (bf16) for model weights and TensorFloat-32 (tf32) for matrix multiplications. All models are trained for 1 epoch on the combined training set with an effective batch size of 4 via gradient accumulation, requiring approximately 20 hours for SurgTEMP convergence.

\rev{LoRA dropout is $0.05$, and adapters are applied to the attention and MLP projections of both the LLM and vision encoder in every fine-tuned condition. The multi-modal projector is initialised from pretrained weights but, being lightweight, is fully optimized rather than LoRA-adapted, in every fine-tuned condition. The modules introduced by SurgTEMP---the TEMP constructor and the learnable grid and frame separator tokens---have no pretrained counterpart, are trained from random initialization, and are likewise fully optimized; they are specific to SurgTEMP and are not added to any baseline. We use AdamW with $\beta_1=0.9$, $\beta_2=0.95$, weight decay $0$, and gradient clipping at $1.0$.}

\rev{\textbf{Randomness control and reported variability.} Every entry in Tables~\ref{tab:overall_results},~\ref{tab:subset_results},~\ref{tab:overlap_metrics},~\ref{tab:ablation_study} and~\ref{tab:golden_set_results_additional} is the mean over three independent runs, with the subscript reporting the standard deviation across those runs; the corresponding figures plot the mean over the same runs. Because the dominant source of stochasticity differs across model groups, the quantity resampled by the three runs is matched to what is actually controllable in each case. For the open-source zero-shot models no training is performed, so the only stochasticity is at inference: we decode with sampling at temperature $1$ under three random seeds, and the reported spread therefore characterizes inference variability. For the fine-tuned models, comprising all fine-tuned baselines and SurgTEMP, we train three separate models under three random seeds, which resamples the initialization of the newly introduced modules, the data ordering, and the stochastic Gumbel-Softmax frame selection, and we then decode every trained model greedily at temperature $0$; the reported spread therefore characterizes training variability and is not confounded with sampling noise. For the proprietary zero-shot model, whose weights and training procedure are inaccessible, we issue three independent queries through the public API at temperature $1$ with a fixed thinking budget, so that the spread reflects inference variability at a fixed reasoning-compute setting. Unless stated otherwise, all margins quoted in the following analyses are differences between run means and should be read against the run-to-run standard deviations reported alongside them.}

For evaluation, we use the open-source \texttt{gpt-oss-120b}~\citep{openai2025gptoss120bgptoss20bmodel} as the LLM judge and verbalizer. To avoid potential judge bias on specific styles, we conduct exploration on multiple judge models: proprietary models including gpt-5.1 and gemini-2.5-pro, and open-source models including gpt-oss-120b~\citep{openai2025gptoss120bgptoss20bmodel}, Mixtral-8x7B-Instruct~\citep{jiang2024mixtralexperts}, and DeepSeek-V3~\citep{deepseekai2025deepseekv3technicalreport}. With fixed evaluation settings including evaluated samples (the test set), scoring prompt, and evaluation criteria, we report their agreement on model performance ranking across each hierarchy of the dataset as shown in Table~\ref{tab:kendall_w}. Notably, our model is ranked first overall across all judge models.

\begin{table}[htbp]
\centering
\caption{Kendall's W coefficient across different judge models showing strong agreement on evaluated models' ranking overall and within each hierarchy on our dataset.}
\label{tab:kendall_w}
\begin{tabular}{cccc}
\hline
Overall & Perception & Assessment & Reasoning \\
\hline
0.852 & 0.811 & 0.864 & 0.782 \\
\hline
\end{tabular}
\end{table}

\subsection{Results Analysis}

\begin{table*}[t!]
\centering
\footnotesize
\setlength{\tabcolsep}{2pt}
\begin{tabular}{l|>{\centering\arraybackslash}m{1.11cm} >{\centering\arraybackslash}m{1.11cm} >{\centering\arraybackslash}m{1.11cm}|>{\centering\arraybackslash}m{1.11cm} >{\centering\arraybackslash}m{1.22cm} >{\centering\arraybackslash}m{1.22cm} >{\centering\arraybackslash}m{1.11cm}|>{\centering\arraybackslash}m{1.11cm} >{\centering\arraybackslash}m{1.11cm}|>{\centering\arraybackslash}m{1.11cm} >{\centering\arraybackslash}m{1.11cm}}
\hline
\multirow{3}{*}{Models} & \multicolumn{3}{c|}{GPT Scores} & \multicolumn{4}{c|}{Overlap Metrics} & \multicolumn{4}{c}{Classification Metrics} \\
\cline{2-12}
        & CR & RL & LG & BLEU & METEOR & ROUGE-L & CIDEr & \multicolumn{2}{c|}{bAcc} & \multicolumn{2}{c}{F1-score} \\
\cline{9-12}
        &  &  &  &  &  &  &  & Score & Rate & Score & Rate \\
\hline
\multicolumn{12}{c}{\rule[-1ex]{0pt}{3ex}\textit{Open-source Zero-shot}} \\
mPLUG-Owl3 & 32.05$_{\pm.05}$ & 45.01$_{\pm.11}$ & 43.03$_{\pm.14}$ & \cellcolor{cyan!20}5.30$_{\pm.00}$ & 23.53$_{\pm.03}$ & 22.86$_{\pm.16}$ & 6.35$_{\pm.13}$ & 32.06$_{\pm.11}$ & \cellcolor{cyan!20}95.08$_{\pm.18}$ & 15.05$_{\pm.04}$ & \cellcolor{cyan!20}97.29$_{\pm.18}$ \\
InternVideo2.5 & 16.16$_{\pm.11}$ & 19.50$_{\pm.08}$ & 16.74$_{\pm.06}$ & 2.74$_{\pm.03}$ & 8.95$_{\pm.05}$ & 9.81$_{\pm.17}$ & 4.41$_{\pm.12}$ & 25.82$_{\pm.03}$ & 90.89$_{\pm.03}$ & 1.21$_{\pm.08}$ & 7.18$_{\pm.08}$ \\
LongVA & 25.84$_{\pm.11}$ & 34.80$_{\pm.15}$ & 34.80$_{\pm.09}$ & 2.20$_{\pm.15}$ & 21.56$_{\pm.10}$ & 14.00$_{\pm.00}$ & 0.33$_{\pm.07}$ & 5.23$_{\pm.03}$ & 9.23$_{\pm.07}$ & 22.58$_{\pm.04}$ & 56.05$_{\pm.02}$ \\
LLaVA-Video & 33.17$_{\pm.14}$ & 40.98$_{\pm.07}$ & 38.76$_{\pm.10}$ & 3.85$_{\pm.20}$ & 21.58$_{\pm.09}$ & 17.13$_{\pm.07}$ & 3.72$_{\pm.02}$ & 27.14$_{\pm.15}$ & 48.74$_{\pm.11}$ & 12.34$_{\pm.10}$ & 64.93$_{\pm.15}$ \\
VideoGPT+ & \cellcolor{cyan!20}36.66$_{\pm.11}$ & \cellcolor{cyan!20}48.18$_{\pm.05}$ & \cellcolor{cyan!20}46.56$_{\pm.10}$ & 3.70$_{\pm.03}$ & 21.88$_{\pm.15}$ & \cellcolor{cyan!20}22.95$_{\pm.14}$ & \cellcolor{cyan!20}14.00$_{\pm.11}$ & \cellcolor{cyan!20}40.04$_{\pm.01}$ & 68.05$_{\pm.09}$ & \cellcolor{cyan!20}24.83$_{\pm.03}$ & 77.92$_{\pm.03}$ \\
Qwen3-VL & 34.41$_{\pm.21}$ & 47.29$_{\pm.23}$ & 43.85$_{\pm.35}$ & 2.15$_{\pm.00}$ & \cellcolor{cyan!20}25.46$_{\pm.00}$ & 15.79$_{\pm.00}$ & 1.84$_{\pm.00}$ & 32.15$_{\pm.22}$ & 89.49$_{\pm.21}$ & 23.47$_{\pm1.02}$ & 67.88$_{\pm.08}$ \\
\hline
\multicolumn{12}{c}{\rule[-1ex]{0pt}{3ex}\textit{Proprietary Zero-shot}} \\
Gemini-3.1-Pro & 33.55$_{\pm.10}$ & 51.55$_{\pm.11}$ & 51.94$_{\pm.10}$ & 2.67$_{\pm.18}$ & 22.30$_{\pm.13}$ & 18.58$_{\pm.11}$ & 6.49$_{\pm.20}$ & 16.77$_{\pm.02}$ & 40.04$_{\pm.32}$ & 10.33$_{\pm.02}$ & 33.10$_{\pm.17}$ \\
\hline
\multicolumn{12}{c}{\rule[-1ex]{0pt}{3ex}\textit{Fine-tuned}} \\
VideoGPT+-ft & 64.00$_{\pm.07}$ & 73.60$_{\pm.04}$ & 71.70$_{\pm.06}$ & 14.86$_{\pm.11}$ & 33.51$_{\pm.12}$ & 31.81$_{\pm.04}$ & 42.09$_{\pm.05}$ & 52.11$_{\pm.17}$ & 94.83$_{\pm.14}$ & 49.20$_{\pm.18}$ & 81.70$_{\pm.14}$ \\
LLaVA-Video-ft & 60.07$_{\pm.12}$ & 67.37$_{\pm.09}$ & 65.41$_{\pm.12}$ & 14.52$_{\pm.06}$ & 32.36$_{\pm.15}$ & 31.62$_{\pm.20}$ & 40.83$_{\pm.01}$ & 51.06$_{\pm.16}$ & 87.91$_{\pm.09}$ & 44.40$_{\pm.14}$ & 62.71$_{\pm.18}$ \\
Qwen3-VL-ft & 61.76$_{\pm.09}$ & 70.54$_{\pm.13}$ & 69.11$_{\pm.05}$ & 14.92$_{\pm.16}$ & 34.62$_{\pm.08}$ & 32.38$_{\pm.11}$ & 41.57$_{\pm.03}$ & 52.32$_{\pm.14}$ & 97.86$_{\pm.22}$ & 44.02$_{\pm.07}$ & 90.42$_{\pm.02}$ \\
ours & \cellcolor{orange!20}71.64$_{\pm.07}$ & \cellcolor{orange!20}81.76$_{\pm.13}$ & \cellcolor{orange!20}79.04$_{\pm.13}$ & \cellcolor{orange!20}15.51$_{\pm.08}$ & \cellcolor{orange!20}36.14$_{\pm.05}$ & \cellcolor{orange!20}34.70$_{\pm.11}$ & \cellcolor{orange!20}42.55$_{\pm.11}$ & \cellcolor{orange!20}56.43$_{\pm.11}$ & \cellcolor{orange!20}100.00$_{\pm.00}$ & \cellcolor{orange!20}52.54$_{\pm.12}$ & \cellcolor{orange!20}91.19$_{\pm.10}$ \\
\hline
\end{tabular}
\caption{Overall performance comparison of baseline models and our proposed method on the CholeVidQA-32K dataset. Models are grouped into Open-source Zero-shot, Proprietary Zero-shot and Fine-tuned categories. Metrics are grouped into GPT Scores (CR: Correctness, RL: Relevance, LG: Linguistic Quality), Overlap Metrics (BLEU, METEOR, ROUGE-L, CIDEr), and Classification Metrics (bAcc: balanced Accuracy, F1: F1-score). For classification metrics, Score represents the metric value and Rate represents the answer rate (\%); Gemini-3.1-Pro is scored under an unanswered-as-wrong convention. Blue shading indicates best performance among open-source zero-shot models; orange shading indicates best performance among fine-tuned models. The proprietary zero-shot group contains a single model and is therefore left unshaded.}\label{tab:overall_results}
\end{table*}

\begin{table*}[t!]
\centering
\footnotesize
\setlength{\tabcolsep}{2pt}
\begin{tabular}{l|>{\centering\arraybackslash}m{1.16cm} >{\centering\arraybackslash}m{1.16cm} >{\centering\arraybackslash}m{1.16cm}|>{\centering\arraybackslash}m{1.16cm} >{\centering\arraybackslash}m{1.16cm} >{\centering\arraybackslash}m{1.16cm}|>{\centering\arraybackslash}m{1.16cm} >{\centering\arraybackslash}m{1.16cm} >{\centering\arraybackslash}m{1.16cm}}
\hline
\multirow{2}{*}{Models} & \multicolumn{3}{c|}{CholecT50-Caption-VQA} & \multicolumn{3}{c|}{Endoscapes-VQA} & \multicolumn{3}{c}{CholeScore-VQA} \\
\cline{2-10}
        & CR & RL & LG & CR & RL & LG & CR & RL & LG \\
\hline
\multicolumn{10}{c}{\rule[-1ex]{0pt}{3ex}\textit{Open-source Zero-shot}} \\
mPLUG-Owl3 & 46.15$_{\pm.01}$ & 54.26$_{\pm.19}$ & 47.50$_{\pm.09}$ & 29.04$_{\pm.08}$ & 42.75$_{\pm.02}$ & 45.30$_{\pm.20}$ & 20.97$_{\pm.05}$ & 38.01$_{\pm.11}$ & 36.28$_{\pm.12}$ \\
InternVideo2.5 & 12.54$_{\pm.16}$ & 14.50$_{\pm.01}$ & 12.65$_{\pm.12}$ & 34.25$_{\pm.06}$ & 41.38$_{\pm.11}$ & 34.97$_{\pm.01}$ & 1.69$_{\pm.12}$ & 2.62$_{\pm.12}$ & 2.59$_{\pm.04}$ \\
LongVA & 47.73$_{\pm.18}$ & 55.71$_{\pm.13}$ & 51.59$_{\pm.14}$ & 5.24$_{\pm.12}$ & 9.74$_{\pm.13}$ & 14.04$_{\pm.06}$ & 24.55$_{\pm.02}$ & 38.94$_{\pm.19}$ & 38.77$_{\pm.06}$ \\
LLaVA-Video & \cellcolor{cyan!20}50.79$_{\pm.17}$ & \cellcolor{cyan!20}56.68$_{\pm.15}$ & \cellcolor{cyan!20}52.54$_{\pm.12}$ & 30.12$_{\pm.08}$ & 41.82$_{\pm.03}$ & 41.47$_{\pm.03}$ & 18.60$_{\pm.16}$ & 24.43$_{\pm.02}$ & 22.27$_{\pm.16}$ \\
VideoGPT+ & 45.99$_{\pm.17}$ & 53.69$_{\pm.01}$ & 46.67$_{\pm.19}$ & \cellcolor{cyan!20}38.75$_{\pm.13}$ & \cellcolor{cyan!20}50.17$_{\pm.12}$ & \cellcolor{cyan!20}53.74$_{\pm.01}$ & \cellcolor{cyan!20}25.25$_{\pm.02}$ & \cellcolor{cyan!20}40.69$_{\pm.01}$ & \cellcolor{cyan!20}39.26$_{\pm.11}$ \\
Qwen3-VL & 42.41$_{\pm.09}$ & 56.39$_{\pm.15}$ & 48.51$_{\pm.40}$ & 36.13$_{\pm.50}$ & 45.65$_{\pm.37}$ & 44.39$_{\pm.32}$ & 24.70$_{\pm.04}$ & 39.82$_{\pm.16}$ & 38.66$_{\pm.34}$ \\
\hline
\multicolumn{10}{c}{\rule[-1ex]{0pt}{3ex}\textit{Proprietary Zero-shot}} \\
Gemini-3.1-Pro & 36.12$_{\pm.09}$ & 46.04$_{\pm.13}$ & 44.14$_{\pm.15}$ & 39.83$_{\pm.13}$ & 70.76$_{\pm.12}$ & 73.67$_{\pm.00}$ & 25.14$_{\pm.19}$ & 37.94$_{\pm.16}$ & 37.96$_{\pm.19}$ \\
\hline
\multicolumn{10}{c}{\rule[-1ex]{0pt}{3ex}\textit{Fine-tuned}} \\
VideoGPT+-ft & 72.29$_{\pm.02}$	&77.40$_{\pm.09}$	&74.75$_{\pm.00}$ & 71.36$_{\pm.10}$	&86.90$_{\pm.02}$	&88.46$_{\pm.00}$ & 48.35$_{\pm.08}$	&56.51$_{\pm.01}$	&51.90$_{\pm.18}$ \\
LLaVA-Video-ft & 71.78$_{\pm.15}$ & 76.44$_{\pm.10}$ & 72.60$_{\pm.18}$ & 66.44$_{\pm.15}$ & 78.50$_{\pm.02}$ & 79.47$_{\pm.10}$ & 41.98$_{\pm.07}$ & 47.17$_{\pm.16}$ & 44.17$_{\pm.09}$ \\
Qwen3-VL-ft & 70.02$_{\pm.12}$ & 76.22$_{\pm.04}$ & 73.23$_{\pm.17}$ & 65.86$_{\pm.06}$ & 75.71$_{\pm.15}$ & 74.59$_{\pm.02}$ & 49.40$_{\pm.10}$ & 59.69$_{\pm.18}$ & 59.51$_{\pm.08}$ \\
ours & \cellcolor{orange!20}74.92$_{\pm.14}$ & \cellcolor{orange!20}81.53$_{\pm.19}$ & \cellcolor{orange!20}78.63$_{\pm.18}$ & \cellcolor{orange!20}73.71$_{\pm.01}$ & \cellcolor{orange!20}89.62$_{\pm.13}$ & \cellcolor{orange!20}90.12$_{\pm.03}$ & \cellcolor{orange!20}66.29$_{\pm.06}$ & \cellcolor{orange!20}74.14$_{\pm.07}$ & \cellcolor{orange!20}68.36$_{\pm.19}$ \\
\hline
\end{tabular}
\caption{GPT Score performance breakdown across the three CholeVidQA-32K subsets: CholecT50-Caption-VQA, Endoscapes-VQA, and CholeScore-VQA. All metrics are GPT-based evaluations with CR: Correctness, RL: Relevance, LG: Linguistic Quality. Blue shading indicates best performance among open-source zero-shot models; orange shading indicates best performance among fine-tuned models. The proprietary zero-shot group contains a single model and is therefore left unshaded.}\label{tab:subset_results}
\end{table*}

\begin{table*}[t!]
\centering
\footnotesize
\setlength{\tabcolsep}{1.5pt}
\begin{tabular}{l|>{\centering\arraybackslash}m{1.00cm} >{\centering\arraybackslash}m{1.11cm} >{\centering\arraybackslash}m{1.11cm} >{\centering\arraybackslash}m{1.00cm}|>{\centering\arraybackslash}m{1.00cm} >{\centering\arraybackslash}m{1.11cm} >{\centering\arraybackslash}m{1.11cm} >{\centering\arraybackslash}m{1.00cm}|>{\centering\arraybackslash}m{1.00cm} >{\centering\arraybackslash}m{1.11cm} >{\centering\arraybackslash}m{1.11cm} >{\centering\arraybackslash}m{1.00cm}}
\hline
\multirow{2}{*}{Models} & \multicolumn{4}{c|}{CholecT50-Caption-VQA} & \multicolumn{4}{c|}{Endoscapes-VQA} & \multicolumn{4}{c}{CholeScore-VQA} \\
\cline{2-13}
        & BLEU & METEOR & ROUGE-L & CIDEr & BLEU & METEOR & ROUGE-L & CIDEr & BLEU & METEOR & ROUGE-L & CIDEr \\
\hline
\multicolumn{13}{c}{\rule[-1ex]{0pt}{3ex}\textit{Open-source Zero-shot}} \\
mPLUG-Owl3 & 3.63$_{\pm.02}$ & 26.30$_{\pm.07}$ & 23.33$_{\pm.00}$ & 8.37$_{\pm.19}$ & \cellcolor{cyan!20}13.05$_{\pm.19}$ & 30.23$_{\pm.09}$ & 30.93$_{\pm.11}$ & 8.28$_{\pm.15}$ & 0.18$_{\pm.16}$ & 14.56$_{\pm.10}$ & 14.69$_{\pm.17}$ & 2.18$_{\pm.08}$ \\
InternVideo2.5 & 1.47$_{\pm.18}$ & 7.26$_{\pm.03}$ & 6.91$_{\pm.18}$ & 10.86$_{\pm.12}$ & 6.80$_{\pm.14}$ & 18.52$_{\pm.15}$ & 21.47$_{\pm.06}$ & 3.13$_{\pm.17}$ & 0.00$_{\pm.15}$ & 0.69$_{\pm.10}$ & 0.43$_{\pm.14}$ & 0.00$_{\pm.12}$ \\
LongVA & 1.46$_{\pm.08}$ & 27.31$_{\pm.08}$ & 15.82$_{\pm.13}$ & 0.17$_{\pm.19}$ & 3.83$_{\pm.19}$ & 19.15$_{\pm.02}$ & 15.92$_{\pm.01}$ & 1.70$_{\pm.16}$ & 0.36$_{\pm.03}$ & 18.66$_{\pm.13}$ & 10.64$_{\pm.00}$ & 0.25$_{\pm.19}$ \\
LLaVA-Video & 2.93$_{\pm.13}$ & 26.39$_{\pm.05}$ & 20.71$_{\pm.06}$ & 8.13$_{\pm.15}$ & 7.80$_{\pm.10}$ & 27.54$_{\pm.13}$ & 22.43$_{\pm.14}$ & 3.39$_{\pm.07}$ & 0.06$_{\pm.10}$ & 10.10$_{\pm.02}$ & 7.94$_{\pm.02}$ & 0.31$_{\pm.18}$ \\
VideoGPT+ & \cellcolor{cyan!20}5.09$_{\pm.02}$ & 26.97$_{\pm.19}$ & \cellcolor{cyan!20}27.07$_{\pm.06}$ & \cellcolor{cyan!20}21.20$_{\pm.05}$ & 12.90$_{\pm.00}$ & 29.41$_{\pm.00}$ & \cellcolor{cyan!20}33.07$_{\pm.09}$ & \cellcolor{cyan!20}9.04$_{\pm.04}$ & 0.00$_{\pm.19}$ & 13.89$_{\pm.11}$ & 14.73$_{\pm.01}$ & \cellcolor{cyan!20}2.33$_{\pm.06}$ \\
Qwen3-VL & 2.75$_{\pm.00}$ & \cellcolor{cyan!20}29.19$_{\pm.00}$ & 17.79$_{\pm.00}$ & 3.01$_{\pm.00}$ & 6.47$_{\pm.00}$ & \cellcolor{cyan!20}33.10$_{\pm.00}$ & 20.75$_{\pm.00}$ & 0.73$_{\pm.00}$ & \cellcolor{cyan!20}0.31$_{\pm.00}$ & \cellcolor{cyan!20}19.14$_{\pm.00}$ & 12.18$_{\pm.00}$ & 0.39$_{\pm.00}$ \\
\hline
\multicolumn{13}{c}{\rule[-1ex]{0pt}{3ex}\textit{Proprietary Zero-shot}} \\
Gemini-3.1-Pro & 3.19$_{\pm.16}$ & 26.15$_{\pm.07}$ & 22.46$_{\pm.00}$ & 9.34$_{\pm.13}$ & 10.33$_{\pm.14}$ & 33.67$_{\pm.13}$ & 27.03$_{\pm.10}$ & 4.64$_{\pm.06}$ & 0.03$_{\pm.19}$ & 14.65$_{\pm.16}$ & 12.37$_{\pm.16}$ & 1.50$_{\pm.18}$ \\
\hline
\multicolumn{13}{c}{\rule[-1ex]{0pt}{3ex}\textit{Fine-tuned}} \\
VideoGPT+-ft & 16.00$_{\pm.15}$ & 39.26$_{\pm.07}$ & \cellcolor{orange!20}38.49$_{\pm.14}$ & 77.27$_{\pm.06}$ & 24.39$_{\pm.13}$ & 42.57$_{\pm.05}$ & 41.42$_{\pm.12}$ & 34.36$_{\pm.01}$ & 3.33$_{\pm.13}$ & 18.40$_{\pm.13}$ & 16.17$_{\pm.16}$ & \cellcolor{orange!20}15.38$_{\pm.08}$ \\
LLaVA-Video-ft & 15.81$_{\pm.19}$ & 38.98$_{\pm.06}$ & 37.75$_{\pm.13}$ & 76.98$_{\pm.05}$ & 24.52$_{\pm.07}$ & 41.82$_{\pm.20}$ & 41.24$_{\pm.10}$ & 33.23$_{\pm.01}$ & 3.69$_{\pm.17}$ & 16.42$_{\pm.10}$ & 14.93$_{\pm.09}$ & 12.52$_{\pm.12}$ \\
Qwen3-VL-ft & 16.11$_{\pm.05}$ & 39.85$_{\pm.13}$ & 38.26$_{\pm.01}$ & 77.13$_{\pm.19}$ & 24.88$_{\pm.11}$ & 43.65$_{\pm.07}$ & 42.72$_{\pm.16}$ & 34.16$_{\pm.04}$ & 3.77$_{\pm.09}$ & 20.35$_{\pm.14}$ & 16.16$_{\pm.02}$ & 13.41$_{\pm.12}$ \\
ours & \cellcolor{orange!20}16.26$_{\pm.19}$ & \cellcolor{orange!20}40.22$_{\pm.05}$ & 38.44$_{\pm.07}$ & \cellcolor{orange!20}77.49$_{\pm.05}$ & \cellcolor{orange!20}25.71$_{\pm.17}$ & \cellcolor{orange!20}45.47$_{\pm.10}$ & \cellcolor{orange!20}44.77$_{\pm.08}$ & \cellcolor{orange!20}34.97$_{\pm.20}$ & \cellcolor{orange!20}3.86$_{\pm.09}$ & \cellcolor{orange!20}23.04$_{\pm.08}$ & \cellcolor{orange!20}21.73$_{\pm.08}$ & 15.34$_{\pm.01}$ \\
\hline
\end{tabular}
\caption{Overlap Metrics performance breakdown across the three CholeVidQA-32K subsets: CholecT50-Caption-VQA, Endoscapes-VQA, and CholeScore-VQA. All metrics measure text similarity between generated and reference answers. Blue shading indicates best performance among open-source zero-shot models; orange shading indicates best performance among fine-tuned models. The proprietary zero-shot group contains a single model and is therefore left unshaded.}\label{tab:overlap_metrics}
\end{table*}

\subsubsection{Overall Performance and Domain Adaptation}
Table~\ref{tab:overall_results} presents the averaged performance across all subsets, evaluated using three complementary metric categories: GPT-based scores, overlap metrics, and classification metrics. 

Among open-source zero-shot models, VideoGPT+ emerges as the strongest baseline, achieving \rev{36.66\% correctness—a 2.25 percentage point margin over the second-best Qwen3-VL (34.41\%)}.

For classification metrics, mPLUG-Owl3 has relatively higher answer rates (with 95\% for bAcc, and 97\% for F1 score), yet struggles to produce accurate predictions. This discrepancy underscores the difficulty of surgical assessment tasks, including CVS criterion evaluation, difficulty findings, adverse event detection, and skill scoring, where high answer rates do not guarantee clinical validity. The gap between answer rate and accuracy reveals that zero-shot models often generate responses that superficially appear valid but fail to correctly apply domain-specific assessment frameworks.

For fine-tuned models, our SurgTEMP substantially outperforms \rev{the LLaVA-Video-ft, VideoGPT+-ft and Qwen3-VL-ft baselines} across all three GPT-based dimensions and all overlap metrics. Notably, our model achieves high answer rates for classification metrics, indicating that the model generates more accurate and confident responses.

\subsubsection{Subset-Specific Performance Patterns}
Tables~\ref{tab:subset_results} and~\ref{tab:overlap_metrics} reveal model performance across our three complementary subsets, with each subset presenting distinct challenges that expose different model capabilities.

\textbf{CholecT50-Caption-VQA Analysis:} For this subset, several zero-shot models demonstrate reasonable transferability to surgical domain perception. \rev{LLaVA-Video and LongVA achieve 50.79\% and 47.73\% correctness respectively, with relevance scores of 56.68\% and 55.71\%}, suggesting that their pre-training on general-domain video data provides some foundational capabilities for surgical scene understanding. In contrast, InternVideo2.5 achieves substantially lower scores (\rev{12.54\% correctness, 14.50\% relevance}), with consistency across all three GPT metrics indicating systematic limitations likely stemming from domain distribution discrepancies in its training corpus.

SurgTEMP achieves better performance over the strongest fine-tuned baseline VideoGPT+-ft on this subset: \rev{+2.63 percentage points in correctness, +4.13 in relevance, and +3.88 in linguistic quality}. For overlap metrics on CholecT50-Caption-VQA (Table~\ref{tab:overlap_metrics}), the trends largely mirror GPT-based evaluations.

\textbf{Endoscapes-VQA Analysis:} Our model demonstrates substantial improvements on this subset, with significant margins over the strongest fine-tuned baseline VideoGPT+-ft: \rev{+2.35 percentage points in correctness, +2.72 in relevance, and +1.66 in linguistic quality}. The overlap metrics (Table~\ref{tab:overlap_metrics}) reflect similar trends; while these margins are less dramatic than GPT-based metrics, they demonstrate improved semantic alignment with reference answers.

The performance gap between fine-tuned and zero-shot models is pronounced on Endoscapes-VQA. Even the best-performing \rev{open-source} zero-shot model (VideoGPT+) achieves only \rev{38.75\%} correctness, representing a \rev{34.96} percentage point deficit compared to our model\rev{, and the proprietary Gemini-3.1-Pro improves on it only marginally (39.83\%)}. This substantial discrepancy demonstrates that CVS assessment is a highly nuanced task requiring specialized domain expertise: models must understand not only anatomical structures but also the specific evaluation criteria defined in clinical guidelines, which cannot be acquired through general-domain pre-training alone.

\textbf{CholeScore-VQA Analysis:} This subset presents the most challenging evaluation scenario due to its variable task complexity and extended temporal context (average 710 seconds). Zero-shot models struggle severely: InternVideo2.5 achieves only \rev{1.69\%} correctness while the best-performing zero-shot model (VideoGPT+) reaches merely \rev{25.25\%} correctness. After domain fine-tuning, \rev{Qwen3-VL-ft demonstrates the strongest improvement among the baselines, reaching 49.40\% correctness---representing a gain of over 24 percentage points compared to the best zero-shot baseline}.

\rev{Our SurgTEMP achieves further improvements over the strongest fine-tuned baseline Qwen3-VL-ft: +16.89 percentage points in correctness, +14.45 in relevance, and +8.85 in linguistic quality.} These substantial margins validate our architectural design motivation: long-range video understanding for comprehensive evaluation requires sophisticated temporal context aggregation mechanisms that can identify and retain information across extended temporal sequences. \rev{The overlap metrics on CholeScore-VQA (Table~\ref{tab:overlap_metrics}) follow the same ordering on BLEU, METEOR and ROUGE-L, where our model leads every baseline by a clear margin; on CIDEr the two are level, with VideoGPT+-ft at $15.38$ against our $15.34$, a difference well inside the run-to-run spread.}

\subsubsection{Task Hierarchy Performance Analysis}
\begin{figure*}[t!]
    \centering
    \includegraphics[width=0.95\textwidth]{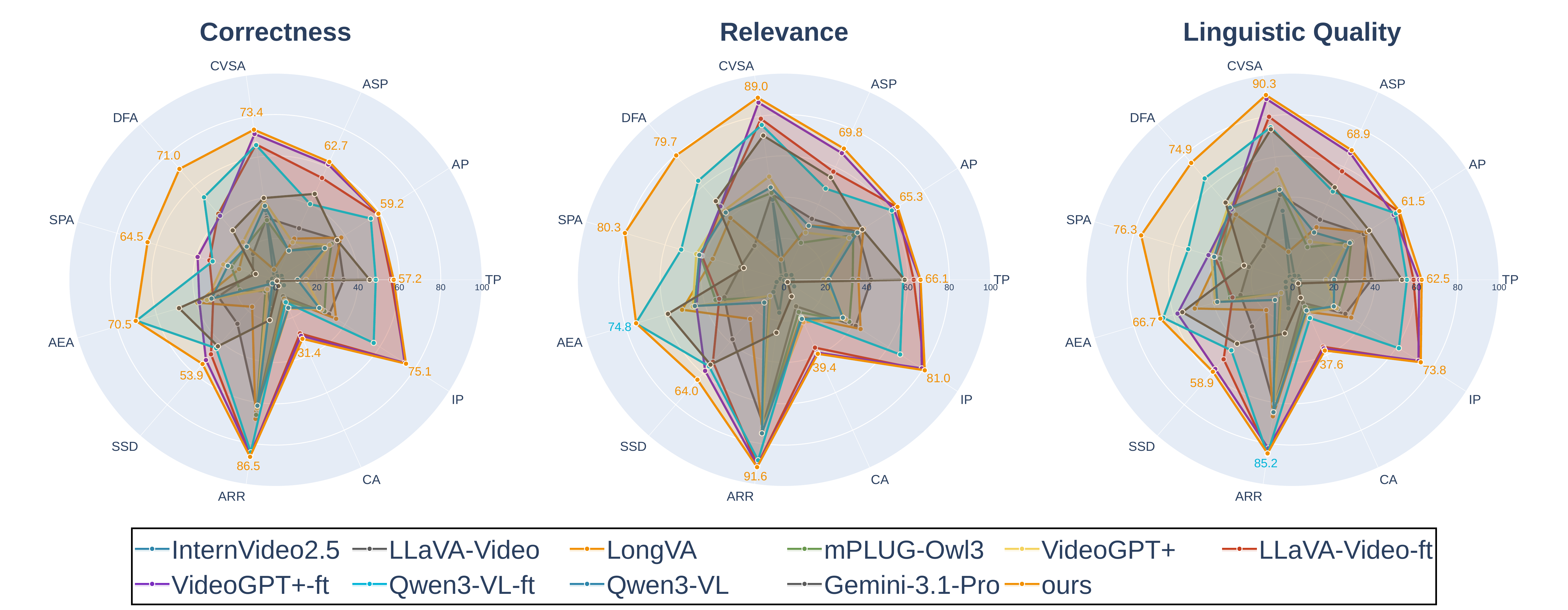}
    \caption{Performance comparison across task hierarchies and granular task categories. The radar chart visualizes model performance on three hierarchical levels: \textbf{Perception} (basic surgical     scene understanding including tool [TP], action [AP], and anatomical structure recognition [ASP]), \textbf{Assessment} (safety-focused assessments including CVS criteria [CVSA], difficulty findings [DFA], skill proficiency [SPA], and adverse events [AEA]), and \textbf{Reasoning} (surgical scene description [SSD], action rationale reasoning [ARR], comprehensive assessment [CA] and intraoperative planing [IP]). Results are averaged across Correctness, Relevance, and Linguistic Quality metrics. Our SurgTEMP demonstrates balanced excellence across all hierarchy levels.}
    \label{fig:radar_chart}
\end{figure*}

To provide deeper insights into model capabilities across different clinical complexity levels, we analyze performance on the three-tiered task hierarchy proposed in our dataset design: \textit{Perception}, \textit{Assessment}, and \textit{Reasoning}. Figure~\ref{fig:radar_chart} visualizes the averaged performance across Correctness, Relevance, and Linguistic Quality metrics for each task category, revealing distinct performance patterns that highlight the unique challenges of surgical assessment.

\textbf{Perception-Level Tasks:} Among \rev{open-source} zero-shot models, LLaVA-Video demonstrates the strongest performance on all three tasks, with LongVA and mPLUG coming in second place with decent action perception recognition capability. VideoGPT+ shows comparatively weaker perception, suggesting its strengths lie more in assessment-level understanding. \rev{All three fine-tuned baselines---LLaVA-Video-ft, VideoGPT+-ft and Qwen3-VL-ft---achieve} substantial improvements over their zero-shot counterparts, validating that surgical domain exposure enables better recognition of specialized instruments, anatomical structures, and procedural actions. Our SurgTEMP on these three tasks shows modest margins over \rev{all three fine-tuned baselines} on action and tool recognition while showing a decent margin on anatomical structures.

\textbf{Assessment-Level Tasks:} This hierarchy represents a more challenging and expertise-demanding category, where most zero-shot models struggle significantly. After domain fine-tuning, \rev{all three fine-tuned baselines} achieve substantial improvements. Our SurgTEMP demonstrates the most improvement on this hierarchy, achieving \rev{69.87\%} average correctness, which shows a notable margin over \rev{all three fine-tuned baselines}. This substantial gain validates our architectural design hypothesis: the TEMP module's text-guided memory construction mechanism enables targeted attention to assessment-related visual cues (e.g. anatomical landmarks for CVS assessment, surgical maneuvers for skill evaluation, abnormal tissue appearances for adverse event detection) that are essential for accurate clinical assessment.

Examining individual tasks, CVSA shows the greatest benefit from domain adaptation, in which zero-shot models achieve less than 40\% correctness, yet after fine-tuning, performance boosts to \rev{the 66--73\% range}. For difficulty findings and skill proficiency assessments, our model shows significant margins over all baselines, which represents the critical need for adaptive attention allocation on temporal span. For adverse event detection, \rev{VideoGPT+-ft and LLaVA-Video-ft still cannot achieve reasonably good performance, whereas Qwen3-VL-ft reaches 70.20\% correctness, close to our 70.53\%}. We hypothesize \rev{the shortfall of the former} is because adverse events rely on detecting fine-grained and sudden status changes of anatomical structures.

\textbf{Reasoning-Level Tasks:} This hierarchy encompasses the most cognitively complex tasks requiring synthesis of multiple information sources. For descriptive tasks SSD and CA, which rely on surgical scene understanding and intraoperative assessment capabilities, zero-shot models fall short. After fine-tuning, \rev{all three fine-tuned baselines} achieve substantial improvements, and our model achieves a visible margin over \rev{each of them}. For the action rationale task, which requires more basic domain understanding including the maneuvers and intent of operative actions, general domain knowledge transfer works better for zero-shot models, and after fine-tuning, the gains remain positive. Yet for the intraoperative planning task that requires contextual analysis and understanding of potential risk management expertise, zero-shot models struggle, with performance mostly below 30\% correctness.

\subsubsection{Ablation Study}
\begin{table}[t!]
\centering
\revtable
\footnotesize
\setlength{\tabcolsep}{2pt}
\begin{tabular}{l|>{\centering\arraybackslash}m{0.55cm} >{\centering\arraybackslash}m{0.55cm} >{\centering\arraybackslash}m{0.55cm} >{\centering\arraybackslash}m{0.55cm} >{\centering\arraybackslash}m{0.55cm}|>{\centering\arraybackslash}m{1.16cm} >{\centering\arraybackslash}m{1.16cm} >{\centering\arraybackslash}m{1.16cm}}
\hline
\multirow{2}{*}{Models} & \multicolumn{5}{c|}{Components} & \multicolumn{3}{c}{GPT Scores} \\
\cline{2-9}
        & VI & TMB & TAS & LS & SCP & CR & RL & LG \\
\hline
w/o TMB & {\color{green}\ding{51}} & {\color{red}\ding{55}} & {\color{green}\ding{51}} & {\color{green}\ding{51}} & {\color{green}\ding{51}} & 66.49$_{\pm.14}$ & 76.43$_{\pm.17}$ & 73.58$_{\pm.05}$ \\
w/o TAS & {\color{green}\ding{51}} & {\color{green}\ding{51}} & {\color{red}\ding{55}} & {\color{green}\ding{51}} & {\color{green}\ding{51}} & 63.93$_{\pm.00}$ & 71.20$_{\pm.03}$ & 68.76$_{\pm.16}$ \\
w/o LS  & {\color{green}\ding{51}} & {\color{green}\ding{51}} & {\color{green}\ding{51}} & {\color{red}\ding{55}} & {\color{green}\ding{51}} & 66.96$_{\pm.13}$ & 76.07$_{\pm.11}$ & 74.20$_{\pm.18}$ \\
w/o SCP & {\color{green}\ding{51}} & {\color{green}\ding{51}} & {\color{green}\ding{51}} & {\color{green}\ding{51}} & {\color{red}\ding{55}} & 69.78$_{\pm.04}$ & 80.49$_{\pm.18}$ & 78.16$_{\pm.15}$ \\
w/o VI  & {\color{red}\ding{55}} & {\color{green}\ding{51}} & {\color{green}\ding{51}} & {\color{green}\ding{51}} & {\color{green}\ding{51}} & 8.14$_{\pm.04}$ & 10.44$_{\pm.15}$ & 9.34$_{\pm.04}$ \\
\hline
Full Model & {\color{green}\ding{51}} & {\color{green}\ding{51}} & {\color{green}\ding{51}} & {\color{green}\ding{51}} & {\color{green}\ding{51}} & \textbf{71.64$_{\pm.07}$} & \textbf{81.76$_{\pm.13}$} & \textbf{79.04$_{\pm.13}$} \\
\hline
\end{tabular}   
\caption{Ablation study evaluating the contribution of each architectural component. VI: Visual Input, TMB: Temporal Memory Bank, TAS: Text-guided Attention Selection, LS: Learnable Separator, SCP: Surgical Competency Progression. {\color{green}\ding{51}} indicates component present, {\color{red}\ding{55}} indicates component removed. GPT Scores: Correctness (CR), Relevance (RL), Linguistic Quality (LG).}\label{tab:ablation_study}
\end{table}

To validate the individual contributions of our proposed architectural components and strategy, we conduct a systematic ablation study by progressively removing key modules from SurgTEMP. Table~\ref{tab:ablation_study} presents the averaged performance across all three CholeVidQA-32K subsets.

\textbf{Temporal Memory Bank (TMB):} Removing the hierarchical temporal memory pyramid construction (ours w/o TMB) results in performance degradation of 5.15, 5.33, and 5.46 percentage points in correctness, relevance, and linguistic quality respectively compared to the full model. This substantial drop validates the importance of the temporal memory bank, which provides a proxy for coarse long-range observation aggregation for surgical video understanding.

\textbf{Text-guided Attention Selection (TAS):} Disabling the text-guided visual token selection mechanism, which we replace the attention-based top-$k$ frame selection with a randomized one, (ours w/o TAS) causes a notable performance drop: correctness drops by 7.71 points, relevance by 10.56 points, and linguistic quality by 10.28 points. This degradation demonstrates that the TAS module's ability to dynamically prioritize surgical task-relevant visual clues.

\textbf{Learnable Separator (LS):} Replacing the learnable separator tokens with fixed all-zero vectors (ours w/o LS) results in performance degradation of 4.68 points in correctness, 5.69 points in relevance, and 4.84 points in linguistic quality. This consistent drop validates that learnable separators, compared to simple zero-vector separators, provide more meaningful structural boundaries that guide hierarchical memory organization. Without content-aware learned separators to demarcate temporal hierarchy levels, the model's ability to maintain coherent spatial-temporal relationships across the memory pyramid is weakened, reducing precision in both spatial and temporal reasoning.

\textbf{Surgical Competency Progression (SCP):} Removing the SCP training scheme (ours w/o SCP) results in moderate performance degradation of 1.86 points in correctness, 1.27 points in relevance, and 0.88 points in linguistic quality. While the performance drop is less severe than removing architectural components, these decreases demonstrate the importance of our progressive training approach that gradually introduces tasks of increasing complexity. The SCP training scheme enables more robust feature representations by first establishing strong perceptual foundations before building higher-level clinical reasoning capabilities.

\textbf{Component Synergy:} The ablation results reveal that all components and training strategies contribute substantially and complementarily to model performance. The cumulative performance gap between the worst ablation (ours w/o TAS: 63.93\% CR) and the full model (71.64\% CR) reaches 7.71 percentage points, demonstrating that our architectural innovations provide meaningful improvements over baseline temporal fusion approaches.

\textbf{Visual Inputs (VI):} To validate that the model does not shortcut by relying solely on textual questions to produce answers, we remove all visual inputs. The dramatic performance collapse confirms both the validity of the dataset generation pipeline and the effectiveness of the model in leveraging visual signals rather than overfitting on textual associations between questions and answers.

\subsubsection{Parameter Sensitivity Analysis}
\label{subsec:param_sensitivity}
\begin{figure*}[t!]
    \centering
    \begin{subfigure}[b]{0.48\textwidth}
        \centering
        \includegraphics[width=\textwidth]{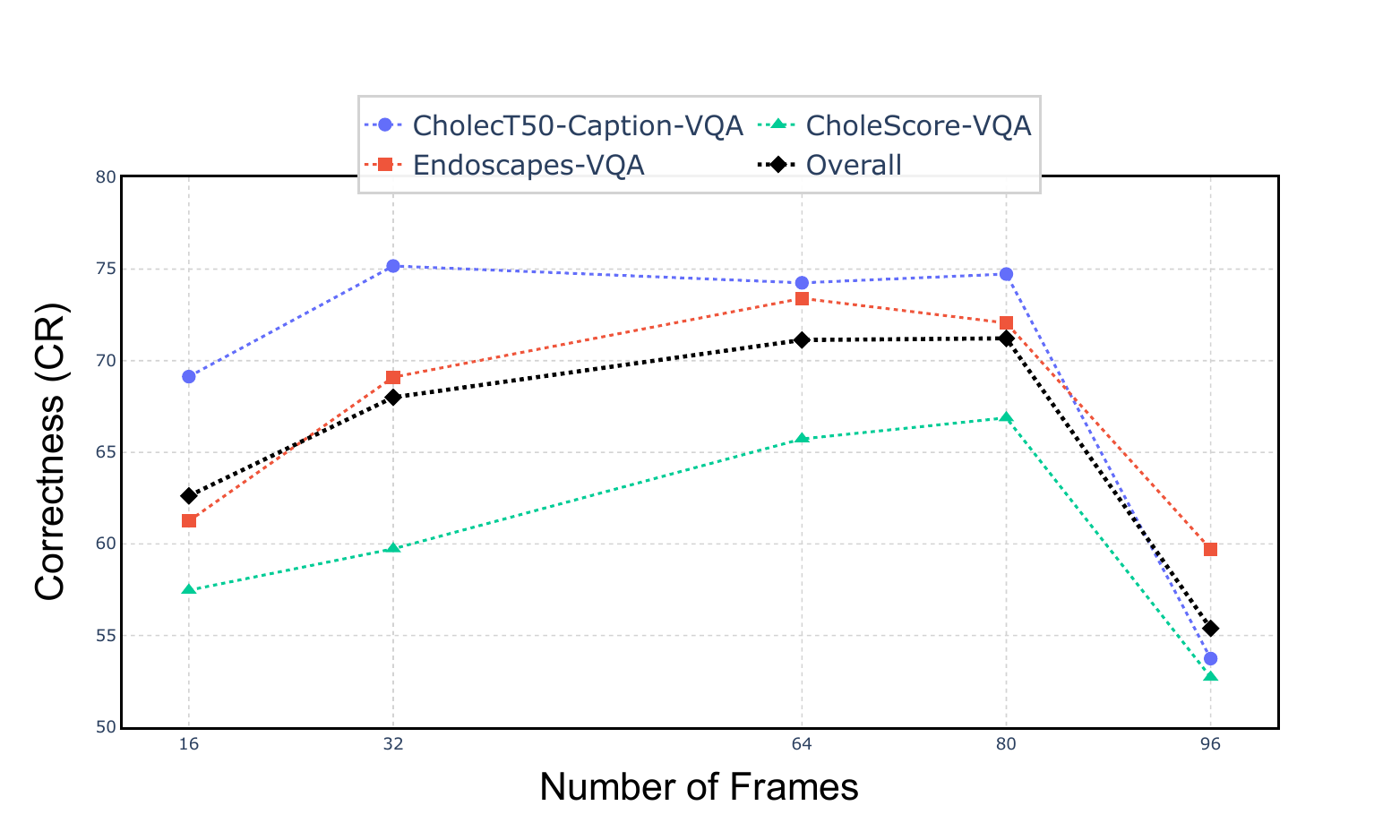}
        \caption{Frame sampling sensitivity}
        \label{fig:frames_sensitivity}
    \end{subfigure}
    \hfill
    \begin{subfigure}[b]{0.48\textwidth}
        \centering
        \includegraphics[width=\textwidth]{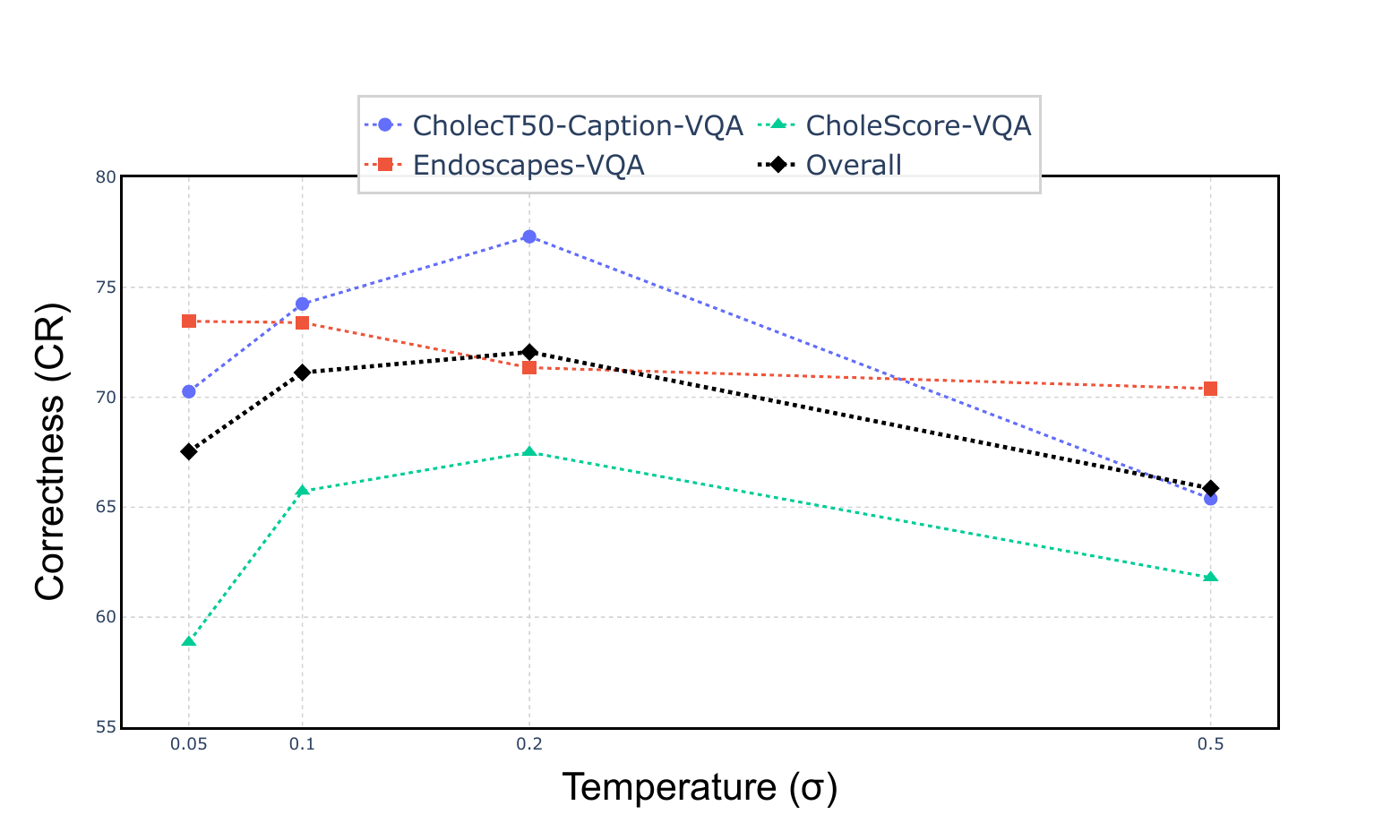}
        \caption{Temperature parameter sensitivity}
        \label{fig:sigma_sensitivity}
    \end{subfigure}
    \caption{Parameter sensitivity analysis of our TEMP module. (a) Impact of frame sampling rate: Correctness (CR) scores as a function of uniformly sampled frames (16, 32, 64, 80, 96). Performance improves with denser temporal sampling up to 64 frames, then degrades at 96 frames \rev{under the current fixed top-$k$ selection setting}. (b) Impact of Gumbel-Softmax temperature ($\sigma$): CR scores across temperature values (0.05, 0.1, 0.2, 0.5). \rev{Optimal temperature exhibits subset-specific patterns, with Endoscapes-VQA benefiting from sharp selection ($\sigma = 0.05$) while the other two subsets perform best with moderate softness ($\sigma = 0.2$).}}
    \label{fig:parameter_sensitivity}
\end{figure*}

\rev{Both sweeps in this subsection are hyper-parameter searches carried out as single runs on one seed, and each varies one factor while holding the other at the value used during the search ($\sigma = 0.1$ for the frame sweep, $64$ frames for the temperature sweep). Their absolute correctness values are therefore not directly comparable to Tables~\ref{tab:overall_results}--\ref{tab:overlap_metrics}, which report the final configuration averaged over three seeds; the sweeps are reported for the shape of the trends they expose, and all statements below refer to positions within a sweep rather than to final model performance.}

\textbf{Frame Sampling Upper-Bound Analysis:} To \rev{characterize} the temporal scaling behavior of our TEMP module, we conduct systematic experiments varying the number of uniformly sampled frames from 16 to 96. As illustrated in Figure~\ref{fig:frames_sensitivity}, model performance exhibits a clear non-monotonic relationship with temporal sampling density across all three subsets. During the initial scaling phase (16 to 64 frames), correctness consistently improves across all subsets as increased temporal coverage enables better understanding of surgical context and procedural progression. However, the rate of improvement varies significantly between subsets. CholecT50-Caption-VQA exhibits early performance saturation, \rev{peaking at 32 frames and gaining nothing thereafter (75.17\% at 32 frames vs.\ 74.25\% at 64 and 74.73\% at 80 frames)}. This early plateau indicates that the subset's relatively short video segments and straightforward question types require limited long-range temporal dependencies, making dense sampling beyond 32 frames unnecessary for this particular task distribution. In contrast, Endoscapes-VQA and CholeScore-VQA demonstrate sustained performance improvements throughout the scaling range, reaching peak performance at 64-80 frames. Endoscapes-VQA achieves 73.40\% correctness at 64 frames compared to 61.25\% at 16 frames (+12.15 points), while CholeScore-VQA \rev{continues to improve over a longer range, peaking only at 80 frames (66.88\% versus 57.47\% at 16 frames, +9.41 points); its total gain is smaller in magnitude than that of Endoscapes-VQA but is accrued more gradually, consistent with evidence that is spread across a procedural phase rather than concentrated in a short clip}. These substantial improvements \rev{indicate} that intraoperative assessment tasks—particularly CVS evaluation requiring fine-grained anatomical detail and skill proficiency assessment requiring comprehensive temporal context—benefit significantly from denser temporal sampling \rev{within the effective operating range of the current TEMP configuration}.

Beyond the optimal operating point at 64 frames, all subsets experience performance degradation at 96 frames, with correctness dropping dramatically: CholecT50-Caption-VQA falls to 53.74\% (-20.99 points from 80 frames), Endoscapes-VQA to 59.71\% (-12.35 points), and CholeScore-VQA to 52.72\% (-14.16 points). We hypothesize this degradation stems from two compounding factors: (1) \textbf{information saturation}, where excessive temporal sampling introduces redundant visual content that dilutes attention to critical frames, and (2) \textbf{capacity limitations} of the TEMP module's top-$k$ selection mechanism, which struggles to effectively filter salient information from an overwhelming pool of 96-frame candidates, leading to suboptimal token selection and reduced reasoning quality. \rev{This result should therefore be interpreted as evidence that the current frame-selection design has an effective temporal range, rather than as evidence of unbounded scalability to arbitrarily long videos.} \rev{Overall correctness is essentially flat between $64$ and $80$ candidate frames ($71.13\%$ versus $71.22\%$) and collapses at $96$, so we retain $64$ frames as the default configuration: among the tested settings it achieves the best trade-off between temporal coverage and computational cost, rather than representing a universal optimum.}

\rev{A complementary control experiment, sweeping the test-time frame budget of the fine-tuned baselines themselves so that they can be compared to SurgTEMP at matched numbers of frames, is reported in Appendix~\ref{appendix:baseline_frames}.}

\textbf{Frame Selection Temperature Parameter Analysis:} Complementing the frame sampling analysis, we investigate the sensitivity of our TEMP module to the Gumbel-Softmax temperature parameter $\sigma$, which controls the sharpness of differentiable frame selection. As shown in Figure~\ref{fig:sigma_sensitivity}, the optimal temperature exhibits a clear inverted-U relationship with performance, with subset-specific peaks that reflect the underlying temporal structure of safety-critical information. Endoscapes-VQA \rev{peaks within this sweep} at $\sigma = 0.05$ (73.46\% correctness), demonstrating that Critical View of Safety assessment benefits from sharp, near-deterministic frame selection to precisely localize anatomical landmarks (cystic duct, cystic artery, hepatocystic triangle) within brief 5-second clips where critical structures appear in specific frames. This sharp selection enables the model to commit decisively to frames containing CVS-relevant anatomical features while filtering out transitional or irrelevant content.

In contrast, CholecT50-Caption-VQA and CholeScore-VQA \rev{peak at $\sigma = 0.2$ (77.31\% and 67.50\% correctness respectively). These two subsets differ sharply in clip length, so what they share is not temporal extent but the fact that their answers draw on evidence spread over the whole clip rather than on a single annotated frame: the caption-derived perception, scene-description and action-rationale questions of CholecT50-Caption-VQA are grounded in the action as it unfolds, while the difficulty, skill and adverse-event questions of CholeScore-VQA accumulate evidence across a procedural phase. Both therefore favour softer selection, which lets complementary frames contribute instead of committing to a single one.} The moderate temperature allows the model to maintain gradient flow during training while preserving sufficient selectivity to identify safety-critical patterns across extended procedural phases.

Performance degradation at extreme temperature values validates our architectural design choices. At $\sigma = 0.5$, excessive softness causes diffuse attention where selection becomes too uniform across frames, diluting safety-critical visual cues with irrelevant content—overall correctness drops to 65.86\%, consistent with the information saturation observed at 96 frames. Conversely, excessively sharp selection ($\sigma = 0.05$) \rev{harms the two subsets whose evidence is distributed across the clip} by over-committing to specific frames during training, limiting gradient flow and reducing the model's ability to discover diverse safety-critical patterns: CholecT50-Caption-VQA and CholeScore-VQA drop to 70.26\% and 58.86\% respectively. \rev{No single temperature is best on every subset; we therefore select $\sigma = 0.2$ as the default because it maximizes correctness averaged over the three subsets ($72.06\%$, against $71.13\%$ at $\sigma = 0.1$ and $67.53\%$ at $\sigma = 0.05$),} balancing differentiability for end-to-end training with sufficient selectivity for effective frame filtering, while demonstrating that the \rev{per-subset} operating point depends on \rev{how task-relevant evidence is distributed within a clip rather than on clip length alone}.

\subsubsection{\rev{Results Analysis on the Expert-Verified Test Subset}}
To validate the alignment between model-generated responses and real-world clinical reasoning processes, we construct \rev{an expert-verified test subset} comprising 300 expert-annotated question-answer pairs spanning four assessment-level tasks. Unlike the main test set where ground-truth answers are derived from structured clinical annotations and repurposed by \rev{an LLM}, this \rev{subset} features answers directly authored by clinical experts based on video content and surgical domain knowledge, providing a benchmark for evaluating models' capability to align with authentic clinical reasoning patterns. \rev{As presented in Table~\ref{tab:golden_set_results_additional}, our SurgTEMP consistently outperforms all baselines across GPT-based metrics on this expert-verified test subset.}

Examining individual task performance reveals distinct patterns reflecting varying assessment complexities. For CVSA, our model outperforms \rev{VideoGPT+-ft by +4.02 points}. In contrast, difficulty findings assessment (DFA) exhibits \rev{larger gains (+7.59 points over Qwen3-VL-ft)}, indicating that identifying operative difficulty factors (e.g., dense adhesions, visceral fat) demands more sophisticated visual pattern recognition and contextual interpretation that our TEMP module's text-guided attention mechanism effectively captures. \rev{The narrowest margin emerges in adverse event assessment (AEA), where our model leads the strongest baseline Qwen3-VL-ft by +1.48 points while still surpassing VideoGPT+-ft by +49.67 points. This spread across baselines validates} that detecting subtle tissue abnormalities, thermal injuries, and bleeding complications requires fine-grained temporal-spatial attention to transient visual cues that standard architectures often overlook. For skill proficiency assessment (SPA), our model achieves \rev{the largest margin among assessment tasks, +24.41 points} over VideoGPT+-ft, confirming the importance of comprehensive temporal context integration for evaluating surgical technique quality across extended procedural phases. These results collectively demonstrate that our model's hierarchical memory construction and text-guided selection mechanisms enable closer alignment with expert clinical reasoning, particularly for tasks requiring detection of subtle safety-critical visual patterns and integration of long-range temporal dependencies.

\begin{table*}[t!]
\centering
\revtable
\small
\begin{tabular}{l|>{\centering\arraybackslash}m{0.9cm} >{\centering\arraybackslash}m{1.2cm}|>{\centering\arraybackslash}m{0.9cm} >{\centering\arraybackslash}m{1.2cm}|>{\centering\arraybackslash}m{0.9cm} >{\centering\arraybackslash}m{1.2cm}|>{\centering\arraybackslash}m{0.9cm} >{\centering\arraybackslash}m{1.2cm}}
\hline
\multirow{2}{*}{Models} & \multicolumn{2}{c|}{CVSA} & \multicolumn{2}{c|}{DFA} & \multicolumn{2}{c|}{AEA} & \multicolumn{2}{c}{SPA} \\
\cline{2-9}
        & GPT & Overlap & GPT & Overlap & GPT & Overlap & GPT & Overlap \\
\hline
\multicolumn{9}{c}{\rule[-1ex]{0pt}{3ex}\textit{Open-source Zero-shot}} \\
mPLUG-Owl3 & 34.26$_{\pm.14}$ & 18.62$_{\pm.06}$ & 37.46$_{\pm.13}$ & 8.80$_{\pm.16}$ & 26.66$_{\pm.19}$ & 6.92$_{\pm.11}$ & 35.88$_{\pm.10}$ & \cellcolor{cyan!20}7.64$_{\pm.05}$ \\
InternVideo2.5 & 35.73$_{\pm.00}$ & 11.22$_{\pm.16}$ & 2.21$_{\pm.16}$ & 0.53$_{\pm.08}$ & 2.64$_{\pm.04}$ & 0.34$_{\pm.11}$ & 5.42$_{\pm.04}$ & 2.15$_{\pm.15}$ \\
LongVA & 10.36$_{\pm.13}$ & 10.59$_{\pm.08}$ & 40.62$_{\pm.13}$ & 7.82$_{\pm.17}$ & \cellcolor{cyan!20}43.75$_{\pm.01}$ & \cellcolor{cyan!20}6.96$_{\pm.06}$ & 34.56$_{\pm.16}$ & 6.69$_{\pm.06}$ \\
LLaVA-Video & 42.16$_{\pm.17}$ & 16.15$_{\pm.15}$ & 20.45$_{\pm.07}$ & 5.57$_{\pm.16}$ & 33.07$_{\pm.12}$ & 4.53$_{\pm.09}$ & 19.09$_{\pm.03}$ & 5.82$_{\pm.02}$ \\
VideoGPT+ & \cellcolor{cyan!20}48.52$_{\pm.07}$ & \cellcolor{cyan!20}19.16$_{\pm.05}$ & \cellcolor{cyan!20}43.29$_{\pm.20}$ & \cellcolor{cyan!20}9.39$_{\pm.05}$ & 25.25$_{\pm.08}$ & 6.21$_{\pm.16}$ & \cellcolor{cyan!20}35.90$_{\pm.19}$ & 6.98$_{\pm.08}$ \\
Qwen3-VL & 45.18$_{\pm.08}$ & 16.82$_{\pm.04}$ & 43.07$_{\pm.12}$ & 8.74$_{\pm.19}$ & 36.67$_{\pm.14}$ & 5.74$_{\pm.03}$ & 32.73$_{\pm.04}$ & 6.41$_{\pm.12}$ \\
\hline
\multicolumn{9}{c}{\rule[-1ex]{0pt}{3ex}\textit{Proprietary Zero-shot}} \\
Gemini-3.1-Pro & 60.19$_{\pm.08}$ & 17.18$_{\pm.03}$ & 49.67$_{\pm.17}$ & 8.58$_{\pm.05}$ & 55.87$_{\pm.15}$ & 7.82$_{\pm.08}$ & 19.62$_{\pm.12}$ & 5.15$_{\pm.18}$ \\
\hline
\multicolumn{9}{c}{\rule[-1ex]{0pt}{3ex}\textit{Fine-tuned}} \\
VideoGPT+-ft & 80.13$_{\pm.01}$ & \cellcolor{orange!20}29.31$_{\pm.09}$ & 42.10$_{\pm.02}$ & 13.35$_{\pm.11}$ & 23.37$_{\pm.14}$ & 8.44$_{\pm.15}$ & 52.53$_{\pm.14}$ & 11.70$_{\pm.04}$ \\
LLaVA-Video-ft & 79.80$_{\pm.04}$ & 29.17$_{\pm.03}$ & 36.75$_{\pm.08}$ & \cellcolor{orange!20}13.75$_{\pm.00}$ & 33.63$_{\pm.05}$ & 8.93$_{\pm.04}$ & 45.56$_{\pm.12}$ & 11.94$_{\pm.06}$ \\
Qwen3-VL-ft & 74.47$_{\pm.12}$ & 28.64$_{\pm.06}$ & 62.99$_{\pm.20}$ & 13.59$_{\pm.05}$ & 71.56$_{\pm.02}$ & 8.69$_{\pm.10}$ & 43.10$_{\pm.01}$ & 12.49$_{\pm.00}$ \\
ours & \cellcolor{orange!20}84.15$_{\pm.07}$ & 28.84$_{\pm.02}$ & \cellcolor{orange!20}70.58$_{\pm.03}$ & 13.49$_{\pm.12}$ & \cellcolor{orange!20}73.04$_{\pm.14}$ & \cellcolor{orange!20}12.02$_{\pm.11}$ & \cellcolor{orange!20}76.94$_{\pm.04}$ & \cellcolor{orange!20}13.68$_{\pm.10}$ \\
\hline
\end{tabular}
\caption{\rev{Comparison on the expert-verified test subset} across assessment-level tasks using the updated model outputs. GPT indicates the averaged scores across correctness, relevance, and linguistic quality; Overlap indicates the weighted average of BLEU, METEOR, ROUGE-L, and CIDEr mapped to a 0-100 scale. Blue shading indicates the best open-source zero-shot result; orange shading indicates the best fine-tuned result. The proprietary zero-shot group contains a single model and is therefore left unshaded.}\label{tab:golden_set_results_additional}
\end{table*}

\subsubsection{Qualitative Results}
\begin{figure*}[t!]
    \centering  
    \includegraphics[width=0.95\textwidth]{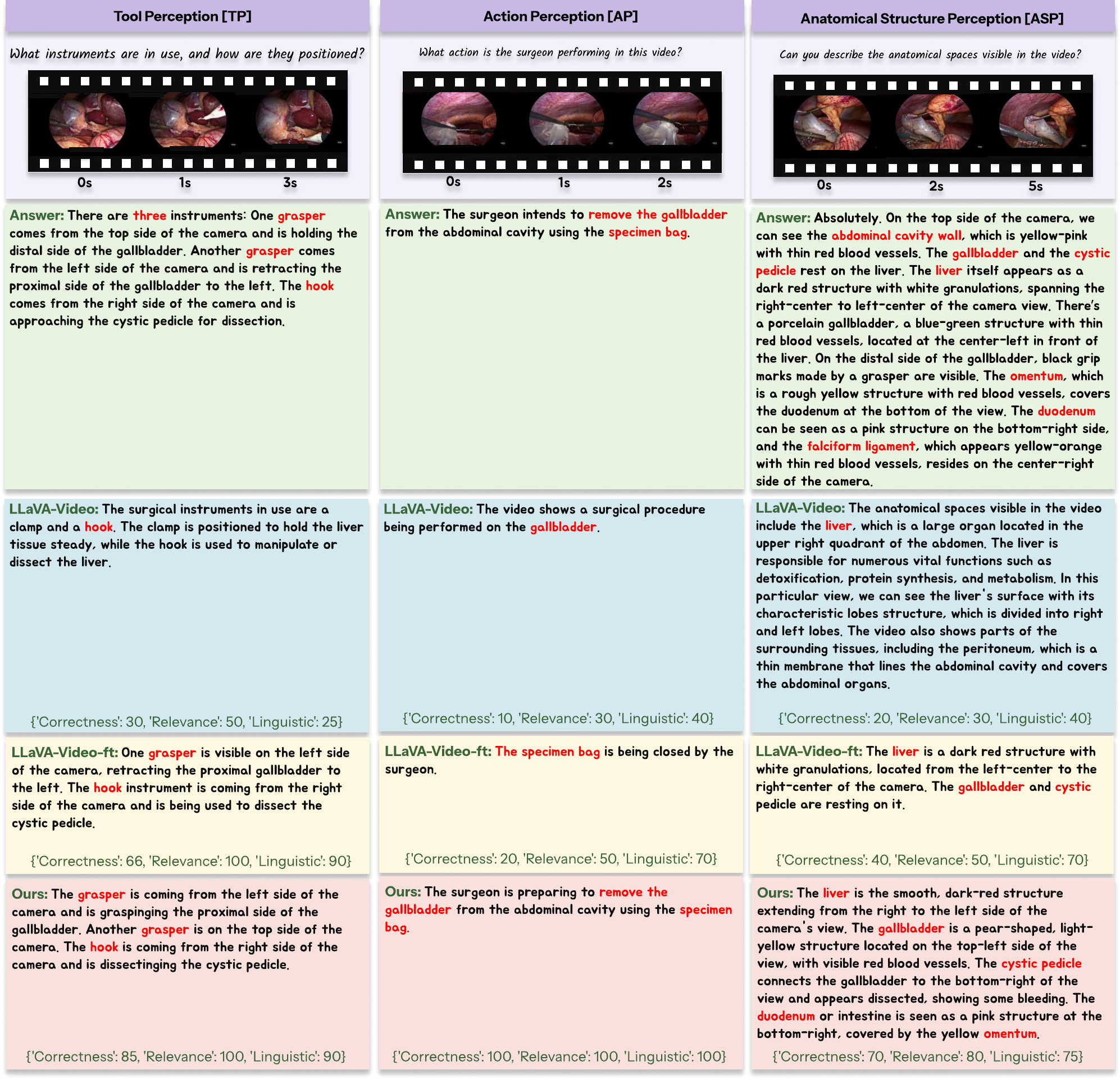}
    \caption{Qualitative comparison of model responses on perception-level tasks, with key answer highlighted with red color font. The figure demonstrates how our SurgTEMP generates clinically accurate and contextually appropriate responses for tasks including tool perception, action perception, and anatomical structure perception, compared to baseline models.}
    \label{fig:perception_qualitative}
\end{figure*}

\begin{figure*}[t!]
    \centering
    \includegraphics[width=0.95\textwidth]{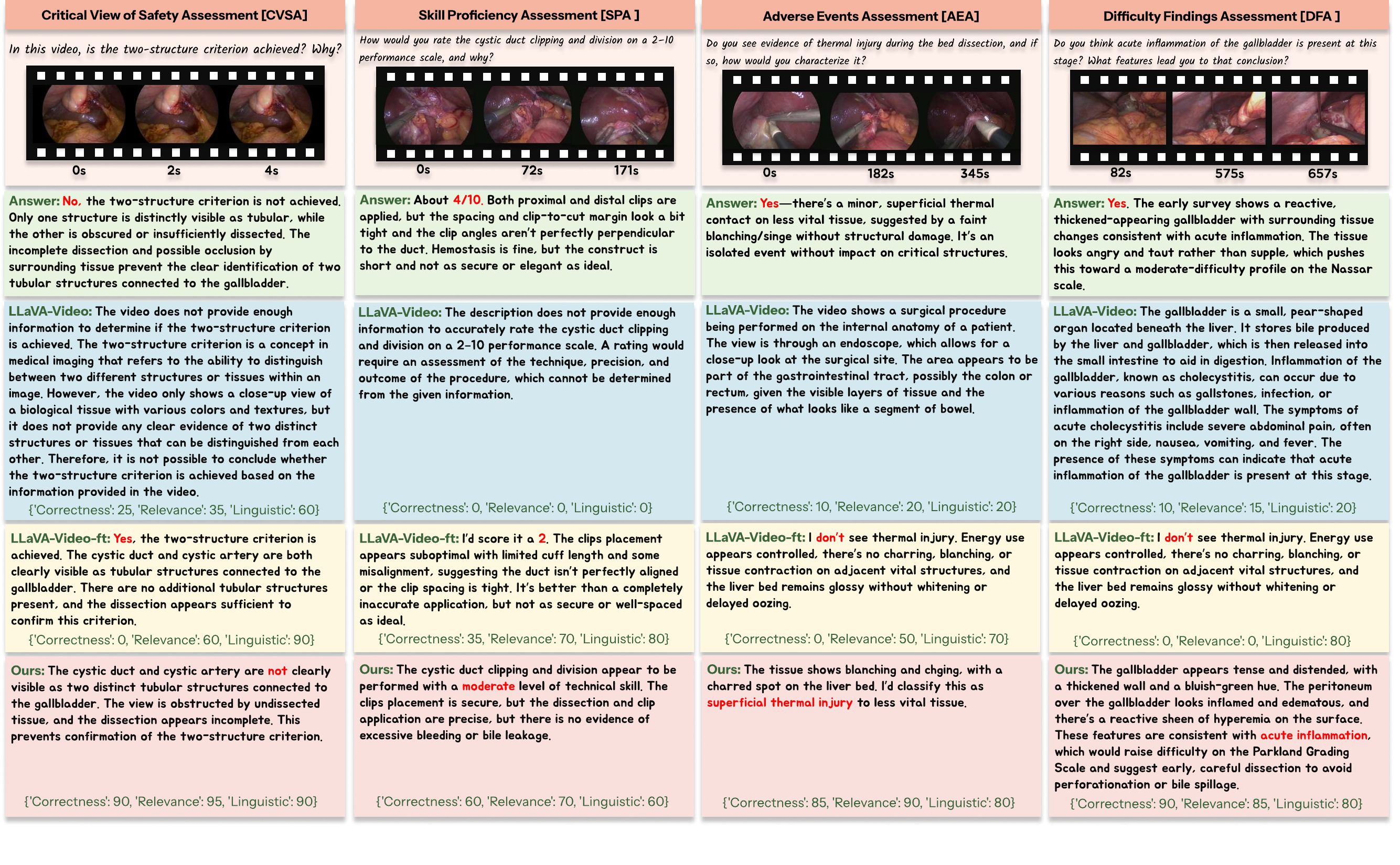}
    \caption{Qualitative comparison of model responses on assessment-level tasks with key answer highlighted with red color font. The figure demonstrates how our SurgTEMP generates clinically accurate and contextually appropriate responses for tasks including CVS, difficulty findings, IAE, and skill proficiency assessment, compared to baseline models.}
    \label{fig:assessment_qualitative}
\end{figure*}

\begin{figure*}[t!]
    \centering
    \includegraphics[width=0.95\textwidth]{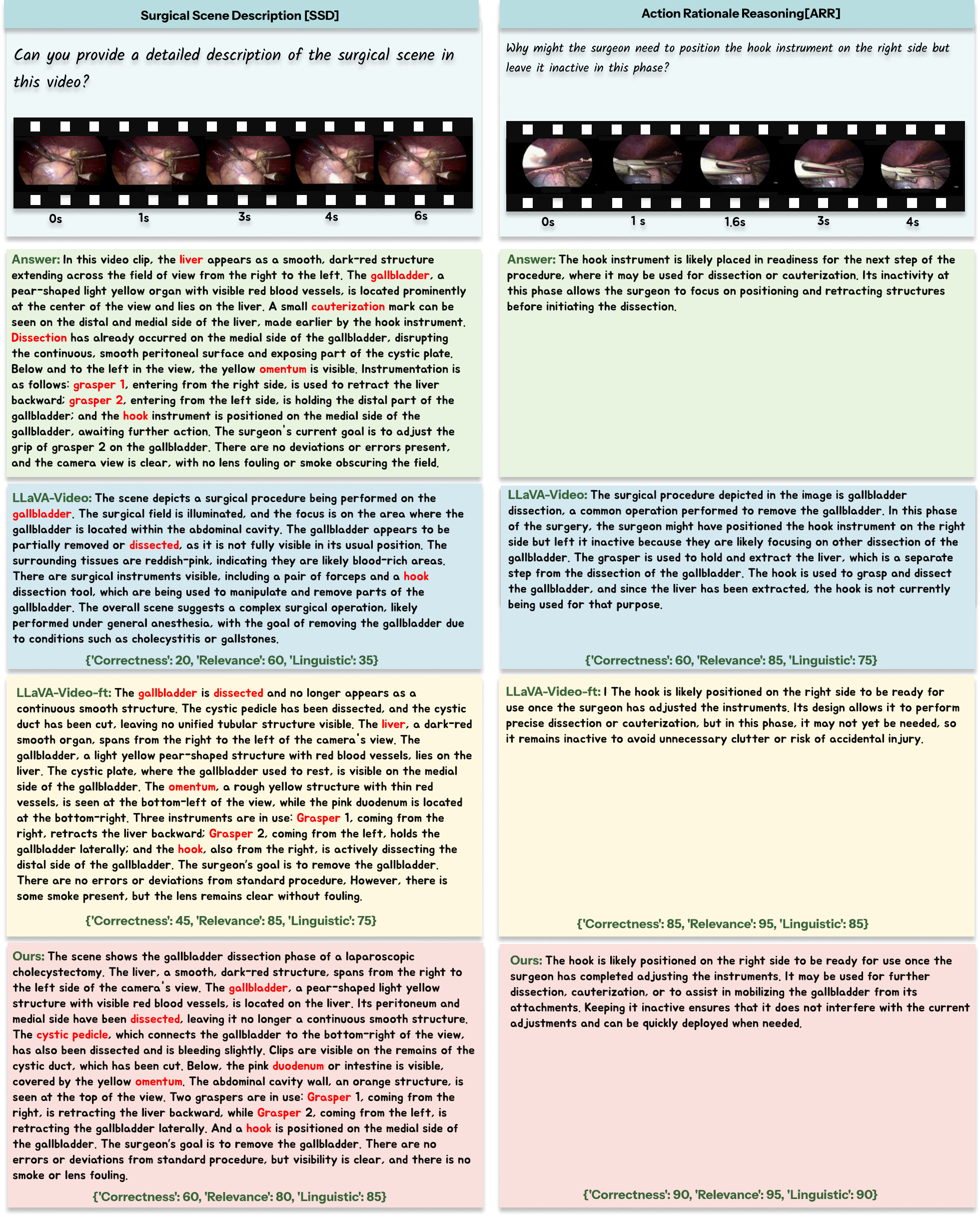}
    \caption{Qualitative comparison of model responses on reasoning-level tasks. The figure illustrates how our SurgTEMP generates comprehensive detailed descriptions and performs complex hypothetical reasoning about surgical scenarios, demonstrating superior temporal understanding and clinical reasoning capabilities compared to baseline models.}
    \label{fig:reasoning_qualitative_1}
\end{figure*}

\begin{figure*}[t!]
    \centering
    \includegraphics[width=0.95\textwidth]{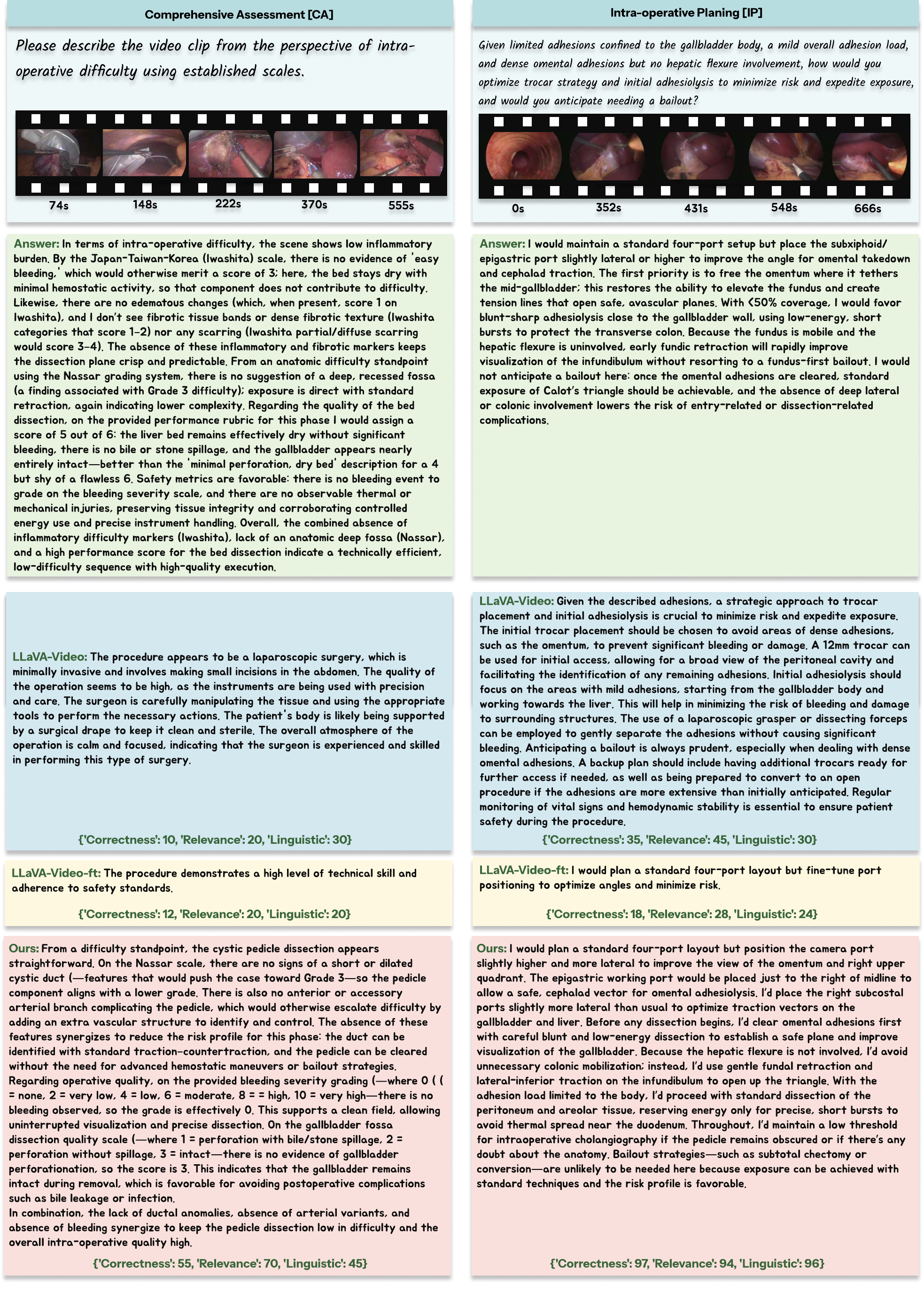}
    \caption{Qualitative comparison of model responses on reasoning-level tasks. The figure illustrates how our SurgTEMP generates comprehensive detailed descriptions and performs complex hypothetical reasoning about surgical scenarios, demonstrating superior temporal understanding and clinical reasoning capabilities compared to baseline models.}
    \label{fig:reasoning_qualitative_2}
\end{figure*}

To provide concrete insights into model behavior on each task, Figure ~\ref{fig:perception_qualitative}, ~\ref{fig:assessment_qualitative}, ~\ref{fig:reasoning_qualitative_1} and ~\ref{fig:reasoning_qualitative_2} presents qualitative examples of model responses on all level questions. The examples demonstrate that our SurgTEMP generates responses that are not only factually correct but also clinically appropriate and linguistically coherent, while baseline models often produce responses that are either too generic, factually incorrect, or fail to apply proper clinical assessment criteria.

\subsection{Frame Selection Visualization}
\begin{figure*}[t!]
    \centering
    \includegraphics[width=0.95\textwidth]{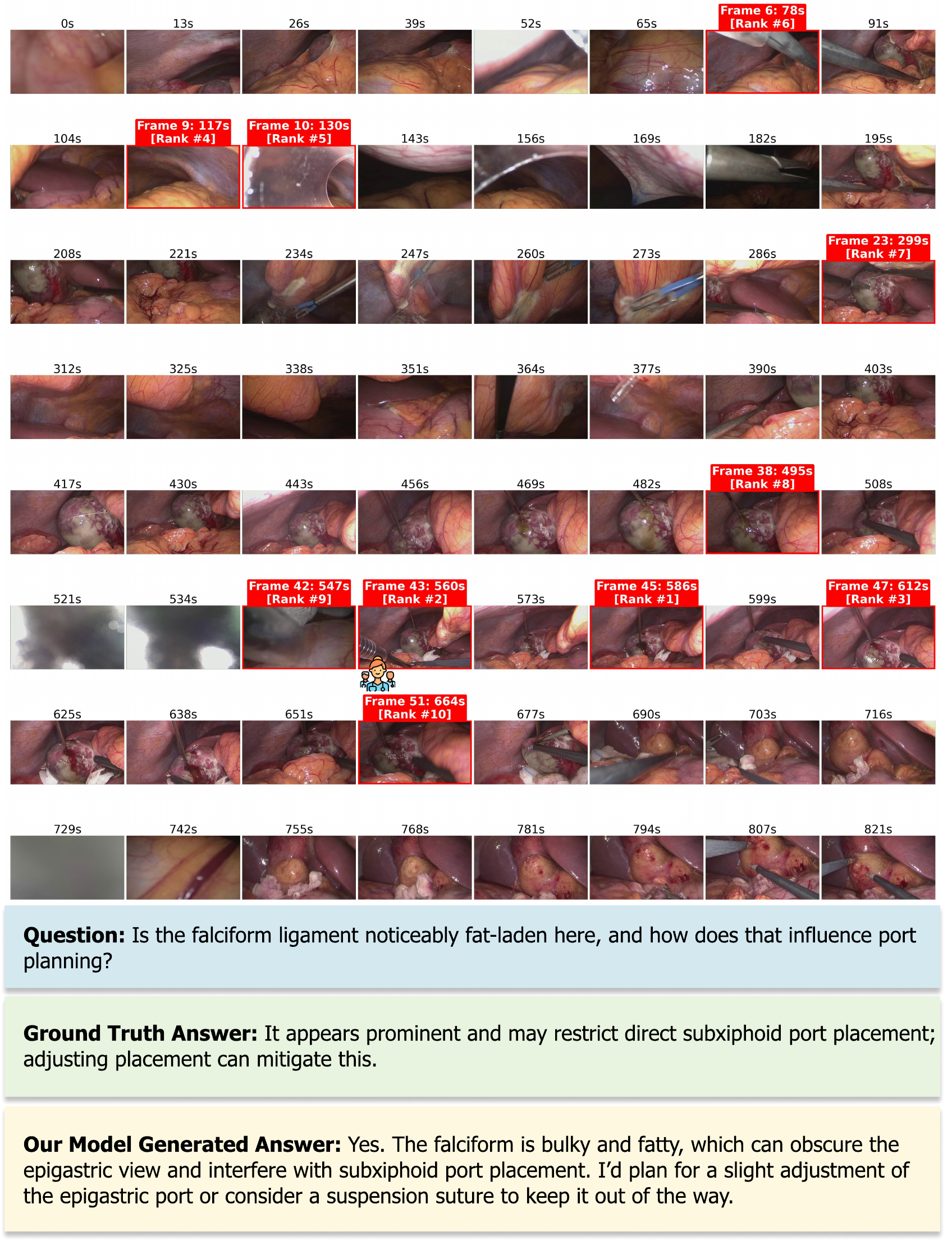}
    \caption{Visualization of the frame selection process in the TEMP module. Sixty-four frames are displayed in an 8×8 grid with timestamps shown above each frame. Ten selected frames are highlighted with red bounding boxes showing their rank during top-k selection. The most informative frame selected by a clinical expert is indicated with an icon in the bottom-left corner. Question, ground truth answer, and generated answer are shown below the frame grid.}
    \label{fig:frame_selection_vis_1}
\end{figure*}

\begin{figure*}[t!]
    \centering
    \includegraphics[width=0.95\textwidth]{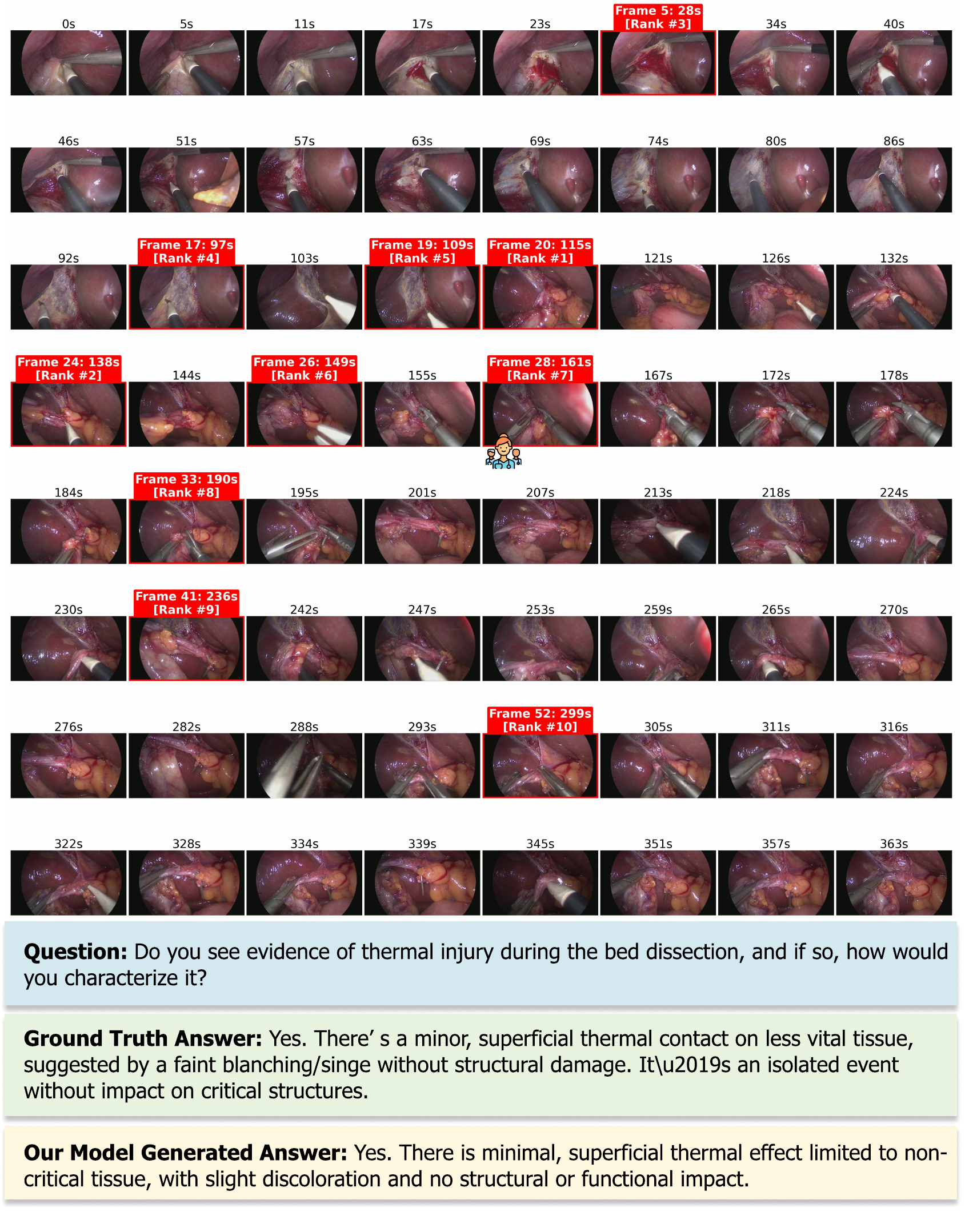}
    \caption{Visualization of the frame selection process in the TEMP module. Sixty-four frames are displayed in an 8×8 grid with timestamps shown above each frame. Ten selected frames are highlighted with red bounding boxes showing their rank during top-k selection. The most informative frame selected by a clinical expert is indicated with an icon in the bottom-left corner. Question, ground truth answer, and generated answer are shown below the frame grid.}
    \label{fig:frame_selection_vis_2}
\end{figure*}

To showcase the effectiveness of the frame selection mechanism in our proposed TEMP module, we provide visualizations in Figures~\ref{fig:frame_selection_vis_1} and~\ref{fig:frame_selection_vis_2}. In the first visualization (Figure~\ref{fig:frame_selection_vis_1}), the model accurately selects the most informative frame as its top-2 selection and actively avoids out-of-body frames that are not helpful to the assessment. In the second visualization (Figure~\ref{fig:frame_selection_vis_2}), the model selects the most informative frame as its top-7 selection, while also distributing attention to injury-relevant frames such as those showing bleeding and focusing on gallbladder bed dissection views.
\section{Discussion}
% 1 page for discussion

\subsection{Limitations and Future Work}
\label{sec:limitations}
While our work establishes a comprehensive benchmark for safety-critical surgical VQA and demonstrates the effectiveness of the proposed SurgTEMP architecture, several limitations merit discussion and suggest promising directions for future research.

\textbf{Dataset Scope and Procedural Coverage:} Our CholeVidQA-32K dataset focuses exclusively on laparoscopic cholecystectomy, which, while being one of the most commonly performed minimally invasive procedures and featuring well-established safety assessment frameworks (CVS, difficulty grading systems, OSATS), limits the generalizability of our findings to other surgical specialties. \rev{Although emerging surgical vision-language resources, such as SurgLaVi~\citep{perez2026surglavilargescalehierarchicaldataset}, and concurrent large-scale surgical multimodal efforts, such as Surg$\Sigma$~\citep{zeng2026surgsigmaspectrumlargescalemultimodal}, are valuable for broadening surgical representation learning, they do not yet provide an open, directly comparable benchmark for variable-length, clinically oriented surgical video question answering. As a result, our evaluation cannot establish how SurgTEMP would behave under substantial out-of-distribution shifts in procedure type, anatomy, operative workflow, camera perspective, instrument set, or assessment criteria. In such settings, the text-guided frame-selection mechanism may attend to visually salient but clinically non-transferable cues, and the SCP curriculum may encode cholecystectomy-specific task dependencies that do not map cleanly to other procedures.} Future work should extend the dataset to encompass diverse procedures such as colorectal surgery, bariatric surgery, and robotic-assisted interventions, each presenting unique anatomical challenges, instrument configurations, and safety considerations. Such expansion would enable investigation of cross-procedural transfer learning and development of procedure-agnostic safety assessment capabilities.

\rev{\textbf{Supervision Provenance and Residual Formulation Bias:} CholeVidQA-32K is constructed by converting existing expert annotations into question--answer form with LLM assistance, and this pipeline leaves residual formulation bias that we do not claim to have removed. First, the underlying supervision is inherited rather than re-derived: the tool, action, and anatomy vocabulary of CholecT50, the three-criterion definition and annotation frequency of CVS in Endoscapes, and the phase-level difficulty, adverse-event, and OSATS rubrics of CholeScore jointly determine what can be asked and what counts as a correct answer. Any labelling convention, class imbalance, or rater tendency present in those source datasets therefore propagates into our QA pairs, and our curation can reorganize this supervision but cannot correct it. Second, the surface form of the questions and answers is shaped by the generation prompts. The iterative quality-control loop documented in Appendix~\ref{appendix:feedback_loop} was designed to surface and correct precisely such prompt-induced artifacts, and we report its review protocol, recurring failure modes, attributed root causes, corrective actions, and before/after examples so that this stage is auditable rather than opaque; it nevertheless mitigates rather than eliminates the effect. Third, LLM-based rewriting imposes a stylistic regularity---phrasing, answer length, and discourse structure---that is not characteristic of spontaneous clinical language, so a model fine-tuned on this corpus may partly acquire this register alongside the clinical content, and the expert-verified test subset only partially controls for it. We regard the systematic characterization and mitigation of LLM-rewriting bias in medical QA corpora as an important and promising open problem, and it is the specific focus of our ongoing follow-up work. Accordingly, CholeVidQA-32K should be read as a clinically grounded benchmark with acknowledged residual formulation bias, rather than as a bias-free clinical resource.}

\rev{\textbf{Benchmark Coverage of Frontier Multimodal Models:} The baselines compared here are selected for being video-capable at comparable input settings, reproducible, and---where weights are open---fine-tunable under a single controlled protocol, which is what allows the comparison to isolate the contribution of surgical-domain adaptation and of the TEMP module. General-purpose multimodal models are released at a pace that no fixed benchmark can track, and the most capable of them are served behind endpoints that are updated without versioning and cannot be fine-tuned on CholeVidQA-32K under a matched recipe, so their scores are neither reproducible at a fixed version nor comparable to a controlled fine-tuned condition. The results reported here should therefore be read as characterizing the models evaluated rather than as establishing a standing ranking against the current frontier. Extending the benchmark to newer frontier multimodal models as their video interfaces, version pinning, and fine-tuning support mature remains necessary future work.}

\rev{\textbf{Temporal Scaling and Frame Selection:} The TEMP module improves long-context modeling by allowing denser candidate-frame sampling followed by text-guided frame selection, but the frame-sampling ablation shows that this benefit is not monotonic. In particular, increasing the candidate pool to 96 frames reduces performance under the current fixed top-$k$ selection design, suggesting that the selector can be challenged by larger sets containing redundant, transitional, or weakly informative frames. Thus, the current architecture should be understood as effective within the tested operating range rather than as a solution that scales without limit to arbitrarily long procedures. Future work should investigate adaptive top-$k$ selection, hierarchical chunk-level filtering, and multi-stage temporal retrieval mechanisms that can adjust the retained evidence according to video length and question complexity.}

\rev{\textbf{Scale of the Expert-Verified Test Subset:} The clinician-authored evaluation subset comprises $300$ QA pairs over $62$ videos, expanded during revision from an initial $121$ pairs over $25$ videos. At this size it supports comparison between models on clinician-written answers, but it is not a validation of the full CholeVidQA-32K generation pipeline, and per-task margins on its smaller strata---adverse events and skill proficiency, at $37$ pairs each---rest on few enough items that they should be read as indicative rather than as precise effect sizes. Enlarging this subset, and extending clinician authorship beyond the assessment level, is a priority for future work.}

\textbf{Handling Uncertainty and Abstention:} Surgical decision-making inherently involves uncertainty, particularly in ambiguous scenarios where visual evidence may be insufficient for definitive safety assessment. While our model achieves high answer rates (up to 100\% for certain tasks), indiscriminate response generation without uncertainty quantification poses potential safety risks in clinical deployment. Future work should incorporate calibrated confidence estimation mechanisms that enable the model to appropriately abstain from answering when evidence is insufficient, provide confidence intervals for safety assessments, and identify scenarios requiring additional clinical expertise or alternative imaging modalities for verification.

\rev{\textbf{Clinical Deployment Readiness:} This study is a retrospective evaluation of a single-procedure benchmark, not a validation of a real-time intraoperative decision-support system. The reported performance and run-to-run variability quantify benchmark behavior under the evaluated protocol, but do not address calibration of individual answers, safe abstention, human--AI interaction, workflow integration, latency, or clinical outcomes. Accordingly, SurgTEMP should be interpreted as a research framework for surgical video QA and a basis for future evaluation, rather than as a system ready to guide intraoperative decisions. Prospective multi-center studies with clinician oversight are required before considering clinical deployment.}

\rev{\textbf{Evaluation Dependence and Clinical Validity:} Although we triangulate performance using overlap metrics, structured classification after LLM-based verbalization, and multi-aspect LLM judgment, open-ended surgical QA evaluation remains partly dependent on the answer style induced by the dataset and the chosen evaluation prompts. The strong agreement among judge models supports the stability of model ranking under the tested automatic protocols, but it does not establish that observed gains correspond exclusively to improved clinical reasoning or would translate to clinical decision quality. In particular, a model may receive higher automatic scores by producing responses that better match the wording, granularity, or conventions of the constructed references. Future studies should therefore complement automatic evaluation with blinded expert adjudication of clinically consequential answers, prospective workflow-oriented assessment, and analyses that separate factual clinical correctness from stylistic alignment.}

\subsection{Conclusion}
This work addresses critical gaps in surgical data science by introducing both a comprehensive benchmark and a novel architectural solution for video question answering in laparoscopic cholecystectomy. We present CholeVidQA-32K, a hierarchical dataset encompassing 32K question-answer pairs across 11 distinct tasks organized into three levels of cognitive complexity—Perception (tool, action, and anatomical structure recognition), Assessment (Critical View of Safety, difficulty findings, adverse events, and skill proficiency evaluation), and Reasoning (surgical scene description, action rationale, comprehensive assessment, and intraoperative planning). Through systematic curation leveraging diverse annotation sources, our dataset establishes a foundation for developing and evaluating surgical VQA systems that extend beyond generic scene understanding.

To address the fundamental challenges of domain-specific visual interpretation, variable spatiotemporal granularity needs, and hierarchical task dependencies, we propose SurgTEMP with the Text-guided Memory Pyramid (TEMP) module. By utilizing textual queries as guiding signals for intelligent visual token selection through differentiable Gumbel-Softmax top-$k$ frame selection and constructing hierarchical memory structures that balance fine-grained spatial details with long-term temporal context, our architecture directly tackles the knowledge-driven nature of surgical videos where anatomical structures often lack strong visual contrast. Furthermore, our Surgical Competency Progression (SCP) training scheme explicitly models task dependencies, progressively training the model from fundamental perception capabilities to higher-level assessment and reasoning tasks.

Comprehensive evaluation using three complementary metric tiers—GPT-based scores (correctness, relevance, linguistic quality), overlap metrics (BLEU, METEOR, ROUGE-L, CIDEr), and classification metrics (balanced accuracy, F1-score)—reveals critical findings: while general-domain video MLLMs demonstrate reasonable zero-shot performance on basic perception tasks, their performance falls short on assessment-level tasks, underscoring the necessity of specialized domain adaptation and architectural solutions. Our SurgTEMP achieves substantial improvements over the baselines across all evaluation dimensions, with notable gains on assessment tasks and the long-context subset, validating that text-guided attention mechanisms and hierarchical temporal memory structures are essential for identifying and reasoning about subtle safety-critical visual cues across extended procedural contexts.

\section*{Acknowledgment }
% {\small

This work has received funding from the European Union (ERC, CompSURG, 101088553). Views and opinions expressed are however those of the authors only and do not necessarily reflect those of the European Union or the European Research Council. Neither the European Union nor the granting authority can be held responsible for them. 
This work was also partially supported by French state funds managed by the ANR under Grants ANR-10-IAHU-02 (IHU Strasbourg), ANR-23-IACL-0004 (AI Cluster Grand Est ENACT) and by the Interdisciplinary Thematic Institute HealthTech (ITI 2021-2028 program of the University of Strasbourg, CNRS and Inserm) via the IdEx Unistra (ANR-10-IDEX-0002) and SFRI (STRATUS project, ANR-20-SFRI-0012).
This work was also granted access to the servers/HPC resources managed by CAMMA, IHU Strasbourg and Unistra Mesocentre.

% \section*{Funding }

% \ifdefined\arxiv
%     \input{credit}
% \fi

\bibliographystyle{model2-names.bst}\biboptions{authoryear}
\bibliography{main}

\appendix
\setcounter{table}{0}
\setcounter{figure}{0}

\section{Data Generation Prompts}
\label{appendix:prompts}

\subsection{CholecT50-Caption-VQA Generation Prompt}
\label{appendix:cholec_prompt}

The following prompt was used to generate question-answer pairs from expert-annotated surgical video captions for the CholecT50-Caption-VQA subset:

\begin{framed}
\noindent\textbf{System Prompt:}
\vspace{0.1cm}

You are an experienced surgeon providing expert insights on laparoscopic cholecystectomy procedures.

\vspace{0.2cm}
\noindent\textbf{Task:} Generate Question-Answer pairs from expert-annotated video captions following a structured template covering: (1) anatomical space description with visual features and spatial relationships, (2) action recognition, (3) action motivation, (4) procedural errors or deviations, and (5) presence of smoke and lens fouling.

\vspace{0.2cm}
\noindent\textbf{Input Format:}\\
\texttt{Caption: \{caption\}}

\vspace{0.2cm}
\noindent\textbf{Output Requirements:}

\textbf{Type 1 - Conversational QA:} Create targeted questions isolating specific informational aspects (tools, anatomy, actions) from the caption. Generate multiple linguistically variable question-answer pairs in conversational tone. Question quantity should reflect caption content richness rather than template structure.

\textbf{Type 2 - Descriptive QA:} Synthesize multiple caption aspects into comprehensive questions requiring detailed elaboration of the surgical scene. Answers should integrate information holistically across caption elements.

\textbf{Type 3 - Reasoning QA:} Design questions requiring clinical reasoning and surgical expertise beyond explicit visual information. Base reasoning scenarios on action motivation from the caption to assess surgical decision-making and procedural rationale.

\vspace{0.2cm}
\noindent\textbf{Format:}\\
\texttt{\# Type 1: Conversational QA}\\
\texttt{Q1: [question] \\
A1: [answer] \\
Q2: [question] \\
A2: [answer] \\
...}\\
\\
\texttt{\# Type 2: Descriptive QA}\\
\texttt{Q1: [question] \\
A1: [answer]}\\
\\
\texttt{\# Type 3: Reasoning QA}\\
\texttt{Q1: [question] \\
A1: [answer]}

\vspace{0.2cm}
\noindent\textbf{Input:} Caption: \{caption\}
\end{framed}

\subsection{Endoscapes-VQA Generation Prompt}
\label{appendix:endoscapes_prompt}

The following prompt was used to generate CVS-focused question-answer pairs from annotated surgical keyframes for the Endoscapes-VQA subset:

\begin{framed}
\noindent\textbf{System Prompt:}
\vspace{0.1cm}

You are an experienced surgeon providing expert insights on Critical View of Safety (CVS) achievement in laparoscopic cholecystectomy.

\vspace{0.2cm}
\noindent\textbf{Task:} Generate Question-Answer pairs from annotated surgical keyframes using CVS annotation guidelines, checklists, and expert examples. Assess whether each CVS criterion is achieved based on visual evidence from color-coded bounding box annotations.

\vspace{0.2cm}
\noindent\textbf{Input Format:}\\
\texttt{Image: [surgical keyframe]}\\
\texttt{CVS Label: \{CVS\_label\}}\\
\texttt{CVS Guidelines: \{CVS\_instructions\}}\\
\texttt{CVS Annotation Checklist: \{CVS\_annotation\_checklist\}}\\
\texttt{CVS Annotation Examples: \{CVS\_flashcards\}}

\vspace{0.2cm}
\noindent\textbf{Visual Grounding:} Detected anatomical structures and tools are annotated with color-coded bounding boxes. Utilize these visual clues in assessment:
\begin{itemize}
    \item cystic\_plate: Green
    \item calot\_triangle: Blue
    \item cystic\_artery: Yellow
    \item cystic\_duct: Magenta
    \item gallbladder: Cyan
    \item tool: Red
\end{itemize}

\vspace{0.2cm}
\noindent\textbf{Output Requirements:}

Generate open-ended questions assessing whether each CVS criterion is achieved and why. Questions must not imply answers. Answers should follow step-by-step reasoning processes comparing visual clues with CVS annotation checklist and examples.

\vspace{0.2cm}
\noindent\textbf{Format:}\\
\texttt{Q1: [question] \\
A1: [answer]}\\
\\
\texttt{Q2: [question] \\
A2: [answer]}\\
\\
\texttt{Q3: [question] \\
A3: [answer]}

\vspace{0.2cm}
\noindent\textbf{Input:} Image: [keyframe], CVS Label: \{CVS\_label\}, CVS Guidelines: \{CVS\_instructions\}, CVS Annotation Checklist: \{CVS\_annotation\_checklist\}, CVS Annotation Examples: \{CVS\_flashcards\}
\end{framed}

\subsection{CholeScore-VQA Generation Prompt}
\label{appendix:lcod_prompt}

The following prompt was used to generate phase-level safety assessment question-answer pairs from multi-dimensional annotations for the CholeScore-VQA subset:

\begin{framed}
\noindent\textbf{System Prompt:}
\vspace{0.1cm}

You are an experienced surgeon providing expert insights on laparoscopic cholecystectomy safety assessment.

\vspace{0.2cm}
\noindent\textbf{Task:} Generate Question-Answer pairs for a phase-level surgical video clip using expert annotations from three safety assessment categories: operative difficulty (Nassar, Sugrue, PGS scales), intraoperative adverse events (SEVERE classification), and technical skill assessment (OSATS framework).

\vspace{0.2cm}
\noindent\textbf{Input Format:}\\
\texttt{Phase:\\ 
\{surgical\_phase\}}\\
\texttt{Findings:}\\
\{finding\_id\} - \{finding\_name\}: \{annotation\_value\}\\
\texttt{Description:\\
\{clinical\_description\}}

\vspace{0.2cm}
\noindent\textbf{Output Requirements:}

\textbf{Type 1 - Conversational QA:} Create targeted questions isolating individual findings without revealing annotations. Focus on clinical assessment rationale and include finding IDs for traceability.

\textbf{Type 2 - Descriptive QA:} Synthesize all findings into comprehensive safety assessments. Specify relevant scale systems and scores while considering synergistic effects between findings.

\textbf{Type 3 - Reasoning QA:} Design hypothetical scenarios modifying specific findings to assess reasoning about alternative procedural contexts and safety implications.

\vspace{0.2cm}
\noindent\textbf{Format:}\\
\texttt{\# Type 1: Conversational QA}\\
\texttt{(\{finding\_ID\}) \\
Q1: [question] \\
A1: [answer]}\\
\\
\texttt{\# Type 2: Descriptive QA}\\
\texttt{Q1: [question] \\
A1: [answer]}\\
\\
\texttt{\# Type 3: Reasoning QA}\\
\texttt{Q1: [question] \\
A1: [answer]}

\vspace{0.2cm}
\noindent\textbf{Input:} Phase: \{phase\}, Findings: \{findings\}
\end{framed}

%% =====================================================================
%% NEW IN REVISION: iterative quality-control feedback loop
%% Reviewer #1, "Opaque Quality Control"
%% =====================================================================
\section{\rev{Iterative Quality-Control Feedback Loop}}
\label{appendix:feedback_loop}

\rev{This appendix documents the iterative feedback loop referenced in
Section~\ref{subsubsec:detailed_curation_process}, i.e.\ what the clinical
review actually found, how each finding was attributed to a root cause, and
what was changed in the next iteration.}

\subsection{\rev{Review Protocol}}
\label{appendix:feedback_protocol}

\rev{After each generation round, we drew a stratified sample of 1\% of the
generated pairs (320 QA pairs), allocated proportionally across the 11 task
categories so that no task category was left unsampled. One clinical expert
reviewed every sampled pair against the four dimensions defined in
Section~\ref{subsubsec:detailed_curation_process}---clinical accuracy,
question clarity, language quality, and annotation consistency---and, for each
rejected pair, recorded a free-text comment describing why the pair was
unacceptable. These comments were then grouped into recurring failure modes,
and each failure mode was attributed to one of three root causes:}

\begin{enumerate}
\item \rev{\textbf{Prompt-specification gap}: the requirement violated by the
generator was never stated in the generation prompt. These were corrected by
adding the missing requirement to the prompt and regenerating the affected
task categories.}
\item \rev{\textbf{Annotation-grouping limit}: the annotation payload passed
into a single generation call was too large or too heterogeneous, so the
generator silently covered only part of it. These were corrected by reducing
the number of findings grouped per call so that each call is anchored to a
bounded set of annotations within one assessment category.}
\item \rev{\textbf{Instruction-following limit}: the requirement was stated in
the prompt but not honoured by the generator. Because re-prompting did not
reliably eliminate these cases, they were handled by deterministic
post-processing---either rule-based rewriting or removal of the offending
pairs---rather than by further prompt engineering.}
\end{enumerate}

\rev{Refined prompts and post-processing rules were applied to the affected
task categories, and a fresh stratified sample was drawn and re-reviewed. The
loop was terminated when a review round surfaced no failure mode that had not
already been addressed.}

\subsection{\rev{Identified Failure Modes and Corrective Actions}}
\label{appendix:feedback_modes}

\rev{Table~\ref{tab:qc_failure_modes} lists every recurring failure mode
surfaced by the review, together with the review dimension it violated, the
attributed root cause, the corrective action, and a pointer to the
corresponding before/after example in
Appendix~\ref{appendix:feedback_examples}. No recurring failure mode was
recorded for CholecT50-Caption-VQA, whose generation is conditioned on
clinician-written captions rather than on structured label payloads.}

\begin{table*}[t]
\centering
\revtable
\caption{\rev{Failure modes identified during the stratified 1\% expert review
of the generated QA pairs. Each mode is attributed to a root cause and mapped
to the corrective action applied in the next iteration of the curation loop.
``Ex.'' indexes the before/after examples in
Appendix~\ref{appendix:feedback_examples}; the two ungrounded-selection modes share example~E4.}}
\label{tab:qc_failure_modes}
\footnotesize
\begin{tabular}{@{}l p{0.24\textwidth} p{0.12\textwidth} p{0.17\textwidth} p{0.10\textwidth} l@{}}
\toprule
\textbf{Subset} & \textbf{Observed problem} & \textbf{Review dimension} & \textbf{Root cause} & \textbf{Action} & \textbf{Ex.} \\
\midrule
\multirow{6}{*}{CholeScore-VQA}
 & Answers describe the finding but never name the assessment scale it belongs to
   & Annotation consistency
   & Scale attribution not requested in the prompt
   & Refine prompt & E1 \\[0.3em]
 & Answers end with observations only, without an explicit graded conclusion
   & Clinical accuracy
   & Conclusion step not requested in the prompt
   & Refine prompt & E2 \\[0.3em]
 & Phases annotated with both difficulty and skill findings yield skill questions only
   & Annotation consistency
   & Too many findings grouped into one generation call
   & Refine grouping limit & E3 \\[0.3em]
 & Questions target findings that are rare or annotated as absent for the phase
   & Annotation consistency
   & Instruction-following limit
   & Remove & E4 \\[0.3em]
 & Questions generated for phases carrying no discriminative assessment annotation (e.g.\ Phase~6)
   & Question clarity
   & Instruction-following limit
   & Remove & E4 \\[0.3em]
 & Conversational questions not answerable from a single annotated finding (descriptive and reasoning types excepted)
   & Question clarity
   & Instruction-following limit
   & Remove & E5 \\
\midrule
\multirow{2}{*}{Endoscapes-VQA}
 & Answers cite the overlaid bounding-box annotations as visual evidence
   & Annotation consistency
   & Instruction-following limit
   & Rewrite & E6 \\[0.3em]
 & Questions not answerable from the binary CVS criterion label
   & Annotation consistency
   & Instruction-following limit
   & Remove & E7 \\
\bottomrule
\end{tabular}
\end{table*}

\subsection{\rev{Before/After Examples}}
\label{appendix:feedback_examples}

\rev{The examples below are representative of each failure mode; question and
answer text is abbreviated for space.}

\noindent\textbf{\rev{Ex.~E1 --- CholeScore-VQA, missing scale attribution
(refine prompt).}}
\begin{framed}
\revtable
\noindent\textbf{Q:} What does the appearance of the gallbladder in this phase
indicate about operative difficulty?

\vspace{0.15cm}
\noindent\textbf{Before:} The gallbladder appears distended and its wall looks
thickened, which makes grasping and retraction more difficult than usual.

\vspace{0.15cm}
\noindent\textbf{After:} The gallbladder is distended with a thickened wall.
Within the Nassar operative difficulty grading of the gallbladder, this
appearance corresponds to the distended/thick-walled category rather than a
floppy, thin-walled gallbladder, and it makes atraumatic grasping and
retraction more difficult.

\vspace{0.15cm}
\noindent\textbf{Prompt change:} the generation prompt now requires every
answer derived from a graded finding to name the scale the finding belongs to
(Nassar, Sugrue, PGS, SEVERE, or OSATS) and to place the observation within
that scale's categories.
\end{framed}

\noindent\textbf{\rev{Ex.~E2 --- CholeScore-VQA, missing graded conclusion
(refine prompt).}}
\begin{framed}
\revtable
\noindent\textbf{Q:} Based on the findings visible in this phase, how would you
assess the overall operative difficulty?

\vspace{0.15cm}
\noindent\textbf{Before:} There are adhesions around the gallbladder, the
hepatocystic triangle is partly obscured, and dissection progresses slowly.

\vspace{0.15cm}
\noindent\textbf{After:} There are adhesions around the gallbladder, the
hepatocystic triangle is partly obscured, and dissection progresses slowly.
Taken together, these findings place the phase at Nassar Grade~III (adhesions
obscuring the gallbladder neck and the hepatocystic triangle) rather than at
Grade~I--II, because the obscured hepatocystic triangle---not the pace of
dissection---is the dominant contributor, and adhesiolysis is required before
the critical structures can be exposed.

\vspace{0.15cm}
\noindent\textbf{Prompt change:} descriptive (comprehensive assessment) answers
must close with an explicit graded conclusion expressed on a named scale
(here the Nassar grade) and must state which finding drives that grade, instead
of terminating after the observations.
\end{framed}

\noindent\textbf{\rev{Ex.~E3 --- CholeScore-VQA, difficulty findings dropped
in favour of skill findings (refine grouping limit).}}
\begin{framed}
\revtable
\noindent\textbf{Before:} for a phase annotated with both operative-difficulty
findings and OSATS skill items, all generated conversational pairs targeted the
skill items; no question was generated for the annotated difficulty findings,
so the difficulty annotations for that phase were left unused.

\vspace{0.15cm}
\noindent\textbf{After:} the same phase yields conversational pairs for the
annotated difficulty findings and for the skill items, e.g.\ ``\textit{Is
there evidence of visceral obesity affecting exposure in this phase?}''
alongside ``\textit{How would you rate the surgeon's instrument handling during
this dissection?}''

\vspace{0.15cm}
\noindent\textbf{Pipeline change:} instead of passing the full phase-level
annotation payload in a single call, findings are grouped into smaller batches
anchored to one assessment category, so that the generator cannot satisfy the
output format by covering only the first category it encounters.
\end{framed}

\noindent\textbf{\rev{Ex.~E4 --- CholeScore-VQA, ungrounded findings and
non-discriminative phases (remove).}}
\begin{framed}
\revtable
\noindent\textbf{Before (rare or absent finding):} \textit{Q:} What does the
presence of a Mirizzi syndrome configuration imply for this dissection?
\textit{A:} It would require conversion to an open approach\ldots{} --- the
finding is not annotated as present for this phase, so the answer is
hypothetical rather than grounded in the source annotation.

\vspace{0.15cm}
\noindent\textbf{Before (non-discriminative phase):} \textit{Q:} What operative
difficulty is observed during Phase~6? \textit{A:} The phase proceeds without
notable difficulty\ldots{} --- the phase carries no discriminative difficulty,
adverse-event, or skill annotation, so the answer is uninformative.

\vspace{0.15cm}
\noindent\textbf{After:} both pairs are removed. Question generation is
restricted to findings annotated as present for the phase, and to phases
carrying at least one discriminative assessment annotation. Counterfactual
formulations are confined to the Intraoperative Planning task, where the
hypothetical framing is explicit.
\end{framed}

\noindent\textbf{\rev{Ex.~E5 --- CholeScore-VQA, conversational question not
anchored to a single finding (remove).}}
\begin{framed}
\revtable
\noindent\textbf{Before:} \textit{Q:} What factors contribute to the difficulty
of this dissection? --- the question spans several findings at once, so the
answer cannot be checked against any individual annotation, and the pair is
neither a conversational pair nor a comprehensive-assessment pair.

\vspace{0.15cm}
\noindent\textbf{After:} the pair is removed. Conversational pairs are retained
only when the question maps to one annotated finding and carries its finding ID
for traceability; multi-finding synthesis is covered by the descriptive
(comprehensive assessment) and reasoning (intraoperative planning) types, which
are exempt from this constraint by design.
\end{framed}

\noindent\textbf{\rev{Ex.~E6 --- Endoscapes-VQA, bounding-box leakage into the
answer (rewrite).}}
\begin{framed}
\revtable
\noindent\textbf{Q:} Is the two-structure criterion of the Critical View of
Safety achieved in this frame, and why?

\vspace{0.15cm}
\noindent\textbf{Before:} Yes. The cystic duct (magenta box) and the cystic
artery (yellow box) are both visible entering the gallbladder (cyan box), and
no third tubular structure is annotated.

\vspace{0.15cm}
\noindent\textbf{After:} Yes. Two tubular structures are seen entering the
gallbladder---the cystic duct medially and the cystic artery superolaterally---and
no additional tubular structure remains attached, which satisfies the
two-structure criterion.

\vspace{0.15cm}
\noindent\textbf{Post-processing change:} the bounding boxes are an annotation
aid supplied to the generator, not evidence available to a model at inference
time. Answers referring to box colours or to the overlay itself are rewritten
so that the justification is expressed in anatomical and spatial terms only.
\end{framed}

\noindent\textbf{\rev{Ex.~E7 --- Endoscapes-VQA, question not answerable from
the CVS label (remove).}}
\begin{framed}
\revtable
\noindent\textbf{Before:} \textit{Q:} How many anatomical structures can be
identified within the hepatocystic triangle in this frame? --- the source
supervision is the binary achievement label of each CVS criterion, so the
answer to a counting question is not verifiable against the annotation.

\vspace{0.15cm}
\noindent\textbf{After:} the pair is removed. Retained questions ask whether a
given CVS criterion is achieved and why, so that the achievement decision is
verifiable against the criterion label while the justification remains
open-ended.
\end{framed}

\section{\rev{Test-Time Frame-Budget Sweep for the Fine-Tuned Baselines}}
\label{appendix:baseline_frames}

\rev{This appendix reports the control experiment referenced in Section~\ref{subsec:param_sensitivity}: because SurgTEMP selects its frames from a $64$-frame candidate pool whereas the fine-tuned baselines receive uniformly sampled frames directly, we examine how those baselines themselves respond to the test-time frame budget.}

\begin{figure*}[t!]
    \centering
    \includegraphics[width=0.98\textwidth]{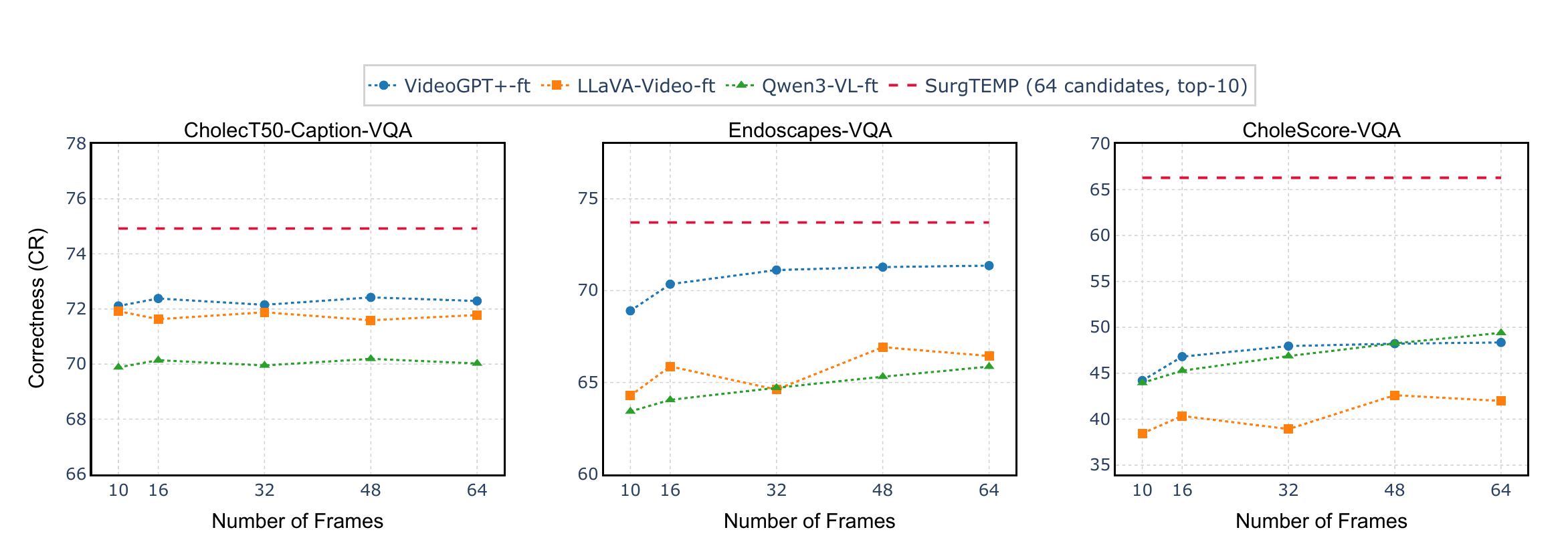}
    \caption{\rev{Test-time frame-budget sweep for the fine-tuned baselines. Correctness (CR) of LLaVA-Video-ft, VideoGPT+-ft, and Qwen3-VL-ft when evaluated at 10, 16, 32, 48, and 64 uniformly sampled frames, shown separately for each CholeVidQA-32K subset. The dashed line marks SurgTEMP at its operating point (64 candidate frames, top-10 selected), which consumes the same number of frames in the LLM as the 10-frame baseline setting.}}
    \label{fig:baseline_frames}
\end{figure*}

\rev{Figure~\ref{fig:baseline_frames} sweeps the number of uniformly sampled evaluation frames from 10 to 64 for LLaVA-Video-ft, VideoGPT+-ft, and Qwen3-VL-ft. The three subsets exhibit distinct behaviours. On CholecT50-Caption-VQA, all three baselines are essentially flat across the sweep, consistent with the short, action-consistent clips of this subset, where a handful of frames already covers the visual evidence required. On Endoscapes-VQA, a mild upward trend appears, reflecting that denser sampling of the 5-second clips helps localise the anatomical landmarks required for CVS assessment, though the effect saturates quickly. CholeScore-VQA shows the most visible test-time scaling, in line with its extended temporal context, but the gains remain modest relative to the spread between models and do not approach the margin held by SurgTEMP. On these two longer-context subsets, the baselines also differ in how they convert the extra frames into accuracy: VideoGPT+-ft realises most of its gain by 16--32 frames and plateaus thereafter; Qwen3-VL-ft scales smoothly and near-monotonically across the sweep, and is the only baseline still improving at 64 frames on CholeScore-VQA, where it overtakes VideoGPT+-ft; and LLaVA-Video-ft fluctuates non-monotonically with the frame budget, indicating that without a selection mechanism performance depends on which frames the uniform sampler happens to draw. None of these patterns is visible on CholecT50-Caption-VQA, where all three baselines stay within a narrow band across the entire sweep. Two observations follow. First, the 64-frame evaluation setting used in Tables~\ref{tab:subset_results} and~\ref{tab:overlap_metrics} is not disadvantageous for the baselines: it is at or near their best budget on every subset, so the comparison is not an artefact of over-feeding frames to models without a selection mechanism. Second, simply increasing the number of frames does not close the gap to SurgTEMP, which indicates that the benefit of denser temporal sampling is realised only when combined with query-conditioned selection rather than uniform sampling alone.}

% \section*{Supplementary Material}
% \noindent There will be no supplementary material.
% 

\end{document}